%% file: main.tex
\documentclass{iopjournal}

\usepackage{amssymb}
\usepackage{amsmath}
\usepackage{booktabs}
\usepackage{makecell}

\usepackage{xr}
\usepackage[backend=bibtex,style=ieee,doi=false,isbn=false,url=true,maxnames=10,minnames=10,giveninits=true]{biblatex}
\AtEveryBibitem{%
  \clearfield{urlyear}%
  \clearfield{urlmonth}%
  \clearfield{urlday}%
}
\newif\ifhighlight
\highlightfalse
\ifhighlight
  \usepackage[color,outerbars]{changebar}
  \cbcolor{red}
  \usepackage{lineno}
\else
  \newcommand{\cbstart}{}
  \newcommand{\cbend}{}
\fi

\newif\ifarXiv
 \arXivtrue
\ifarXiv
    \makeatletter
    \renewcommand{\articletype}[1]{}
    \makeatother
    \fancyhfoffset[L]{0mm}
    \renewcommand{\headrulewidth}{0pt} 
    
    \usepackage{graphicx}
    \usepackage{booktabs}
    \usepackage{rotating}      % for sideways table (S11)
    \usepackage{array}
    \usepackage{amssymb}
    \usepackage{caption}
    \usepackage{placeins}   % for \FloatBarrier
    
\fi

\begin{document}

\ifhighlight
  \linenumbers
\fi

\articletype{Paper} %	 e.g. Paper, Letter, Topical Review...

\title{ERP-XTTN: Interpretable Prototype-Guided Cross-Attention for Cross-Subject ERP Classification}

\author{Charlotte Genevier Wyman$^{1,*}$\orcid{0009-0000-8927-0485} and Leanne Hirshfield$^{1}$\orcid{0000-0003-0111-6948}}

\affil{$^1$University of Colorado Boulder, Boulder, CO, United States of America}

\affil{$^*$Author to whom any correspondence should be addressed.}

\email{charlotte.wyman@colorado.edu}

\keywords{EEG classification, brain-computer interface, event-related potentials, cross-subject generalization, interpretable deep learning, cross-attention}

\begin{abstract}
\input{01-abstract}
\end{abstract}

\section{Introduction}
\label{sec:introduction}
\input{02-introduction}

\section{Methods}
\label{sec:methods}
\input{03-methods}

\cbstart 
\section{Results}
\label{sec:results}
\input{04-results}
\cbend

\section{Discussion}
\label{sec:discussion}
\input{05-discussion}

\section{Conclusion}
\label{sec:conclusion}
\input{06-conclusion}

\ack{The authors declare no conflicts of interest. The authors used Claude Opus 4.6–4.8 (Anthropic) during the preparation of this work for the following purposes: editing human-written text for grammar, clarity, and structure; assisting with the drafting and debugging of analysis code; supporting literature review by helping locate relevant publications; 
\cbstart and generating the architecture diagram in Figure~\ref{fig:erp_xttn_architecture} using Figma\cbend. All AI-generated text was critically revised by the authors, all references were independently verified, and all data figures and quantitative results derive from the analyses reported in the paper, conducted on publicly available datasets. The authors reviewed and approved the final manuscript and take full responsibility for its content.}

\funding{This research received no external funding.}
% This section is a list of funder names and grant numbers

\roles{Charlotte Genevier Wyman: Conceptualization, Methodology, Software, Formal Analysis, Investigation, Writing -- Original Draft, Writing -- Review \& Editing, Visualization.\newline Leanne Hirshfield: Supervision, Writing -- Review \& Editing.}

\data{All datasets used in this study are publicly available.\\
BNCI Horizon 2020 ErrP dataset: \url{https://bnci-horizon-2020.eu/database/data-sets}\\
HRI ErrP dataset: \url{https://github.com/stefan-ehrlich/dataset-ErrP-HRI}\\
ERP CORE dataset: \url{https://erpinfo.org/erp-core}\\
ERP-XTTN code and results are available at \url{https://github.com/cgenevier/ERP-XTTN}; the version used to produce these results is archived on Zenodo \cite{wyman2026erpxttn_code}.}

\suppdata{\cbstart Supplementary material includes: Section~\ref{sup:datasets} (dataset and paradigm descriptions), Section~\ref{sup:full} (full-montage results), Section~\ref{sup:epmn-native} (EPMN native-recipe evaluation), Section~\ref{sup:validation} (architecture validation procedures: grounding interventions and ablation grid), Table~\ref{sup:tab-fullmontage-auroc} (full-montage AUROC), Table~\ref{sup:tab-fullmontage-ba} (full-montage balanced accuracy), Tables~\ref{sup:tab-grounding-full-routing}--\ref{sup:tab-grounding-full-amp} (routing and amplitude grounding interventions, full montage), Table~\ref{sup:tab-epmn-native} (EPMN native-recipe comparison), Table~\ref{sup:tab-balacc} (three-channel balanced accuracy), Tables~\ref{sup:tab-persubj-hri}--\ref{sup:tab-persubj-n400} (per-subject AUROC for all nine datasets at the three-channel montage), Table~\ref{sup:tab-wilcoxon} (full Wilcoxon signed-rank statistics), Tables~\ref{sup:tab-corr}--\ref{sup:tab-corr-perseed} (outcome-conditioned waveform correlations and per-seed stability), Tables~\ref{sup:tab-grounding-3-routing}--\ref{sup:tab-grounding-3-amp} (routing and amplitude grounding interventions, three-channel montage), Table~\ref{sup:tab-ablation} (architecture ablation grid), Table~\ref{sup:tab-polarity-contrast} (polarity inversion under the grounded versus learned readout), Figure~\ref{sup:fig-prototypes-all} (difference-wave prototypes for all nine datasets), and Figure~\ref{sup:fig-morphology-all} (outcome-conditioned grand-average waveforms for all nine datasets).\cbend}

\section*{References}
\printbibliography[heading=none]

\ifarXiv

    \clearpage
    
    % -- S-style numbering for supplementary tables/figures ---
    \renewcommand{\thesection}{S\arabic{section}}
    \renewcommand{\thesubsection}{S\arabic{section}.\arabic{subsection}}
    \setcounter{section}{0}
    \renewcommand{\thetable}{S\arabic{table}}
    \renewcommand{\thefigure}{S\arabic{figure}}
    \setcounter{table}{0}
    \setcounter{figure}{0}
    
    \section*{Supplementary Material}
    This document contains supplementary material for the main manuscript. Section~\ref{sup:datasets} provides full dataset and paradigm descriptions. Section~\ref{sup:full} reports classification and grounding results at the full available montage. Section~\ref{sup:epmn-native} evaluates EPMN under its native training recipe. Section~\ref{sup:validation} specifies the architecture validation procedures, including the prototype-corruption ladder, routing permutation controls, amplitude window-localization controls, and the single-knob ablation grid.

    Tables are organized as follows: Table~\ref{sup:tab-fullmontage-auroc} (full-montage AUROC) and Table~\ref{sup:tab-fullmontage-ba} (full-montage balanced accuracy); Tables~\ref{sup:tab-grounding-full-routing}--\ref{sup:tab-grounding-full-amp} (routing and amplitude grounding interventions at the full montage); Table~\ref{sup:tab-epmn-native} (EPMN native-recipe comparison); Table~\ref{sup:tab-balacc} (three-channel balanced accuracy); Tables~\ref{sup:tab-persubj-hri}--\ref{sup:tab-persubj-n400} (per-subject AUROC for all nine datasets at the three-channel montage, one table per dataset); Table~\ref{sup:tab-wilcoxon} (full Wilcoxon signed-rank statistics); Tables~\ref{sup:tab-corr}--\ref{sup:tab-corr-perseed} (outcome-conditioned waveform correlations and per-seed stability); Tables~\ref{sup:tab-grounding-3-routing}--\ref{sup:tab-grounding-3-amp} (routing and amplitude grounding interventions at the three-channel montage); Table~\ref{sup:tab-ablation} (architecture ablation grid); and Table~\ref{sup:tab-polarity-contrast} (polarity inversion under the grounded versus learned readout). Figure~\ref{sup:fig-prototypes-all} shows difference-wave prototypes across all nine datasets, and Figure~\ref{sup:fig-morphology-all} shows outcome-conditioned grand-average waveforms for all nine datasets.

    \section{Dataset and Paradigm Descriptions}
    \label{sup:datasets}
    \input{supplementary-datasets}
    
    \section{Full-Montage Results}
    \label{sup:full}
    \input{supplementary-full}

    \section{EPMN Native}
    \label{sup:epmn-native}
    \input{supplementary-epmn}

    \section{Architecture Validation: Full Procedures}
    \label{sup:validation}
    \input{supplementary-validation}

    \input{supplementary-tables}
    \clearpage
    
    \input{supplementary-figures}
    \clearpage

\fi

\end{document}

%% file: 01-abstract.tex
\textit{Objective:} Interpretable brain-computer interface classifiers that generalize across subjects without calibration remain an open challenge. We evaluated whether prototype-based cross-attention can provide competitive, inherently interpretable event-related potential (ERP) classification across diverse paradigms under deployment-compatible conditions. 
\textit{Approach:} We propose ERP-XTTN (ERP Cross-Attention), a cross-attention architecture that routes 
\cbstart 
input electroencephalographic peaks to fixed difference-wave prototypes via query-key-only cross-attention with no value projection. Classification is based directly on prototype similarity and a separate measure of component amplitude, so that the prototype content contributes to every decision by construction. 
\cbend
Prototypes are derived automatically from prominent extrema in the training-fold grand-average difference wave. We evaluated across three public sources (BNCI Horizon 2020, HRI Cursor, and ERP CORE) encompassing eight ERP components (ERN, LRP, ErrP, N170, P300, N2pc, MMN, N400). Evaluations used leave-one-subject-out (LOSO) cross-validation with causal filtering at a three-channel montage, compared against EEGNet, 
\cbstart 
EEG-Deformer, ERP Prototypical Matching Net (EPMN), and
\cbend
xDAWN with Riemannian geometry (xDAWN+RG). 
\textit{Main results:} At three channels, the mean performance gap between the best baseline and ERP-XTTN was 
\cbstart 
0.025 area under the receiver operating characteristic curve (AUROC). Prototype interventions confirmed that decisions depend on the physiological content of the prototypes rather than on the routing attention pattern alone. False positives morphologically resembled true positives more than true negatives did across all datasets,
\cbend
indicating classification errors are neurophysiologically explicable. \textit{Significance:} ERP-XTTN generalizes across diverse ERP morphologies under causal, calibration-free conditions, while 
\cbstart 
retaining competitive performance and decisions that depend directly on physiological prototype content at a three-channel montage. Unlike post-hoc explanation methods for black-box models, the basis of each decision is directly observable in the trained model itself. 
\cbend
To our knowledge, this is the first epoch-level LOSO benchmark on ERP CORE. 

%% file: 02-introduction.tex
Event-related potentials (ERPs) are time-locked neural responses measurable via electroencephalography (EEG) that index sensory, attentional, and cognitive processing \cite{luck2014} and underlie a range of brain-computer interface (BCI) applications such as error detection and spelling systems \cite{wolpaw2002}. Deployment-ready ERP classification requires meeting several constraints simultaneously. For online compatibility, filtering must be causal: acausal filtering incorporates future samples, producing accuracy estimates that cannot be reproduced in any forward-processing pipeline. Practical deployment often limits channel count, as many commercial and mobile devices lack research-grade electrode coverage \cite{sabio2024}. Classification must generalize across subjects, since per-subject calibration delays setup and limits scalability \cite{lotte2015}, and standard calibration pipelines fail to produce a functional classifier for a non-negligible portion of users \cite{vidaurre2010}. Finally, clinical and safety-critical applications demand that classification decisions be interpretable, not merely accurate.  

These constraints are compounded by the diversity of ERP morphologies. ERPs span frontocentral error responses such as the error-related negativity (ERN), posterior visual components such as the N170, diffuse, low signal-to-noise ratio (SNR) responses such as N400, lateralized components such as N2pc and lateralized readiness potential (LRP), passive auditory responses such as mismatch negativity (MMN), and broader error-related potential (ErrP) families \cite{luck2014}. A general cross-subject framework must handle this diversity without requiring component-specific architecture or preprocessing choices. 

Cross-subject ERP classification has been addressed from multiple directions. General-purpose deep learning architectures such as EEGNet \cite{lawhern2018, solon2017, lotte2018} have been evaluated across multiple BCI paradigms; CNN-transformer architectures have demonstrated competitive cross-subject ErrP performance under leave-one-subject-out (LOSO) evaluation \cite{ren2023, ren2024}; 
\cbstart 
and dense convolutional transformers such as EEG-Deformer \cite{ding2024} have improved decoding across a range of EEG tasks by capturing coarse- and fine-grained temporal dynamics within each transformer block. These architectures are not inherently interpretable, however; post-hoc approaches to explaining their decisions are discussed below. 
\cbend
Classical pipelines such as shrinkage linear discriminant analysis (LDA) \cite{blankertz2011} achieve strong cross-subject ERP classification and can reveal discriminative spatial patterns, but their interpretability rests on a fixed decision rule rather than trial-specific evidence for each classification. xDAWN spatial filtering \cite{rivet2009} with Riemannian geometry (xDAWN+RG) features also perform well cross-subject \cite{li2020, barachant2014} but lack obvious physiological interpretation \cite{lotte2018}. Calibration-free cross-subject ErrP classification has been demonstrated with generalized LDA \cite{schonleitner2020}, an xDAWN+RG pipeline with a support vector machine classifier \cite{kumar2021}, ensemble classifiers \cite{bhattacharyya2017}, and online PCA-based pipelines \cite{lopes-dias2021}, though without inherent interpretability. Domain adaptation and domain generalization methods address cross-subject distribution shift in ERP-based BCIs \cite{wu2017, wu2022, luo2026}, but are not designed for interpretability.  

Interpretability for deep EEG architectures has been pursued post-hoc and under offline preprocessing, including EEGNet's feature-visualization and DeepLIFT relevance analyses \cite{lawhern2018} and Grad-CAM applied to ERP classifiers \cite{jalilpour2025}. Post-hoc importance estimation has also been used to guide channel and time-window selection for ErrP decoding, with the estimates validated against error-processing neurophysiology \cite{chan2025}. These methods can identify features associated with a trained model's predictions and map them to known neurophysiological phenomena, in some cases at the single-trial level, but because they are approximations computed after training, their faithfulness to the decision process is not guaranteed; they are not evidence that is, by construction, the basis of each classification. 
\cbstart 
Prototype-based approaches, including ERP Prototypical Matching Net (EPMN) \cite{wei2022} and Attention-ProNet \cite{zhang2024}, achieve zero-calibration cross-subject decoding in P300-based rapid serial visual presentation (RSVP) tasks by matching trials to class prototypes in a learned embedding space. Because these prototypes are abstract summaries of features learned by the model rather than recognizable EEG waveforms, they cannot be interpreted directly in neurophysiological terms, and the approach has been demonstrated only within a single paradigm.
\cbend
Discriminative canonical pattern matching with time-domain ERP templates has been demonstrated across multiple ERP paradigms but only within-subject \cite{xiao2020}.  

Cross-subject evaluation on ERP CORE has been reported using fold-based splits \cite{aristimunha2023} and time-point-wise LOSO decoding \cite{qin2026}, but neither provides an epoch-level LOSO classification benchmark under deployment-compatible constraints such as causal filtering. To our knowledge, no prior approach combines inherent interpretability, LOSO evaluation, causal filtering, and coverage across diverse ERP components.

\cbstart 
This work demonstrates that prototype-guided attention routing can be constrained so that prototype content is necessary to every classification, and that the resulting classifier remains competitive across diverse ERP paradigms under deployment-compatible constraints. ERP-XTTN (ERP Cross-Attention) \cite{wyman2026} was introduced for error-related potentials, using a constrained, hand-specified polarity sequence to locate prototype windows based on the training subjects’ grand-average difference wave. Here, we redesign the decision pathway so that every prediction depends directly on the physiological content of the prototypes – a property we term \emph{groundedness} – verified empirically by corrupting prototypes at inference and observing that classification degrades systematically, with polarity inversion reversing discriminative performance rather than merely abolishing it. We additionally generalize ERP-XTTN to eight ERP components via 
\cbend
automatic peak detection, requiring no component-specific architecture, preprocessing, or training choices. We evaluate across three public sources (BNCI Horizon 2020, HRI Cursor, and ERP CORE) encompassing eight ERP components, using LOSO cross-validation with causal infinite impulse response (IIR) filtering 
\cbstart 
at a minimal three-channel montage, and compare against EEGNet, EEG-Deformer, EPMN,
\cbend
and xDAWN+RG. To our knowledge, this is the first epoch-level LOSO benchmark on ERP CORE under deployment-compatible constraints. Beyond classification performance, we 
\cbstart 
show that the model’s classification errors are neurophysiologically explicable. 
\cbend

%% file: 03-methods.tex
This section describes the nine datasets, the shared preprocessing pipeline, the ERP-XTTN architecture, the 
\cbstart 
four baseline methods, the training procedure, the evaluation protocol and statistical analyses, and the architecture validation procedures. 
\cbend

\subsection{Datasets}
\label{sec:datasets}
\input{03-methods-01-datasets}

\subsection{Preprocessing}
\label{sec:preprocessing}
\input{03-methods-02-preprocessing}

\subsection{Architecture (ERP-XTTN)}
\label{sec:architecture}
\cbstart 
\input{03-methods-03-architecture}
\cbend

\subsection{Baselines}
\label{sec:baselines}
\input{03-methods-04-baselines}

\subsection{Training}
\label{sec:training}
\input{03-methods-05-training}

\subsection{Evaluation}
\label{sec:evaluation}
\input{03-methods-06-evaluation}

\cbstart 
\subsection{Architecture Validation}
\label{sec:architecture-validation}
\input{03-methods-07-architecturevalidation}
\cbend

%% file: 03-methods-01-datasets.tex
Nine datasets spanning eight ERP components, with ErrP represented by two datasets, were evaluated 
\cbstart 
at a targeted three-channel montage. They are drawn from three public sources: the BNCI Horizon 2020 dataset (BNCI) \cite{chavarriaga2010}, the HRI Cursor ErrP dataset (HRI) \cite{ehrlich2019}, and ERP CORE \cite{kappenman2021}. Together they cover frontocentral error responses (ERN, ErrP), lateralized motor and attentional components (LRP, N2pc), and stimulus-evoked sensory and cognitive responses (N170, MMN, N400, P300), spanning a wide range of ERP morphologies. Trials were classified per dataset as error vs. correct (BNCI, HRI, ERN), left- vs. right-hand (LRP), deviant vs. standard (MMN), face vs. car (N170), left- vs. right-target (N2pc), unrelated vs. related (N400), and target vs. non-target (P300). 
\cbend
The original ERP CORE publication \cite{kappenman2021} reports smaller per-component samples (N = 34–39) after exclusions based on behavioral-accuracy and artifact-rejection criteria tied to their ERP-averaging pipeline. We retained all 40 participants because our preprocessing omits the baseline correction, re-referencing, and ocular-artifact correction those criteria presuppose, and because retaining all participants provides a more stringent test of cross-subject generalization. 
\cbstart 
Table~\ref{tab:datasets} contains a high-level summary of the datasets; full per-paradigm descriptions are provided in Supplementary Section~\ref{sup:datasets}.
\cbend
This study used only publicly available, de-identified datasets; no new human subjects data were collected.  

\cbstart 
\setlength{\tabcolsep}{2pt}
\begin{table}
\caption{Datasets evaluated. Electroencephalography (EEG) channels only; electrooculography and reference channels were excluded where applicable. Full Channels (Full Ch) gives the full montage recorded for each dataset; the 3-channel montage is the evaluation condition reported in the main text, chosen to cover the scalp region of the event-related potential (ERP) of interest. The detection channel is the single electrode on which prototype time windows are located by peak detection (Section~\ref{sec:prototypes}); for the ERP CORE datasets it is the canonical site reported for that component by Kappenman et al.~\cite{kappenman2021}, and for BNCI and HRI it is Cz, the central midline site available in both montages. Trials/Subject refers to total classification epochs across both classes, summed across sessions where applicable (BNCI includes two sessions per subject). BNCI: BNCI Horizon 2020 dataset; HRI: human--robot interaction; ERP CORE: ERP Compendium of Open Resources and Experiments; ErrP: error-related potential; ERN: error-related negativity; LRP: lateralized readiness potential; MMN: mismatch negativity; N2pc: N2 posterior contralateral.}
\label{tab:datasets}
\centering
\begin{tabular}{lcccccc}
\hline
\textbf{Dataset / ERP} & \textbf{Time-Locked} & \textbf{Total} & \textbf{Trials/Subject} & \textbf{Full} & \textbf{3-Ch} & \textbf{Detection} \\
 & \textbf{Event} & \textbf{Subjects} & \textbf{(approx)} & \textbf{Ch} & \textbf{Montage} & \textbf{Channel} \\
\hline
BNCI / ErrP & Feedback & 6 & 1073 & 64 & Fz, Cz, Pz & Cz \\
HRI / ErrP & Feedback & 11 & 483 & 27 & Fz, Cz, Pz & Cz \\
ERP CORE / ERN & Response & 40 & 401 & 30 & FCz, Cz, Pz & FCz \\
ERP CORE / LRP & Response & 40 & 356 & 30 & C3, Cz, C4 & C3 \\
ERP CORE / MMN & Stimulus & 40 & 981 & 30 & FCz, Cz, Pz & FCz \\
ERP CORE / N170 & Stimulus & 40 & 160 & 30 & PO7, Oz, PO8 & PO8 \\
ERP CORE / N2pc & Stimulus & 40 & 320 & 30 & PO7, Pz, PO8 & PO7 \\
ERP CORE / N400 & Stimulus & 40 & 120 & 30 & Cz, CPz, Pz & CPz \\
ERP CORE / P300 & Stimulus & 40 & 200 & 30 & Fz, Cz, Pz & Pz \\
\hline
\end{tabular}
\end{table}
\cbend

%% file: 03-methods-02-preprocessing.tex
A causal IIR 4th-order Butterworth 1–10 Hz bandpass filter was applied to all datasets. This frequency range was selected to capture the primary spectral content of ERP components while attenuating muscle artifact and high-frequency noise, and was fixed across all datasets prior to evaluation. 
\cbstart 
Causal filtering ensures that each output sample depends only on past input, so that reported performance is reproducible in an online, forward-processing pipeline. 
\cbend
BNCI was downsampled from 512 Hz to 256 Hz after filtering. HRI was recorded at 1000 Hz but distributed pre-downsampled to 256 Hz, requiring no further downsampling. All ERP CORE datasets were downsampled from 1024 Hz to 256 Hz after filtering. 

Baseline correction, re-referencing, and spatial filtering were omitted to evaluate performance under constraints compatible with real-time, zero-calibration deployment, and all datasets were epoched to 0–800 ms after the dataset-specific time-locking event. For the response-locked ERP CORE LRP and ERN paradigms, this window captures post-response lateralized motor activity and error processing respectively, rather than pre-response preparatory components. All datasets were evaluated using 
\cbstart 
a 3-channel set tailored to the ERP of interest (Table~\ref{tab:datasets}); results using the full montage available for each dataset are reported in Supplementary Table~\ref{sup:tab-fullmontage-auroc}. 
\cbend

%% file: 03-methods-03-architecture.tex
\begin{figure}
 \centering
 \includegraphics[width=0.99\textwidth]{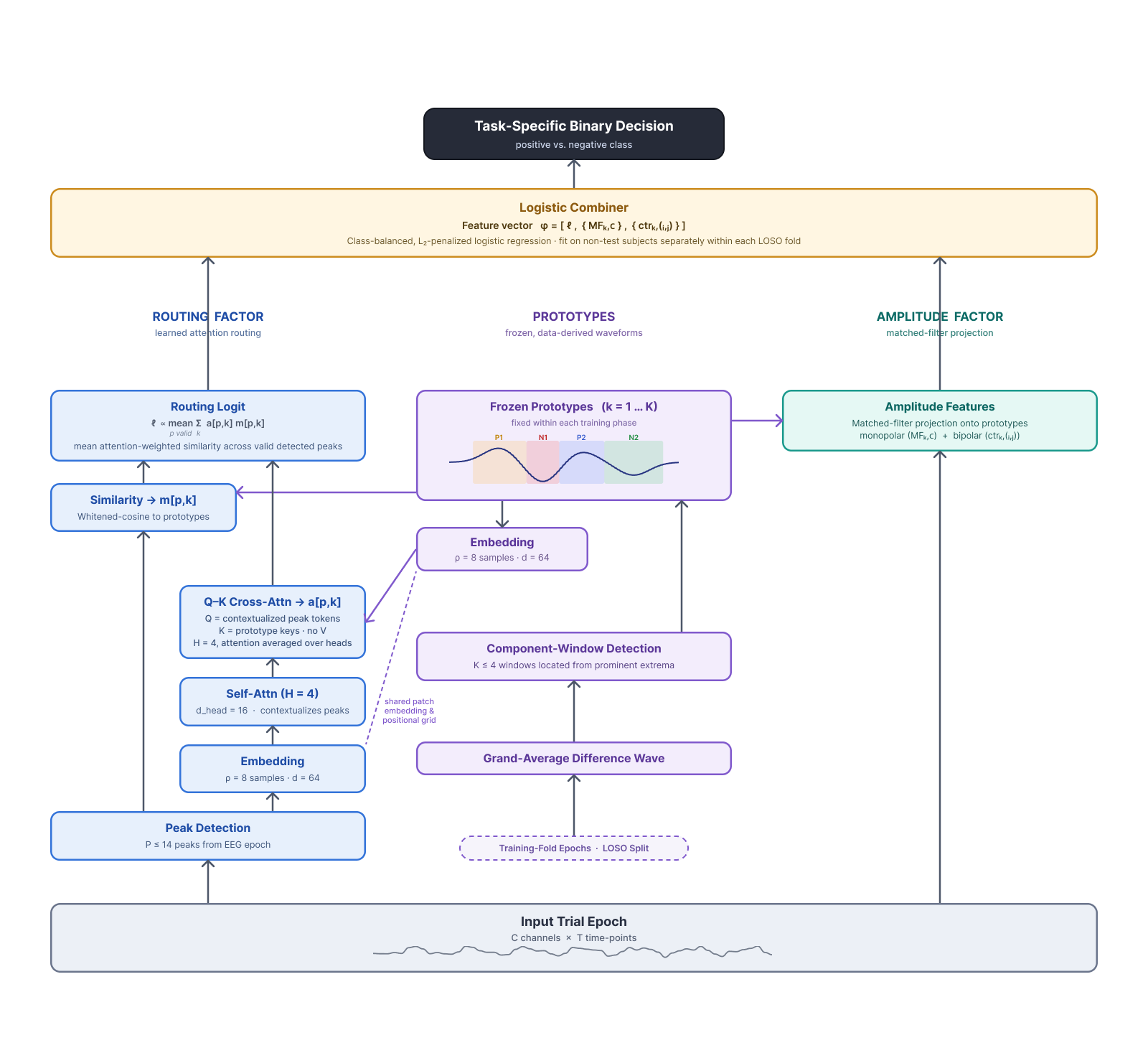}
 \caption{ERP-XTTN architecture. The model classifies from two factors defined against the same frozen prototypes. The routing factor (left) tokenizes the input trial into detected peaks, embeds them, contextualizes them via multi-head self-attention, and computes query-key (QK)-only cross-attention against prototype keys; the routing logit is the attention-weighted sum of whitened-cosine similarities $m[p,k]$ between each peak $p$ and each prototype $k$, which carry no learned parameters. The amplitude factor (right) projects the trial onto the same prototypes via monopolar and bipolar matched filters. Prototypes (center) are derived from the training fold's grand-average difference wave by automatic peak detection on the detection channel (Section~\ref{sec:prototypes}); up to $K = 4$ component windows are retained. The two factors are combined by $L_2$-penalized logistic regression fit on the remaining training subjects, so that no data from the held-out subject informs either factor or the combiner.}
\label{fig:erp_xttn_architecture}
\end{figure}

The proposed model, ERP-XTTN, classifies trial epochs from two factors, each defined against the same set of fold-specific ERP prototypes (Figure~\ref{fig:erp_xttn_architecture}). The first factor, \emph{routing}, measures how closely the trial's own deflections resemble each prototype in shape, and where in the trial they do so. The trial is tokenized into its own detected peaks, which serve as queries in a query-key (QK)-only cross-attention against the prototypes, and the decision is read out as the attention-weighted similarity between each peak and each prototype. That similarity is computed directly from the signal and carries no learned parameters, so prototype content enters every decision by construction; the learned attention weights determine only how those similarities are combined. The second factor, \emph{amplitude}, measures how large each component is at its own latency, by matched filtering the trial against the same prototypes in both monopolar and bipolar channel derivations. The two factors are combined by a fold-specific $L_2$-penalized logistic regression fit only to the training subjects, so that no data from the held-out subject informs either the routing factor or the combiner. 

The subsections below describe the peak units (Section~\ref{sec:units}), prototype construction (Section~\ref{sec:prototypes}), the similarity that defines the match (Section~\ref{sec:match}), the routing readout (Section~\ref{sec:routing-readout}), and the amplitude factor and combiner (Section~\ref{sec:amplitude}). Design choices introduced here are tested empirically in Section~\ref{sec:architecture-validation}, and the results of those tests are reported in Section~\ref{sec:validation}. 

Throughout, a trial epoch is written $x \in \mathbb{R}^{C \times T}$ for $C$ channels and $T$ time samples ($T = 206$ at 256~Hz, $C = 3$ in the three-channel condition), and $x_{\mathrm{det}} \in \mathbb{R}^{T}$ denotes the signal at the dataset's detection channel (Table~\ref{tab:datasets}). Quantities estimated from the training fold are used unchanged for the held-out subject.

% --- PEAK UNITS --- %
\subsubsection{Peak Units}\label{sec:units}
ERP-XTTN tokenizes each trial into its own detected deflections, so that each token corresponds to an identifiable feature of that trial's waveform. The detection-channel signal is smoothed with a Gaussian kernel ($\sigma_t = 3$ samples, reflect-padded). Candidate peaks are the strict local extrema of the smoothed signal (both polarities) whose topographic prominence exceeds $0.02$ in $z$-scored units, with non-maximum suppression at a minimum separation of $40$~ms applied separately to positive and negative candidates before pooling. No prescribed polarity sequence or minimum latency is imposed. The $P = 14$ most prominent peaks are retained and restored to temporal order; trials yielding fewer are padded, and a mask $\mu_p \in \{0,1\}$ marks valid slots, with $P^{\mathrm{val}} = \sum_p \mu_p$. 

Each peak is bounded by its flanking inflection points (sign changes of the discrete second derivative of the smoothed signal). Windows are variable in width, are neither clamped nor constrained to be disjoint, and default to the epoch boundaries where no inflection is found.

Each peak segment is resampled to a fixed width $\rho = 8$ samples and linearly projected to a $d = 64$ dimensional embedding, 
\begin{equation} 
u_p = \mathrm{Resample}_{\rho}\big( x[:, s_p{:}e_p] \big) \in \mathbb{R}^{C \times \rho}, \qquad z_p^{0} = W_e\, \mathrm{vec}(u_p) + b_e, 
\label{eq:embed} 
\end{equation} 
and positionally encoded at its temporal centre by linearly interpolating the learned positional table $E \in \mathbb{R}^{n_g \times d}$ ($n_g = \lfloor T/\rho \rfloor = 25$) at the fractional patch index of the peak. Padded slots are zeroed and excluded from self-attention and from the readout. Dropout ($p = 0.3$) is applied to the peak embeddings during training. 

Peak detection is performed inside the forward pass, on the augmented signal, rather than as a preprocessing step; tokenization is therefore part of the model and is subject to the same augmentation as the rest of the input (Section~\ref{sec:training}).

% --- PROTOTYPES --- %
\subsubsection{Prototypes}\label{sec:prototypes}
Prototypes are component waveforms derived from the training fold's grand-average difference wave. For classes labeled $y = 1$ (target, error, or the equivalent positive class of each paradigm; Supplementary Section~\ref{sup:datasets}) and $y = 0$, 
\begin{equation} 
\bar{d} = \frac{1}{|\{y = 1\}|} \sum_{n : y_n = 1} x_n \;-\; \frac{1}{|\{y = 0\}|} \sum_{n : y_n = 0} x_n \;\in\; \mathbb{R}^{C \times T}, 
\label{eq:diffwave} 
\end{equation} 
with $\bar{s} = G_{\sigma_p}(\bar{d}_{\mathrm{det}})$ its smoothed detection-channel trace ($\sigma_p = 2$ samples). 

Component windows are located on $\bar{s}$ by the same prominence criterion used for trial peaks ($0.02$), subject to a minimum latency of 50~ms, a minimum separation of 80~ms, and a requirement that the signal value at each retained extremum matches its polarity sign. The most prominent such deflections are retained, up to a maximum of $K_{max} = 4$. Each is expanded to the interval between its flanking zero crossings of $\bar{s}$, clamped to a width of 40--200~ms and constrained to be non-overlapping. Deflections whose window fails to contain them are discarded, so the number of prototypes $K \leq 4$ adapts per fold to the number of components the difference wave actually supports. 

Each prototype $k$ contributes two quantities over its window $[s_k, e_k)$ of native width $w_k = \min(e_k - s_k,\, w_{\max})$, $w_{\max} = \lfloor 0.2 f_s \rceil$: a routing key $\Pi_k = \mathrm{Resample}_{\rho}(\bar{d}[:, s_k{:}e_k]) \in \mathbb{R}^{C \times \rho}$, embedded on the same path as the peak units, and a prototype waveform $\tau_k = \bar{d}[:, s_k{:}e_k] \in \mathbb{R}^{C \times w_k}$, which is used for the match (Section~\ref{sec:match}) and for the amplitude factor (Section~\ref{sec:amplitude}). Prototypes are frozen during training.

% --- SIMILARITY --- %
\subsubsection{Similarity}\label{sec:match}
Peak segments and prototypes are compared in shape only, independently of amplitude, after joint spatiotemporal whitening estimated from within-class residuals on the training fold. A fixed shrinkage coefficient ($0.9$) toward a scaled identity keeps the estimate well conditioned; full details of the whitening procedure are provided in the released code~\cite{wyman2026erpxttn_code}. Critically, the match is evaluated at the prototype's \emph{native} window width $w_k$, not at the embedding width $\rho$. 

The match between peak $p$ and prototype $k$ is the cosine similarity of their whitened representations:
\begin{equation} 
m[p,k] = \frac{\langle v_{p,k},\, \tilde{\tau}_k \rangle}{\lVert v_{p,k} \rVert \, \lVert \tilde{\tau}_k \rVert} \; \in [-1, 1], 
\label{eq:match} 
\end{equation} 
where $\tilde{\tau}_k$ is the whitened prototype and $v_{p,k}$ is the peak segment resampled to width $w_k$ and whitened by the same operator. Equation~\eqref{eq:match} contains no learned parameters and receives no gradient.

% --- ROUTING READOUT --- %
\subsubsection{Routing Readout}\label{sec:routing-readout}
Peak embeddings first pass through a single pre-normalized multi-head self-attention layer ($H = 4$ heads, $d_h = 16$) with residual connection, so that each peak is represented in the context of the other deflections in the same trial; padded slots are masked and dropout ($p = 0.3$) is applied to the attention weights during training. Prototype keys are formed on the shared embedding path and positionally encoded at the patch index of their window midpoint. Cross-attention is then computed with peaks as queries and prototypes as keys, using separate layer normalizations and linear projections. No value projection is used; the per-head softmax distributions are averaged across all $H$ heads to produce attention weights $a[p,k]$. 

The routing logit is the attention-weighted sum of the matches, normalized by the number of valid peaks so that trials with different numbers of detected deflections are comparable: 
\begin{equation} 
\ell_{\mathrm{route}} = s_{\mathrm{scale}} \cdot \frac{\sum_{p,k} a[p,k]\, m[p,k]\, \mu_p}{\max\big(1,\, P^{\mathrm{val}}\big)} + b. 
\label{eq:routelogit} 
\end{equation} 
The learned quantities in Equation~\eqref{eq:routelogit} are the attention weights $a$ and the scalars $s_{\mathrm{scale}}$ and $b$. The matches $m$, the prototypes, and the whiteners are fixed.

% --- AMPLITUDE AND COMBINER --- %
\subsubsection{Amplitude and Combiner}\label{sec:amplitude}
A second factor measures component amplitude by matched filtering the trial against the same prototypes in monopolar and bipolar channel derivations. The monopolar projection measures how strongly the trial expresses prototype $k$ at channel $c$ over that prototype's window, and the bipolar projection measures the same quantity in the derivation $i - j$, capturing the spatial gradient the prototype predicts between two sites: 
\begin{align} 
\mathrm{MF}_{k,c} &= \frac{\big\langle x[c, s_k{:}e_k],\; \tau_k[c] \big\rangle}{\lVert \tau_k \rVert_F}, 
\label{eq:mfchan} \\[2pt] 
\mathrm{ctr}_{k,(i,j)} &= \frac{\big\langle x[i, s_k{:}e_k] - x[j, s_k{:}e_k],\; \tau_k[i] - \tau_k[j] \big\rangle}{\lVert \tau_k \rVert_F}, \qquad i < j . 
\label{eq:mfctr} 
\end{align} 
Both are normalized by the prototype norm only.  

The routing factor and amplitude factor are combined in a second stage, on the frozen output of the first. For each fold, the routing factor $f_s$ is trained without the held-out subject $s$; features are then assembled as 
\begin{equation} 
\phi = \Big[\, \ell_{\mathrm{route}},\; \{ \mathrm{MF}_{k,c} \}_{k,c},\; \{ \mathrm{ctr}_{k,(i,j)} \}_{k,\, i<j} \,\Big] \;\in\; \mathbb{R}^{\,1 + KC + K\binom{C}{2}}, 
\label{eq:features} 
\end{equation} 
and a logistic combiner is fit on the non-test subjects' features, computed through the same $f_s$, and applied to the test subject $s$: 
\begin{equation} 
\hat{p}_s = \sigma\big( \beta_s^{\top} \phi_s(f_s) + \beta_{0,s} \big), \qquad (\beta_s, \beta_{0,s}) = \arg\min\; \mathcal{L}_{\ell_2}\big( \{ \phi_o(f_s),\, y_o \}_{o \neq s} \big), 
\label{eq:fusion} 
\end{equation} 
with $\ell_2$ penalty (regularization strength 1.0), balanced class weighting, and L-BFGS optimization. The two factors are kept separate rather than trained end to end; this design choice is tested in Section~\ref{sec:validation}. 

Equation~\eqref{eq:features} has dimension $1 + KC + K\binom{C}{2}$: one routing logit plus an amplitude block that grows quadratically in the number of channels, from 24 amplitude features in the three-channel condition (25 total) to 8{,}320 at a 64-channel montage (8{,}321 total), both at $K{=}4$. The consequences of this scaling are taken up in Supplementary Section~\ref{sup:full}. In the representative three-channel configuration the model has 28{,}546 trainable parameters against up to 96{,}840 fixed, data-derived buffers (at $K{=}4$).

%% file: 03-methods-04-baselines.tex
\cbstart 
We compared ERP-XTTN against four baselines spanning the dominant approaches to cross-subject ERP classification: a compact convolutional network (EEGNet), a recent convolutional transformer (EEG-Deformer), a metric-based prototype method (EPMN), and a classical feature-engineering pipeline (xDAWN+RG). EEGNet and xDAWN+RG are long-standing benchmarks that have been evaluated directly against one another across ERP paradigms \cite{lawhern2018}; EEG-Deformer and EPMN situate ERP-XTTN against, respectively, a modern high-capacity architecture and the closest prototype-based comparator. 
\cbend

\subsubsection{EEGNet} 
As a deep learning baseline, we used EEGNet with $F1=8, D=2$, and a kernel length of 128 samples (half the 256 Hz sampling rate), yielding 2,017 trainable parameters for the 3-channel configuration. Dropout of 0.25 was applied in both blocks, with batch normalization after every convolution, average pooling (kernel sizes 4 and 8 following Blocks 1 and 2 respectively), and ELU activations, matching the cross-subject configuration of Lawhern et al. \cite{lawhern2018}. Max-norm weight constraints of 1.0 (depthwise convolution) and 0.25 (classifier) were enforced after each optimizer step, also per the original specification. 

\subsubsection{EEG-Deformer} 
\cbstart 
As a recent deep-learning baseline, we used EEG-Deformer \cite{ding2024}, a convolutional transformer whose hierarchical blocks combine a fine-grained temporal convolution branch with self-attention over coarse-grained tokens, and which densely propagates purified intermediate representations to the final embedding. We ported the reference implementation to our codebase \cite{wyman2026erpxttn_code} without architectural modification, retaining the published configuration. In the representative three-channel configuration this yields approximately $5.6 \times 10^{5}$ trainable parameters. 

\subsubsection{EPMN}
As a prototype-based baseline, we used EPMN \cite{wei2022}, a metric-learning method for zero-calibration ERP classification. EPMN maps each trial through a convolutional feature extractor to an $L_2$-normalized embedding and classifies by a softmax over squared Euclidean distance to per-class prototypes formed from the support set. The model is trained with the classification and metric losses as described by Wei et al. \cite{wei2022}. We did not identify an official implementation at the time of analysis, so we reimplemented the method from the original description; our implementation is available with the rest of the code \cite{wyman2026erpxttn_code}. Because EPMN's episodic meta-training differs from the protocol used for the other neural models, we evaluated it both under our shared LOSO two-phase protocol (Sections~\ref{sec:training} and~\ref{sec:evaluation}) and under its own native episodic recipe (Supplementary Section~\ref{sup:epmn-native}).
\cbend

\subsubsection{xDAWN+RG} 
As a classical baseline, we used xDAWN+RG with logistic regression, implemented via the pyriemann \cite{barachant2015} library. For each LOSO fold, xDAWN spatial filters \cite{rivet2009} (nfilter = 4; effectively capped at 3 per class in the 3-channel configuration) were estimated from the training pool to enhance evoked responses. The resulting trials were represented using xDAWN covariance matrices, projected into the Riemannian tangent space, and classified using logistic regression on the tangent-space features. Covariance matrices were estimated with Ledoit-Wolf shrinkage and projected to tangent space using the affine-invariant Riemannian metric. Logistic regression used $L_2$ regularization with $C=1.0$ and the L-BFGS solver. Unlike the neural models, xDAWN+RG applied no explicit class re-weighting and used no iterative epoch-based optimization, early stopping, or data augmentation; it was fit directly on the full training pool within each LOSO fold.

%% file: 03-methods-05-training.tex
EEGNet, 
\cbstart 
EEG-Deformer,
\cbend 
and ERP-XTTN were trained using AdamW (lr $= 1\times10^{-3}$, weight decay $= 1\times10^{-4}$), batch size 128, and class-weighted binary cross-entropy with logits loss, with the positive-class weight set to $n_{\text{negative}} / n_{\text{positive}}$. The learning rate schedule used linear warmup for 5 epochs followed by cosine annealing over 100 epochs, after which it was held at $1\times10^{-5}$; gradient norms were clipped at 1.0. Data augmentation, applied on-the-fly during training only, consisted of temporal jitter (uniform random shift in $[-10,~+10]$ samples with zero-padding at exposed boundaries) and additive Gaussian noise ($\sigma=0.1$) in normalized units. Early stopping used a trial-level 15\% stratified validation split from the training pool, stratified jointly by training subject and class; the held-out test subject contributed no trials to either split. The selection criterion was area under the receiver operating characteristic curve (AUROC; patience 15, maximum 250 epochs). The final model was then retrained from scratch on all training and validation data for that epoch count; 
\cbstart 
the validation split was used solely for epoch selection.  

EPMN was trained under the same optimizer, schedule, augmentation, and two-phase early-stopping procedure, with two differences intrinsic to its metric-learning design: optimization was episodic (each training subject serves in turn as query domain against the remaining subjects' support set), and epoch selection uses a subject-level rather than trial-level validation split. The held-out test subject was not included in the support set or the validation split.  
\cbend

No hyperparameter optimization was performed; 
\cbstart 
all settings were fixed across datasets and channel configurations for each model.
\cbend
Neural model experiments were conducted in PyTorch on an NVIDIA RTX PRO 6000 Blackwell Server Edition GPU using cloud compute; xDAWN+RG used the pyriemann library on CPU. 

%% file: 03-methods-06-evaluation.tex
All models were evaluated using LOSO cross-validation \cite{kunjan2021}: 
\cbstart 
each
subject served as the test set exactly once, with all remaining subjects forming the training set. No data from the held-out subject were used during training or combiner fitting.
\cbend
All models received identical preprocessing, channel selections, and train/test splits for each dataset, ensuring that performance differences reflect only the classification method. 
\cbstart 
Within each fold, channels were z-scored from training pool statistics;
\cbend
held-out subjects were normalized with those same statistics. The primary evaluation metric was AUROC, which is threshold-independent and robust to class imbalance. Balanced accuracy was computed at a fixed decision threshold of 0.5 on the predicted positive-class probability, applied uniformly across all subjects and configurations. This uncalibrated threshold simulates deployment conditions where per-subject optimization is unavailable. 

\cbstart 
All comparisons used mean-of-fold AUROC: per-subject AUROC was computed within each LOSO fold, averaged across the five seeds within subject, and then across subjects. Per-subject AUROCs were never pooled across folds. To compare ERP-XTTN against each baseline, we used the Wilcoxon signed-rank test on the per-subject paired differences, with the Hodges-Lehmann estimator and its exact distribution-free 95\% confidence interval as the effect-size measure, and controlled the false discovery rate across all 36 tests (nine datasets $\times$ four baselines) using the Benjamini-Hochberg procedure. The performance gap $\Delta$ (best baseline minus ERP-XTTN AUROC) is not subjected to a significance test, since the best baseline is selected post hoc. 
\cbend

%% file: 03-methods-07-architecturevalidation.tex
Two questions arise from the design of Section~\ref{sec:architecture}: whether the decisions genuinely depend on prototype content, and whether the specific architectural and prototype-construction choices are justified. We address the first with interventions on the trained model that test whether prototype content is necessary to the decision, and the second with single-knob ablations that test whether each design choice earns its accuracy. The results of both are reported in Section~\ref{sec:validation}. 

Because the match $m[p,k]$ is a fixed function of the prototype with no learned parameters (Section~\ref{sec:match}), a prototype can be corrupted at inference on the frozen model and the decision re-evaluated without retraining. Corruptions alter only the match pathway; the routing key and attention weights remain fixed. This dissociation cannot occur during normal operation, where the routing key and the match template derive from the same prototype; its purpose is to isolate the contribution of prototype content from that of routing-pattern geometry. 

For the routing factor, we applied a graded ladder of prototype corruptions, culminating in polarity inversion. The first rung replaces each prototype's match template with variance-matched Gaussian noise; because the routing keys and attention weights are left intact, the AUROC attained in this condition captures whatever discriminability remains without prototype content, and it therefore replaces a nominal 0.5 as the reference level for the remaining rungs. We refer to it as the noise null. Because the match is a signed cosine, negating the prototype should not merely remove discriminative information but reverse it, driving AUROC below the noise null for a grounded model. Permutation controls shuffled the trial-to-prototype pairing to test whether the specific correspondence carries the signal. The amplitude factor was tested by displacing the matched-filter window to off-target latencies and by permuting matched-filter feature vectors across trials, thereby breaking their correspondence with the labels. Single-knob ablations varied the two-factor design (routing-only, amplitude-only, end-to-end training), the whitening, and architectural hyperparameters (prototype-count cap, peak-detection prominence, attention heads, self-attention layer presence, unconstrained learned readout) on four datasets spanning the difficulty range (ERN, HRI ErrP, P300, N400). Datasets on which the factor under test performs at or near chance are referred to below as floor datasets; effects of an intervention or ablation cannot be resolved against a floor, so those datasets are excluded from the corresponding comparisons. Full procedural specifications are given in Supplementary Section~\ref{sup:validation}. 

%% file: 04-results.tex
\subsection{Classification Performance}
\label{sec:classification}
\input{04-results-01-classification}

\subsection{Physiological Structure of Model Decisions}
\label{sec:physiological-structure}
\input{04-results-02-structure}

\subsection{Architecture Validation and Ablations}
\label{sec:validation}
\input{04-results-03-validation}

%% file: 04-results-01-classification.tex
\begin{table}
\centering
\small
\setlength{\tabcolsep}{5pt}
\caption{Classification performance (area under the receiver operating characteristic curve, AUROC; mean $\pm$ standard deviation across subjects) at the three-channel montage under leave-one-subject-out (LOSO) cross-validation. Neural models (ERP-XTTN, EEGNet, EEG-Deformer, EPMN) report seed-averaged means across five training seeds; xDAWN+RG is deterministic (single run). \textbf{Bold} marks the highest mean per dataset. $\Delta$ = best baseline minus ERP-XTTN (performance gap, in AUROC); the best baseline is selected post hoc per dataset and $\Delta$ is reported descriptively. Datasets ordered by mean AUROC across all five methods, descending. Full-montage results are in Supplementary Table~\ref{sup:tab-fullmontage-auroc}; three-channel per-subject AUROC values in Supplementary Tables~\ref{sup:tab-persubj-hri}--\ref{sup:tab-persubj-n400}. EEGNet: compact convolutional neural network \cite{lawhern2018}; EEG-Deformer: dense convolutional transformer \cite{ding2024}; EPMN: ERP Prototypical Matching Net \cite{wei2022}; xDAWN+RG: xDAWN spatial filtering with Riemannian geometry classification. Dataset abbreviations are defined in Table~\ref{tab:datasets}.}
\label{tab:auroc}
\begin{tabular}{lcccccc}
\toprule
Dataset & ERP-XTTN & EEGNet & EEG-Deformer & EPMN & xDAWN+RG & $\Delta$ \\
\midrule
HRI & 0.837 $\pm$ 0.101 & \textbf{0.858 $\pm$ 0.101} & 0.852 $\pm$ 0.097 & 0.831 $\pm$ 0.100 & 0.837 $\pm$ 0.107 & \textit{0.021} \\
ERN & 0.824 $\pm$ 0.110 & \textbf{0.869 $\pm$ 0.093} & 0.840 $\pm$ 0.097 & 0.795 $\pm$ 0.096 & 0.793 $\pm$ 0.130 & \textit{0.045} \\
LRP & 0.761 $\pm$ 0.080 & 0.792 $\pm$ 0.084 & 0.786 $\pm$ 0.081 & 0.769 $\pm$ 0.080 & \textbf{0.796 $\pm$ 0.074} & \textit{0.035} \\
BNCI & 0.771 $\pm$ 0.084 & \textbf{0.792 $\pm$ 0.088} & 0.787 $\pm$ 0.084 & 0.727 $\pm$ 0.102 & 0.785 $\pm$ 0.082 & \textit{0.021} \\
N170 & 0.737 $\pm$ 0.105 & 0.732 $\pm$ 0.105 & 0.730 $\pm$ 0.106 & 0.712 $\pm$ 0.104 & \textbf{0.748 $\pm$ 0.104} & \textit{0.011} \\
P300 & 0.664 $\pm$ 0.071 & \textbf{0.723 $\pm$ 0.087} & 0.717 $\pm$ 0.082 & 0.674 $\pm$ 0.077 & 0.660 $\pm$ 0.091 & \textit{0.059} \\
N2pc & 0.639 $\pm$ 0.081 & 0.648 $\pm$ 0.092 & 0.641 $\pm$ 0.092 & 0.634 $\pm$ 0.082 & \textbf{0.649 $\pm$ 0.093} & \textit{0.010} \\
MMN & 0.587 $\pm$ 0.049 & 0.589 $\pm$ 0.049 & 0.587 $\pm$ 0.050 & 0.562 $\pm$ 0.038 & \textbf{0.593 $\pm$ 0.048} & \textit{0.006} \\
N400 & 0.565 $\pm$ 0.077 & \textbf{0.586 $\pm$ 0.087} & 0.578 $\pm$ 0.079 & 0.573 $\pm$ 0.077 & 0.580 $\pm$ 0.081 & \textit{0.021} \\
\midrule
\textit{Mean} & 0.709 & \textbf{0.732} & 0.724 & 0.697 & 0.716 & \textbf{0.025} \\
\bottomrule
\end{tabular}
\end{table}

\begin{figure}
\centering
\includegraphics[width=0.9\textwidth]{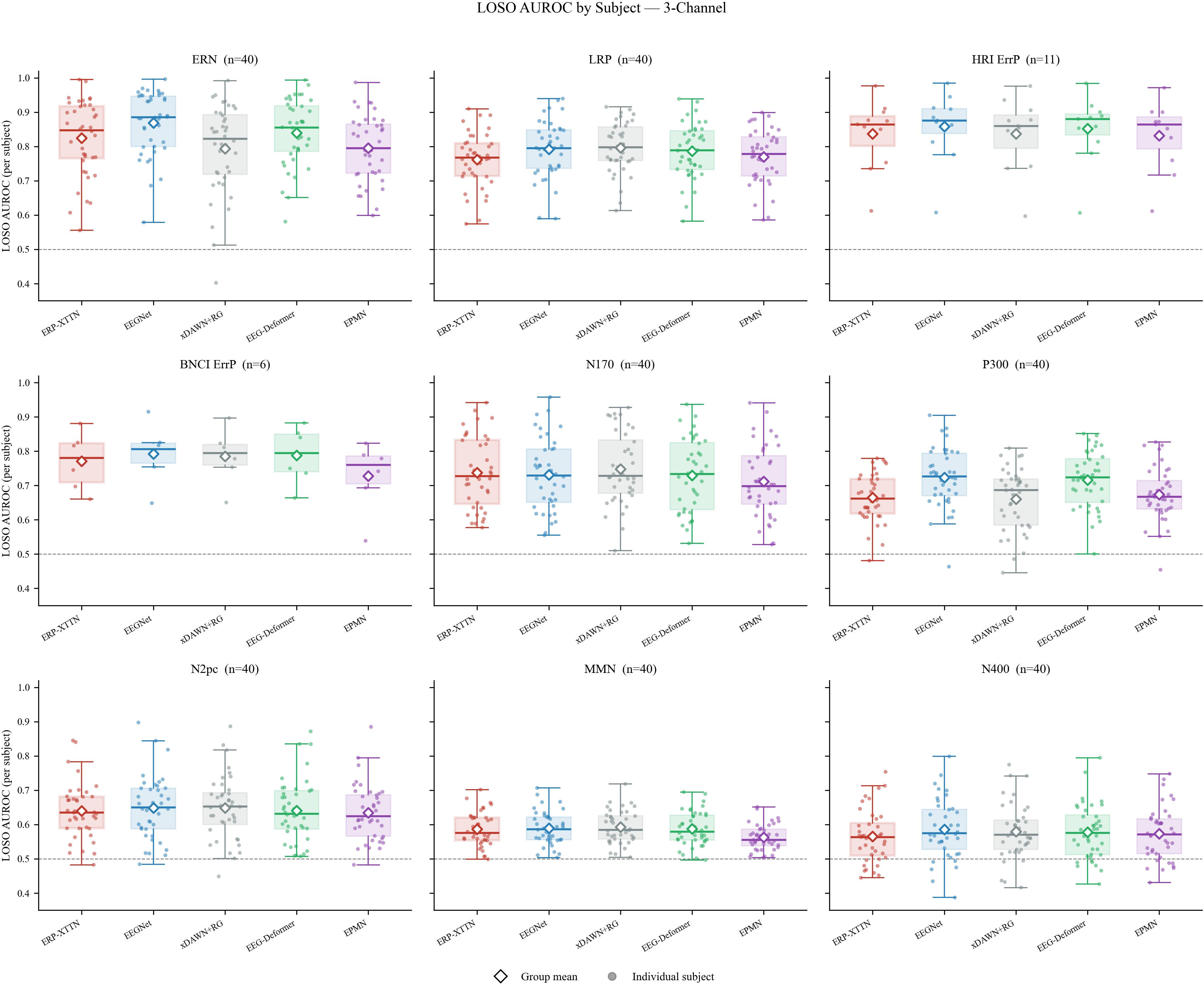}
\caption{Per-subject area under the receiver operating characteristic curve (AUROC) distributions at the three-channel montage for all nine datasets under leave-one-subject-out (LOSO) cross-validation. Each point is one held-out subject's AUROC averaged across five training seeds (xDAWN+RG: single deterministic run). Boxes show the interquartile range with median line; diamonds show the group mean. Dashed line at 0.5 indicates chance. Datasets ordered by mean AUROC across all five methods, descending (Table~\ref{tab:auroc}); sample sizes are indicated per panel. EEGNet: compact convolutional neural network \cite{lawhern2018}; EEG-Deformer: dense convolutional transformer \cite{ding2024}; EPMN: ERP Prototypical Matching Net \cite{wei2022}; xDAWN+RG: xDAWN spatial filtering with Riemannian geometry classification. Dataset abbreviations are defined in Table~\ref{tab:datasets}.}
\label{fig:persubject}
\end{figure}

Table~\ref{tab:auroc} reports mean LOSO AUROC for all five methods across the nine datasets at the three-channel montage, and Figure~\ref{fig:persubject} shows the corresponding per-subject distributions. Cross-subject difficulty followed a highly consistent ordering across all five methods (mean pairwise Spearman $\rho=0.96$), from HRI and ERN (easiest) through to MMN/N400 (hardest), with the few differences being swaps between near-tied datasets. Per-subject AUROCs were averaged over five training seeds (xDAWN+RG is deterministic, single run). 
Across-seed standard deviations of the dataset-level means were $\leq 0.008$ for ERP-XTTN, EEGNet, and EEG-Deformer, and $\leq 0.020$ for EPMN, small relative to the cross-subject standard deviations (0.05–0.11) reported in Table~\ref{tab:auroc}. Results are reported as seed-averaged means. 

EEGNet was the top baseline on five datasets (BNCI, HRI, ERN, P300, N400) and xDAWN+RG on four (LRP, MMN, N2pc, N170). Across datasets, the descriptive performance gap between the best baseline and ERP-XTTN ($\Delta$, in AUROC; Table~\ref{tab:auroc}) averaged 0.025 (range 0.006–0.059). Per-subject AUROC values are reported in Supplementary Tables~\ref{sup:tab-persubj-hri}–\ref{sup:tab-persubj-n400}. 

\begin{figure}
\centering
\includegraphics[width=1\textwidth]{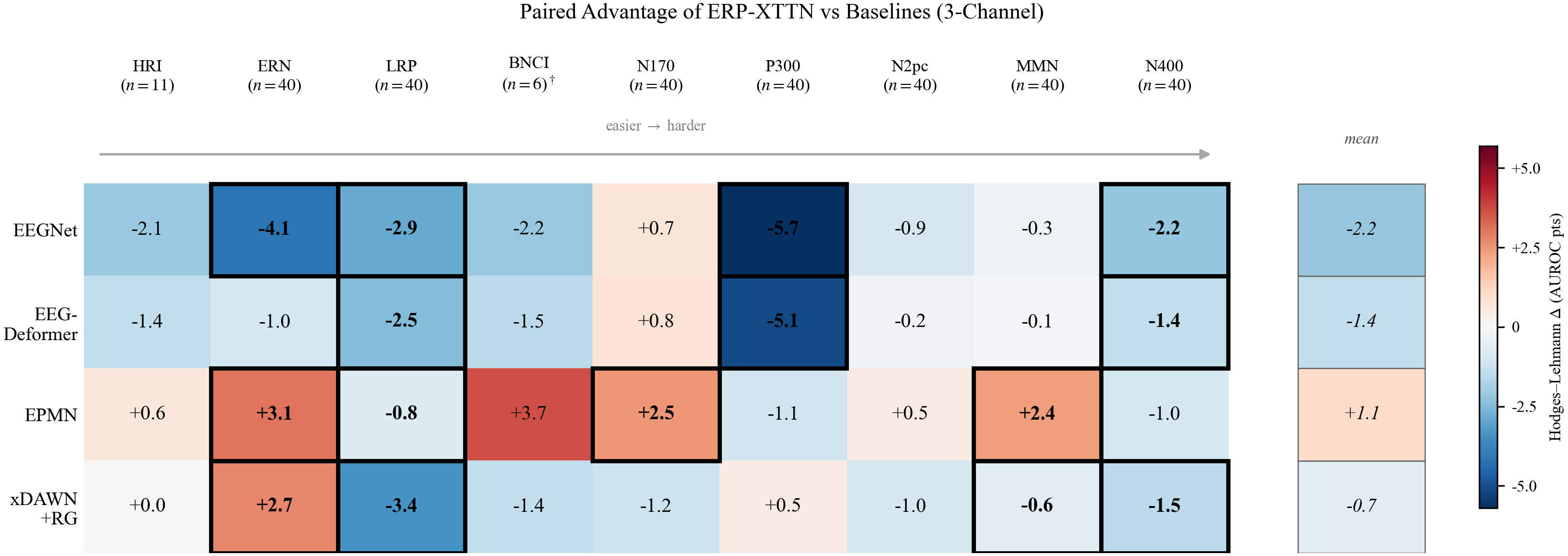}
\caption{Paired per-subject comparison of ERP-XTTN against each baseline at the three-channel montage. Each cell shows the Hodges-Lehmann estimate of the paired difference (ERP-XTTN minus baseline) in area under the receiver operating characteristic curve (AUROC) points, computed from per-subject AUROC values averaged across five training seeds (xDAWN+RG: single deterministic run). Red indicates ERP-XTTN higher; blue indicates baseline higher. Bold text with black border: significant after false-discovery-rate correction across all 36 tests (nine datasets $\times$ four baselines; two-sided Wilcoxon signed-rank test, Benjamini-Hochberg $q < 0.05$). Full test statistics are reported in Supplementary Table~\ref{sup:tab-wilcoxon}. Datasets ordered by mean AUROC across all five methods, descending (Table~\ref{tab:auroc}). $\dagger$\,BNCI ($n = 6$): the exact signed-rank $p$ floors at 0.031 and cannot survive correction at this sample size; the estimate is reported descriptively. Right column: mean across datasets. EEGNet: compact convolutional neural network \cite{lawhern2018}; EEG-Deformer: dense convolutional transformer \cite{ding2024}; EPMN: ERP Prototypical Matching Net \cite{wei2022}; xDAWN+RG: xDAWN spatial filtering with Riemannian geometry classification. Dataset abbreviations are defined in Table~\ref{tab:datasets}.}
\label{fig:fig_paired_heatmap}
\end{figure}

In paired per-subject comparisons (Figure~\ref{fig:fig_paired_heatmap}), no significant difference between ERP-XTTN and EEGNet was detected on HRI, BNCI, N170, N2pc, or MMN after false-discovery-rate correction. Hodges-Lehmann estimates indicated significantly lower AUROC for ERP-XTTN compared with EEGNet on P300 (-5.7 points), ERN (-4.1 points), LRP (-2.9 points), and N400 (-2.2 points). Against xDAWN+RG the comparison was mixed: ERP-XTTN significantly exceeded it on ERN and fell below it on LRP, MMN, and N400. Against EPMN, the closest prototype-based baseline, ERP-XTTN was higher on six of nine datasets, significantly so on three (ERN, N170, MMN), and significantly lower only on LRP. EEG-Deformer separated from ERP-XTTN on only three paradigms (P300, LRP, N400). Full test statistics are reported in Supplementary Table~\ref{sup:tab-wilcoxon}.

Median single-trial latencies on a consumer CPU (Apple M2, 24 GB RAM, single-threaded, no GPU) ranged from $0.18$~ms (EEGNet) to $2.82$~ms (ERP-XTTN), all well within real-time budgets; ERP-XTTN's higher latency reflects its unoptimized peak-detection tokenizer. 

All five methods, ERP-XTTN included, improved under the full montage (Supplementary Table~\ref{sup:tab-fullmontage-auroc}). Because ERP-XTTN's amplitude features scale with channel count (over 8{,}000 at 64 channels), we take the three-channel montage as the interpretable operating point; the full-montage variant is reported for completeness and its interpretability examined in Section~\ref{sup:full}.

%% file: 04-results-02-structure.tex
\begin{figure}
\centering
\includegraphics[width=0.9\textwidth]{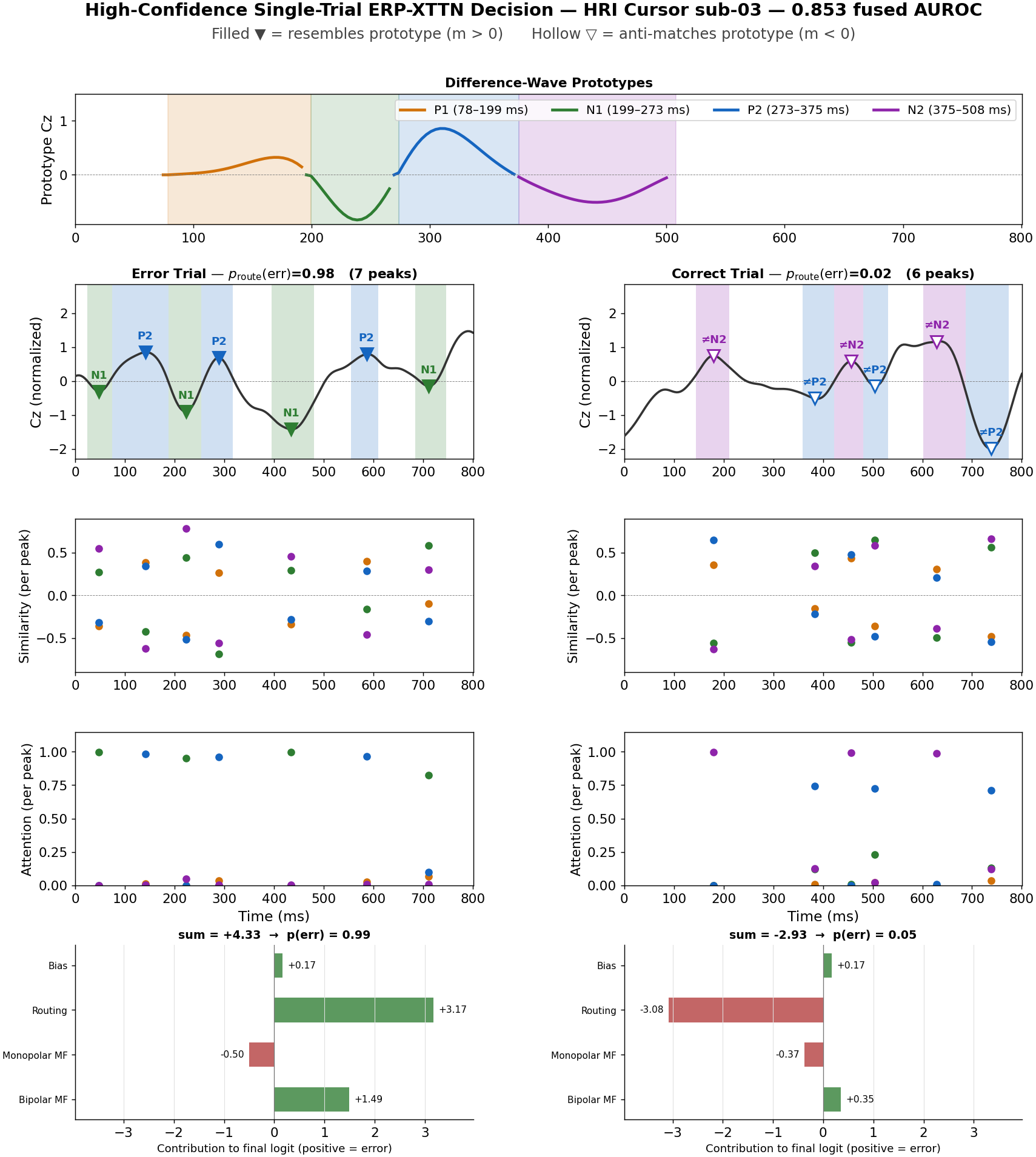}
\caption{Single-trial decision decomposition for HRI ErrP sub-03 from one representative training seed (seed 1; ERP-XTTN two-factor (fused) area under the receiver operating characteristic curve (AUROC) $= 0.853$) in the 3-channel configuration. Top: difference-wave prototypes at detection-channel Cz with shaded component windows (color-coded). Second row: detection-channel waveforms for a high-confidence true positive (error trial, left; $p_{\mathrm{route}}(\text{err}) = 0.98$) and true negative (correct trial, right; $p_{\mathrm{route}}(\text{err}) = 0.02$), selected by routing-factor-only confidence. Shaded bands span each detected peak's window and are colored, like the markers at each peak's center, by the prototype receiving that peak's largest attention-weighted similarity $|a[p,k]\, m[p,k]|$. Filled markers indicate positive similarity ($m > 0$), hollow markers indicate negative similarity ($m < 0$). Third row: per-peak similarity $m[p,k]$ to each prototype, color-coded by prototype. Fourth row: per-peak cross-attention weight $a[p,k]$, the learned routing score indicating how strongly peak $p$ attends to prototype $k$. The pre-combiner routing logit is the learned-scale and bias transformation of the valid-peak-normalized, attention-weighted similarity sum (Section~\ref{sec:routing-readout}). Bottom row: exact additive contributions to the final logistic-combiner logit from routing, monopolar matched filtering (Monopolar MF), bipolar matched filtering (Bipolar MF), and the combiner bias. Values label the signed contribution of each term; panel titles give their sum and the resulting fused probability (error trial, left, $p(\text{err}) = 0.99$; correct trial, right, $p(\text{err}) = 0.05$). All quantities are computed directly from the trained routing model and fitted combiner, without post-hoc approximation. Prototypes are color-coded consistently across all rows. Dataset abbreviations are defined in Table~\ref{tab:datasets}.}
\label{fig:single-trial}
\end{figure}

\begin{figure}
\centering
\includegraphics[width=0.75\textwidth]{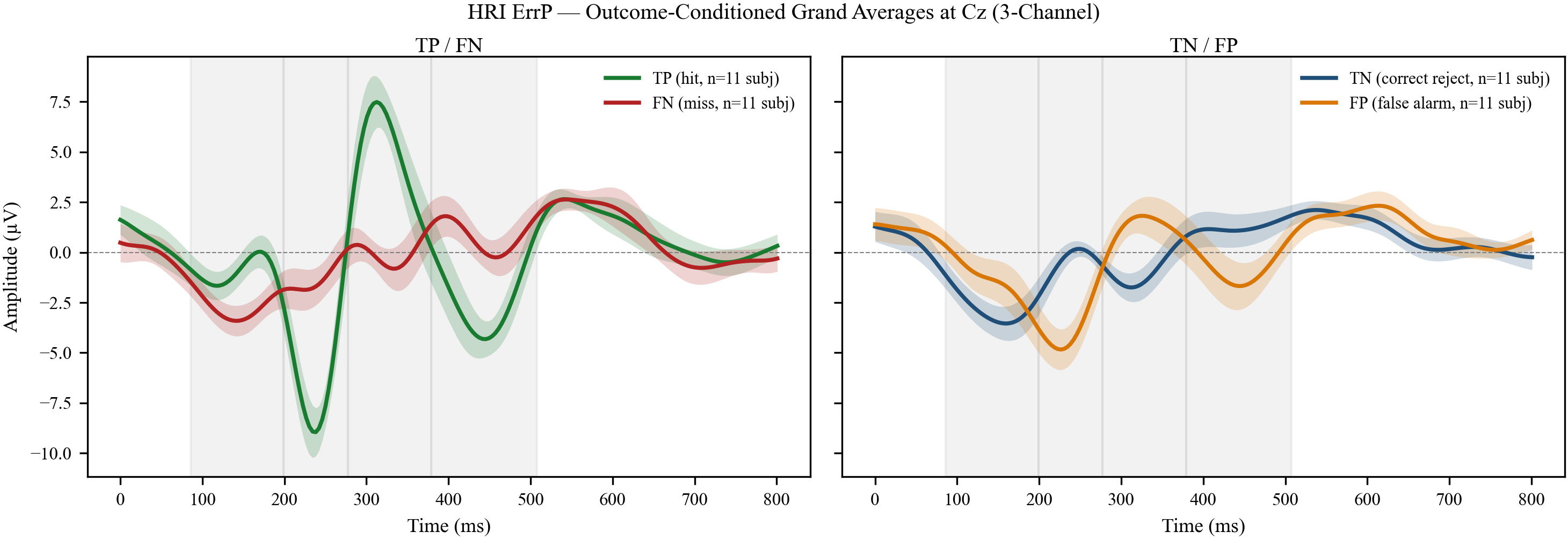}
\caption{Outcome-conditioned grand-average waveforms at Cz for HRI ErrP (3-channel configuration, $n = 11$ subjects). Left: true positive (TP, correctly classified error trials) and false negative (FN, missed error trials) waveforms with standard-error ribbons. Right: true negative (TN, correctly classified correct trials) and false positive (FP, falsely flagged correct trials) waveforms with standard-error ribbons. Shaded grey regions indicate detected prototype windows. Trial counts are summed across all subjects and folds. Waveforms are averaged across subjects, which suppresses subject-idiosyncratic morphology; the per-subject correlations quantifying this comparison are reported in Supplementary Table~\ref{sup:tab-corr}. False positives morphologically resemble true positives across the full ErrP complex but at reduced amplitude, while true negatives show substantially less morphological similarity to true positives, consistent with classification operating through waveform-prototype correspondence. Outcome-conditioned waveforms for all nine datasets are shown in Supplementary Figure~\ref{sup:fig-morphology-all}. Dataset abbreviations are defined in Table~\ref{tab:datasets}.}
\label{fig:morphology}
\end{figure}

We examined ERP-XTTN's decision structure at two complementary levels: single-trial decision decomposition and outcome-conditioned waveform morphology. Figure~\ref{fig:single-trial} shows this decomposition for one high-confidence error trial and one high-confidence correct trial from a representative HRI subject, reporting each detected peak's signed similarity $m$ to each prototype, the attention weight $a$ it receives, and the matched-filter contributions. Single-trial decompositions for all subjects are available in the public code repository \cite{wyman2026erpxttn_code}. 

We next examined whether ERP-XTTN's individual classification outcomes could be explained by trial-level waveform morphology by comparing grand-average waveforms conditioned on model predictions. On HRI ErrP at Cz (Figure~\ref{fig:morphology}), true positives showed larger and sharper ERP components than false negatives, which were attenuated. False positives closely resembled true positives in waveform shape across all four prototype windows (P1, N1, P2, and N2) but at reduced amplitude, whereas true negatives showed substantially less morphological similarity to true positives. Outcome-conditioned waveforms for all nine datasets are shown in Supplementary Figure~\ref{sup:fig-morphology-all}. 

We quantified this relationship by computing, for each subject, the Pearson correlation between that subject's mean true-positive and mean false-positive waveforms (TP$\leftrightarrow$FP) and between their mean true-positive and mean true-negative waveforms (TP$\leftrightarrow$TN) at the detection channel. Per-subject correlations were summarized as the mean across subjects for a single reference seed. TP$\leftrightarrow$FP correlations were positive and substantial across datasets (r = 0.60–0.86; Table~\ref{sup:tab-corr}). TP$\leftrightarrow$TN correlations were generally lower but remained positive, with one exception: on MMN, TP$\leftrightarrow$TN was negative (r = -0.31). Although bootstrap confidence intervals overlapped on several datasets, the ordering TP$\leftrightarrow$FP $>$ TP$\leftrightarrow$TN held across all nine datasets and all five seeds (Table~\ref{sup:tab-corr-perseed}). 

%% file: 04-results-03-validation.tex
Results of the grounding interventions and ablations described in Section~\ref{sec:architecture-validation} are reported below; full three-channel results are in Supplementary Tables~\ref{sup:tab-grounding-3-routing}–\ref{sup:tab-ablation}. ERP-XTTN's routing factor responded systematically to corruption of the physiological prototypes (Supplementary Section~\ref{sup:validation}). Inverting each prototype’s polarity drove routing below the noise null on all nine datasets, with the largest reversals occurring where routing-only performance was strongest (e.g. HRI, $0.81 \to 0.19$, against a null near $0.50$). The reversals were smaller on MMN and N400, where routing-only performance was closer to chance. Both trial-to-prototype permutations returned routing to the noise null throughout. The same polarity inversion applied to an unconstrained learned readout over the same attention tensor had no effect (HRI $0.82 \to 0.82$), confirming that the learned head classifies from routing-pattern geometry rather than prototype content (Table~\ref{sup:tab-polarity-contrast}).

The early-window control reduced amplitude-factor discriminability on every dataset where the amplitude factor carried signal except ERN (e.g. HRI $0.74 \rightarrow 0.58$); ERN's routing factor showed the full corruption response, so the grounding evidence for that dataset rests on routing. Reductions under the late-window displacement were smaller and less consistent. Permuting matched-filter features across trials reduced discriminability to 0.54–0.60, well below the on-target values; the floor lies above 0.5 because the diagnostic reports the best of many individual features (Supplementary Section~\ref{sup:grounding-interventions}).

The ablation grid was run on four datasets spanning the difficulty range; on N400 routing was near chance, so it is treated as a floor dataset. On the three remaining datasets, the two-factor model exceeded either factor alone (Table~\ref{sup:tab-ablation}). On N400, amplitude alone performed comparably to and numerically above the fused model (0.578 versus 0.565), and routing added no measurable benefit. The two factors were strongly correlated at the trial level (on HRI, $r = 0.69$ overall, and $0.48$ and $0.66$ within the correct and error classes). Training the two factors end to end, rather than fitting the combiner on the frozen routing model, reduced accuracy on every non-floor ablation dataset. The learned readout gave no accuracy advantage over the grounded model on ERN and HRI and performed worse on the harder datasets, falling to chance on N400. The model was otherwise insensitive to its architectural choices: across the four ablation datasets, varying the attention-head count and peak-detection prominence, replacing the whitened match with a raw cosine match, and removing the self-attention layer each changed AUROC by at most $0.017$ (Supplementary Table~\ref{sup:tab-ablation}).

At the full montage, the prototype-corruption ladder still drove routing below the noise null on every dataset where routing carried signal, but the amplitude-localization controls became uninformative (Supplementary Section~\ref{sup:full}).

%% file: 05-discussion.tex
\cbstart 
ERP-XTTN achieves competitive cross-subject classification across diverse ERP paradigms while providing physiologically grounded, directly inspectable decisions. We discuss the evidence supporting this interpretability claim, its performance tradeoffs, the contribution of the LOSO benchmark, and the study’s limitations. 

\subsection{Physiologically Interpretable Decisions}
\label{sec:interpretability}
\input{05-discussion-01-interpretability}

\subsection{Performance Tradeoffs}
\label{sec:performance}
\input{05-discussion-02-performance}
\cbend

\subsection{LOSO Benchmark}
\label{sec:loso}
\input{05-discussion-03-loso}

\subsection{Limitations}
\label{sec:limitations}
\input{05-discussion-04-limitations}

%% file: 05-discussion-01-interpretability.tex
The central interpretability claim of ERP-XTTN is not that attention weights are inherently explanatory, but that the physiological content of the prototypes is necessary to the decision. The grounded readout enforces this by construction, and the intervention analyses of Section~\ref{sec:validation} confirm it empirically. Where the routing factor carried discriminative signal, prototype polarity inversion drove its AUROC below the noise null rather than merely toward chance, and both trial-to-prototype permutations eliminated the signal. By contrast, an unconstrained learned readout over the same attention tensor was unchanged by polarity inversion (Table~\ref{sup:tab-polarity-contrast}) and provided no accuracy advantage. The amplitude controls likewise showed that the factor's discriminative value depended on the expected component window: displacing the matched filter into the pre-component period reduced discriminability on every dataset where the amplitude factor carried signal, with the exception of ERN, where the signal persisted into that window. The grounding evidence for ERN therefore rests on its routing factor, which showed the full corruption response. Reductions under the weaker late-window displacement, which can overlap the original window, were less consistent. These results indicate that ERP-XTTN’s decisions depend on the physiological content and trial-to-prototype correspondence of the evidence, not only on the geometry of the learned attention pattern.   

Unlike post-hoc explanation methods applied to black-box classifiers, ERP-XTTN's displayed quantities (per-peak similarities, attention weights, and matched-filter projections) are the quantities through which the prediction was actually produced. The TP$\leftrightarrow$FP analysis further supports this interpretation: across all nine datasets, false positives morphologically resembled true positives more than true negatives did, consistent with classification operating through waveform-prototype correspondence rather than through features opaque to the observer. Classification errors are therefore neurophysiologically explicable. MMN provides an informative special case: TP$\leftrightarrow$TN was negative, reflecting the opposing waveforms elicited by standard and deviant tones, while TP$\leftrightarrow$FP remained higher, a pattern that itself is neurophysiologically expected.  

%% file: 05-discussion-02-performance.tex
ERP-XTTN was competitive across diverse ERP paradigms but not the highest-performing method on any dataset. The descriptive performance gap from the best baseline averaged 0.025 AUROC across datasets, with the largest differences on P300 and ERN and near-matched performance on N170, N2pc, MMN, HRI, and BNCI. Against EPMN, the closest prototype-based comparator, ERP-XTTN was higher on six of nine datasets; against EEG-Deformer, a substantially larger convolutional transformer, it differed significantly on only three paradigms. These comparisons suggest that making the decision pathway physiologically grounded does not preclude competitive cross-subject performance, although unconstrained architectures retain an advantage on some datasets. 

In a controlled ablation holding the peak tokenizer and attention tensor fixed, replacing the grounded match with an unconstrained learned readout provided no accuracy advantage, while training the routing and amplitude pathways end to end reduced performance on every non-floor ablation dataset. The two-stage fusion therefore improves accuracy, and the grounded readout costs none.

The three-channel montage is the clearest operating point for ERP-XTTN. Although all five methods improved with the full montage, ERP-XTTN’s amplitude feature space grows quadratically with channel count, reaching over 8{,}000 features for high-density recordings. At that dimensionality, the routing pathway remained sensitive to prototype corruption, but the amplitude-localization controls became uninformative, weakening the groundedness claim for the full model. The full-montage results demonstrate that the architecture can exploit additional spatial information, but the minimal montage matches the deployment constraint the work targets and provides the strongest balance between classification performance, computational simplicity, and physiologically interpretable decision evidence. 

%% file: 05-discussion-03-loso.tex
To the best of our knowledge, these are the first published epoch-level LOSO results on ERP CORE under deployment-compatible (causal) preprocessing constraints. The ordering of classification difficulty across ERP components, with 
\cbstart 
HRI and ERN easiest and MMN and N400 hardest, was highly consistent across all five methods (mean pairwise Spearman $\rho=0.96$; Tables~\ref{tab:auroc} and~\ref{sup:tab-fullmontage-auroc}). This consistency across convolutional and transformer architectures, metric-based prototype learning, and classical spatial filtering, and its agreement with prior cross-subject results on ERP CORE \cite{aristimunha2023}, suggests that the ordering primarily reflects ERP signal and paradigm 
\cbend
properties rather than classifier-specific behavior. The cross-component difficulty ordering is useful independently of ERP-XTTN: it characterizes which components generalize cross-subject under matched preprocessing and offers practitioners a guide to which paradigms are viable for calibration-free deployment. 

On the two previously benchmarked ErrP datasets (HRI and BNCI), ERP-XTTN's LOSO performance is competitive with published results. On HRI, Schönleitner et al. \cite{schonleitner2020} reported 72.7\% balanced accuracy with generalized LDA using 27 channels; ERP-XTTN achieved 
\cbstart 
76.6\%
\cbend
with three midline channels. On BNCI, Ren et al. \cite{ren2024} reported a best per-session LOSO AUROC of 0.755 with two channels (FCz, Cz); ERP-XTTN achieved 
\cbstart 
0.771 with three channels under causal filtering. 

All results reported here use causal IIR filtering under calibration-free LOSO evaluation; many published LOSO benchmarks use acausal filtering instead \cite{schonleitner2020, ren2023}, so their reported performance is not reproducible in a forward-processing pipeline. 
\cbend

%% file: 05-discussion-04-limitations.tex
\cbstart 
Although the decision pathway is grounded within a trained model, the particular routing pattern is not identifiable across independent retrainings. Different random initializations can converge to different routing patterns while achieving similar classification performance. Consequently, routing should be interpreted as an explanation of a specific trained model rather than as a unique decomposition of the classification task. The TP$\leftrightarrow$FP morphological ordering (Section~\ref{sec:physiological-structure}) is unaffected by this limitation: it reflects the fused decision's prediction-conditioned waveforms rather than any particular routing solution, and was stable across all five seeds (Table~\ref{sup:tab-corr-perseed}).
\cbend

Several methodological limitations should be noted. First, all evaluations were offline; online or real-time deployment may introduce additional challenges not captured here. Second, prototypes were frozen after extraction from the training set and may not capture individual variation in ERP morphology; adaptive prototypes that update during inference could address this but would complicate the 
\cbstart 
groundedness guarantee and intervention analyses. 
\cbend
Third, the ERP CORE datasets each contain 40 subjects, moderate for LOSO evaluation but insufficient to characterize performance on clinical populations or underrepresented demographics. Fourth, the LRP epoch window (0–800 ms post-response) captures post-response lateralized activity rather than the canonical pre-response lateralized readiness potential. Fifth, training hyperparameters including early-stopping patience and learning rate were shared across all neural models without per-model optimization; the reported gap may not reflect the full capacity of each architecture. Finally, balanced accuracy lagged AUROC across all five methods (Table~\ref{sup:tab-balacc}), indicating that the fixed, uncalibrated 0.5 decision threshold is suboptimal under LOSO. Post-hoc probability calibration or threshold calibration could address this without altering any of the architectures.

%% file: 06-conclusion.tex
ERP-XTTN generalizes from ErrP to seven additional ERP components via automatic prototype detection, achieving competitive cross-subject classification under causal, calibration-free LOSO conditions with a mean 
\cbstart 
performance gap of 0.025 AUROC at 3 channels. Interventions on the frozen model verified that decisions depend on physiological prototype content rather than on the attention pattern alone. The architecture’s grounded decision pathway makes individual predictions directly inspectable and yields morphologically explicable classification errors. 
\cbend
To our knowledge, this work also provides the first epoch-level LOSO benchmark on ERP CORE.

%% file: supplementary-datasets.tex
This section provides the eliciting paradigm, class definitions, and per-class trial counts for each dataset summarized in Section~\ref{sec:datasets} and Table~\ref{tab:datasets} of the main text. 

\subsection{BNCI Horizon 2020 013-2015 (BNCI) - ErrP}
BNCI \footnote{BNCI Horizon 2020 013-2015 is publicly available at \url{https://bnci-horizon-2020.eu/database/data-sets}} is a 64-channel EEG dataset designed to elicit ErrP \cite{chavarriaga2010}. Six subjects were asked to monitor the performance of an agent moving a cursor toward a target; errors occurred on roughly 20\% of trials. Data were collected over two sessions, each averaging 110 error trials and 426 correct trials for a total of roughly 536 trials per session. 

\subsection{HRI Cursor (HRI) - ErrP} 
HRI \footnote{HRI Cursor is publicly available at \url{https://github.com/stefan-ehrlich/dataset-ErrP-HRI}} is an EEG dataset recorded with 32 active electrodes (27 scalp EEG, all analyzed here, plus 3 electrooculography and 2 mastoid reference) designed to elicit ErrP during a choice-reaction-time task with cursor-movement feedback \cite{ehrlich2019}. Eleven subjects responded to one of three target stimuli via keypress, and feedback was presented as a cursor moving toward or away from the target; errors occurred on roughly 35\% of trials. Only the cursor scenario was used in this work; a companion robot-head-turn scenario from the same study was excluded. Subjects had an average of 164 error trials and 319 correct trials for a total of roughly 483 trials. 

\subsection{ERP CORE} 
The remaining seven datasets are drawn from ERP CORE \footnote{ERP CORE is publicly available at \url{https://erpinfo.org/erp-core}} \cite{kappenman2021}, a publicly available resource providing standardized paradigms for 7 ERPs and data from 40 subjects across 30 EEG channels. Note that the original ERP CORE publication \cite{kappenman2021} reports smaller per-component samples (N = 34–39) after exclusions based on behavioral-accuracy and artifact-rejection criteria tied to their ERP-averaging pipeline. We retained all 40 participants because our preprocessing omits the baseline correction, re-referencing, and ocular-artifact correction those criteria presuppose, and because retaining all participants provides a more stringent test of cross-subject generalization.

\textbf{ERN:} The error-related negativity was elicited using an Eriksen flankers task in which subjects identified the direction of a central arrowhead flanked by congruent or incongruent distractors. Trials were response-locked and classified as incorrect vs correct responses. Subjects averaged approximately 45 incorrect and 356 correct responses. 

\textbf{LRP:} The lateralized readiness potential was elicited using the same flankers task, restricted to trials with correct responses and analyzed with respect to response hand. Trials were response-locked and classified as left-hand vs right-hand responses. Subjects averaged approximately 177 left-hand and 179 right-hand responses. 

\textbf{MMN:} The mismatch negativity was elicited using a passive auditory oddball task. Standard ($80~dB,~p=0.8$) and deviant ($70~dB,~p=0.2$) tones were presented while subjects watched a silent video. Trials were classified as deviant vs standard. Subjects averaged approximately 199 deviant and 782 standard trials. 

\textbf{N170:} The N170 was elicited using a face perception task in which subjects judged whether each stimulus was an intact object (faces or cars) or a texture (scrambled faces or scrambled cars). Only intact stimuli were used; trials were classified as face vs car. Subjects averaged 80 face and 80 car trials. 

\textbf{N2pc:} The N2pc is a lateralized ERP component associated with covert attentional selection. It was elicited using a visual search task in which subjects viewed arrays of colored squares in both hemifields. On each trial, they identified whether a small notch in the color-defined target square appeared on the top or bottom edge. Trials were classified as left-target vs right-target. Subjects averaged approximately 160 left-target and 160 right-target trials. 

\textbf{N400:} The N400 was elicited using a word pair judgment task. A red prime word was followed by a green target word, and subjects indicated whether the pair was semantically related or unrelated. Trials were classified as unrelated vs related. Subjects averaged 60 unrelated and 60 related trials. 

\textbf{P300:} The P300 was elicited using an active visual oddball task in which letters were presented equiprobably and one was designated the target per block. Trials were classified as target vs non-target. Subjects averaged 40 target and 160 non-target trials.

\bigskip
This study used only publicly available, de-identified datasets. No new human subjects data were collected. Ethical approval for the original data collection was obtained by the original investigators: the ERP CORE dataset was approved by the Institutional Review Board at the University of California, Davis \cite{kappenman2021}; the HRI ErrP dataset was approved by the ethics commission of the Faculty of Medicine, Technische Universität München (reference number 236/15s) \cite{ehrlich2019}. The original publication describing the BNCI dataset \cite{chavarriaga2010} does not include a formal ethics statement; the data were collected at the Idiap Research Institute, Martigny, Switzerland. 

%% file: supplementary-full.tex
The main text reports all classification and grounding results at the three-channel montage. This section reports the corresponding results at the full available montage for each dataset. 

All five methods improved under the full montage (Table~\ref{sup:tab-fullmontage-auroc}; balanced accuracy in Table~\ref{sup:tab-fullmontage-ba}). The mean performance gap between the best baseline and ERP-XTTN was 0.036 AUROC, compared with 0.025 at three channels. The cross-dataset difficulty ordering was largely preserved. ERP-XTTN's amplitude feature block in the combiner feature vector (Equation~7) grows quadratically with channel count, reaching 8,320 amplitude features at 64 channels ($K = 4$), compared with 24 at three channels. The per-feature interpretability that is straightforward at three channels becomes impractical at research-grade montages. 

The prototype-corruption ladder produced the same qualitative pattern as at three channels: polarity inversion drove routing below the noise null on every dataset where routing carried signal, and both permutations returned routing to the noise-null range (Table~\ref{sup:tab-grounding-full-routing}). The amplitude-localization controls, however, were less informative at the full montage (Table~\ref{sup:tab-grounding-full-amp}): displaced windows did not consistently reduce best-feature AUROC, consistent with reduced specificity as the number of candidate channel and contrast features increased. The three-channel montage remains the recommended operating point for applications requiring physiologically interpretable decisions.

%% file: supplementary-epmn.tex
To confirm the model ranking does not depend on our shared training protocol, we additionally evaluate EPMN under its native recipe described by Wei et al. \cite{wei2022}: the input time axis is resampled to 64 samples; optimization uses stochastic gradient descent (lr $10^{-4}$, momentum 0.9) with the learning rate halved every 5 epochs; each episode draws $N_q=12$ query trials with class templates built from $N_s=18$ trials sampled per class (re-sampled each episode); class balance is achieved by majority-class down-sampling with unweighted cross-entropy; and no augmentation is applied. The two-phase subject-level validation scaffolding is unchanged, so only the training recipe differs. Table~\ref{sup:tab-epmn-native} reports the results. Native-recipe EPMN was lower than shared-protocol EPMN on all three datasets, with a mean difference of 0.024 AUROC. Relative to ERP-XTTN, native EPMN was lower on ERN and P300 and tied at displayed precision on N400. Thus, the shared protocol did not disadvantage EPMN.

%% file: supplementary-validation.tex
This section specifies the grounding interventions and the ablation grid summarized in Section~\ref{sec:architecture-validation} of the main text. All analyses used the LOSO folds, five seeds, and mean-of-fold AUROC of the main evaluation protocol.

\subsection{Grounding Interventions}\label{sup:grounding-interventions} 
The grounding interventions tested whether the discriminative information carried by the routing and amplitude factors arose from their intended physiological content. The two factors were evaluated separately, as described below. All interventions were conducted on all nine datasets under the five-seed LOSO protocol and were applied to each fold's frozen model without retraining unless otherwise specified.

For the routing factor, five corruption conditions were applied to each prototype's match template: time reversal; replacement with variance-matched Gaussian noise (the \emph{noise null}); preservation of the amplitude spectrum while replacing the phase spectrum with independent uniform random values; polarity inversion through multiplication by $-1$; and a cross-component condition in which each slot's match template, window, and whitener were cyclically transferred to the next prototype slot while the routing key remained fixed. For the noise null, phase-randomized, time-reversed, and polarity-inverted conditions, each prototype was transformed within its original window and re-whitened using that window's fixed whitener. For the cross-component condition, the next prototype slot's template, window, and whitening specification were transferred together while the original routing key remained fixed. The match values $m[p,k]$ were then recomputed on the frozen model.

The unmodified prototypes provided the intact reference condition (\emph{Routing-only}; Tables~\ref{sup:tab-grounding-full-routing} and~\ref{sup:tab-grounding-3-routing}), whereas the corruption conditions were interpreted relative to the noise null. Because the routing keys and attention weights remained fixed, this null captured any residual discriminability recoverable from routing-pattern geometry alone and therefore replaced a nominal AUROC of 0.5 as the reference level. All five corruption conditions left the routing keys, detected trial peaks, and attention weights $a[p,k]$ unchanged. The routing key and match template derive from the same prototype waveform at different resolutions; corrupting the match template while holding the routing key fixed therefore creates a dissociation that cannot occur during normal operation. This dissociation isolates whether the readout depends on prototype content or only on attention-weight geometry.

Two permutation controls tested whether the trial-to-prototype correspondence carried discriminative information. In the first, the prototype columns of the match matrix were permuted within each trial, breaking the correspondence between attention column $k$ and match column $k$. In the second, match values were shuffled across trials, breaking the correspondence between each trial's attention weights and its own match values. If the routing decision is grounded in trial-specific prototype correspondence, both permutations should reduce performance to the noise null.

The amplitude factor was assessed separately. A trial-permutation control shuffled matched-filter features across trials, breaking their correspondence with the trial labels; a grounded amplitude factor should therefore collapse to this permutation reference. As with the noise null for the routing factor, this reference replaces a nominal AUROC of 0.5: because the diagnostic maximizes over individual features, it retains an upward bias under permutation. A window-localization control recomputed the matched filters over windows displaced from each component's detected latency by $-300$\,ms, clipped at the epoch start, and by $+150$\,ms. If the amplitude factor captures activity localized to the detected ERP component, its discriminative value should decrease under displacement. The two displacements are not equivalent tests: the early displacement ($-300$ ms) moves the window into the pre-component period, whereas the late displacement ($+150$ ms) is smaller than the maximum prototype window width (200 ms) and may partially overlap either the original window or a subsequent component, so it is the weaker control.

\subsection{Ablation Grid}\label{sup:ablation-grid}
Each ablation was a single-knob deviation from the model of Section~\ref{sec:architecture}. Ablations were run at the three-channel montage on HRI ErrP, ERN, P300, and N400 under the five-seed LOSO protocol and were compared with the unmodified model. Table~\ref{sup:tab-ablation} reports the full grid. The two-factor design was tested by removing either factor, producing routing-only and amplitude-only variants, and by training the routing and amplitude pathways end to end rather than fitting the combiner on the frozen outputs. The contribution of whitening was tested separately by replacing the whitened cosine with a raw cosine. Architectural hyperparameters were varied one at a time: maximum prototype count $K \in \{2, 6\}$ against the default cap of $4$, peak-detection prominence $\{0.01, 0.05\}$ against the default $0.02$, attention heads $\{2, 8\}$ against the default $4$, and removal of the self-attention layer. A final configuration replaced the grounded readout with the unconstrained learned classification head of the original ERP-XTTN \cite{wyman2026}; this arm was evaluated both for classification AUROC and under the grounding interventions of Section~\ref{sup:grounding-interventions}.

%% file: supplementary-tables.tex
% Table S1 — Full-montage AUROC
\begin{table}
\centering
\setlength{\tabcolsep}{5pt}
\caption{%
Full-montage classification performance (area under the receiver operating characteristic curve, AUROC; mean $\pm$ standard deviation across subjects) under leave-one-subject-out (LOSO) cross-validation. Channel counts vary by dataset (Table~\ref{tab:datasets} of the main text): 64 for BNCI, 27 for HRI, and 30 for all ERP CORE datasets. Neural models (ERP-XTTN, EEGNet, EEG-Deformer, EPMN) report seed-averaged means across five independent training seeds; xDAWN+RG is deterministic (single run). \textbf{Bold} marks the highest mean per dataset. $\Delta$ = best baseline minus ERP-XTTN (performance gap, in AUROC); the best baseline is selected post hoc per dataset and $\Delta$ is reported descriptively. Datasets ordered as in main text Table~\ref{tab:auroc} (three-channel mean AUROC, descending). EEGNet: compact convolutional neural network \cite{lawhern2018}; EEG-Deformer: dense convolutional transformer \cite{ding2024}; EPMN: ERP Prototypical Matching Net \cite{wei2022}; xDAWN+RG: xDAWN spatial filtering with Riemannian geometry classification. Dataset abbreviations are defined in Table~\ref{tab:datasets} of the main text.%
}
\label{sup:tab-fullmontage-auroc}
\begin{tabular}{lcccccc}
\toprule
Dataset & ERP-XTTN & EEGNet & EEG-Deformer & EPMN & xDAWN+RG & $\Delta$ \\
\midrule
HRI & 0.850 $\pm$ 0.098 & 0.869 $\pm$ 0.092 & \textbf{0.877 $\pm$ 0.099} & 0.844 $\pm$ 0.101 & 0.869 $\pm$ 0.085 & \textit{0.027} \\
ERN & 0.925 $\pm$ 0.055 & \textbf{0.951 $\pm$ 0.042} & 0.942 $\pm$ 0.047 & 0.927 $\pm$ 0.052 & 0.928 $\pm$ 0.054 & \textit{0.026} \\
LRP & 0.850 $\pm$ 0.067 & \textbf{0.871 $\pm$ 0.070} & 0.868 $\pm$ 0.064 & 0.843 $\pm$ 0.066 & 0.854 $\pm$ 0.074 & \textit{0.021} \\
BNCI & 0.776 $\pm$ 0.062 & \textbf{0.845 $\pm$ 0.063} & 0.842 $\pm$ 0.089 & 0.811 $\pm$ 0.072 & 0.820 $\pm$ 0.061 & \textit{0.069} \\
N170 & 0.768 $\pm$ 0.099 & 0.755 $\pm$ 0.100 & 0.774 $\pm$ 0.104 & 0.732 $\pm$ 0.095 & \textbf{0.793 $\pm$ 0.098} & \textit{0.025} \\
P300 & 0.700 $\pm$ 0.084 & 0.772 $\pm$ 0.092 & \textbf{0.780 $\pm$ 0.095} & 0.732 $\pm$ 0.088 & 0.722 $\pm$ 0.090 & \textit{0.080} \\
N2pc & 0.682 $\pm$ 0.076 & 0.698 $\pm$ 0.088 & \textbf{0.701 $\pm$ 0.089} & 0.670 $\pm$ 0.075 & 0.672 $\pm$ 0.077 & \textit{0.019} \\
MMN & 0.611 $\pm$ 0.046 & 0.617 $\pm$ 0.048 & 0.614 $\pm$ 0.049 & 0.572 $\pm$ 0.036 & \textbf{0.626 $\pm$ 0.052} & \textit{0.015} \\
N400 & 0.723 $\pm$ 0.075 & \textbf{0.761 $\pm$ 0.079} & 0.727 $\pm$ 0.080 & 0.720 $\pm$ 0.084 & 0.712 $\pm$ 0.078 & \textit{0.038} \\
\midrule
\textit{Mean} & 0.765 & \textbf{0.793} & 0.792 & 0.761 & 0.777 & \textbf{0.036} \\
\bottomrule
\end{tabular}
\end{table}

% Table S2 — Full-montage balanced accuracy
\begin{table}
\centering
\setlength{\tabcolsep}{5pt}
\caption{%
Full-montage balanced accuracy (mean $\pm$ standard deviation across subjects, LOSO) at threshold 0.5. The decision threshold was fixed at 0.5 on the predicted positive-class probability and applied uniformly across all subjects and configurations, simulating deployment conditions where per-subject threshold optimization is unavailable. \textbf{Bold} marks the highest mean per dataset. $\Delta$ = best baseline minus ERP-XTTN. ERP-XTTN reports the fused (routing $+$ amplitude) decision; neural models report seed-averaged means across five training seeds; xDAWN+RG is deterministic (single run). Datasets ordered as in main text Table~\ref{tab:auroc} (three-channel mean AUROC, descending). Three-channel balanced accuracy is in Table~\ref{sup:tab-balacc}. Abbreviations as in Table~\ref{sup:tab-fullmontage-auroc}.%
}
\label{sup:tab-fullmontage-ba}
\begin{tabular}{lcccccc}
\toprule
Dataset & ERP-XTTN & EEGNet & EEG-Deformer & EPMN & xDAWN+RG & $\Delta$ \\
\midrule
HRI & 0.774 $\pm$ 0.094 & 0.771 $\pm$ 0.098 & 0.787 $\pm$ 0.093 & 0.753 $\pm$ 0.099 & \textbf{0.788 $\pm$ 0.099} & \textit{0.014} \\
ERN & 0.852 $\pm$ 0.072 & \textbf{0.868 $\pm$ 0.081} & 0.830 $\pm$ 0.096 & 0.806 $\pm$ 0.098 & 0.756 $\pm$ 0.110 & \textit{0.016} \\
LRP & 0.769 $\pm$ 0.065 & \textbf{0.780 $\pm$ 0.072} & 0.765 $\pm$ 0.063 & 0.752 $\pm$ 0.059 & 0.771 $\pm$ 0.069 & \textit{0.011} \\
BNCI & 0.694 $\pm$ 0.046 & 0.753 $\pm$ 0.073 & \textbf{0.757 $\pm$ 0.089} & 0.708 $\pm$ 0.082 & 0.693 $\pm$ 0.076 & \textit{0.063} \\
N170 & 0.686 $\pm$ 0.079 & 0.677 $\pm$ 0.079 & 0.693 $\pm$ 0.082 & 0.653 $\pm$ 0.067 & \textbf{0.710 $\pm$ 0.083} & \textit{0.024} \\
P300 & 0.636 $\pm$ 0.065 & 0.694 $\pm$ 0.076 & \textbf{0.701 $\pm$ 0.077} & 0.632 $\pm$ 0.069 & 0.577 $\pm$ 0.066 & \textit{0.065} \\
N2pc & 0.626 $\pm$ 0.057 & 0.636 $\pm$ 0.064 & \textbf{0.640 $\pm$ 0.065} & 0.617 $\pm$ 0.054 & 0.614 $\pm$ 0.056 & \textit{0.014} \\
MMN & 0.578 $\pm$ 0.035 & \textbf{0.583 $\pm$ 0.035} & 0.573 $\pm$ 0.035 & 0.536 $\pm$ 0.023 & 0.503 $\pm$ 0.005 & \textit{0.005} \\
N400 & 0.662 $\pm$ 0.062 & \textbf{0.681 $\pm$ 0.066} & 0.653 $\pm$ 0.064 & 0.648 $\pm$ 0.067 & 0.651 $\pm$ 0.069 & \textit{0.019} \\
\midrule
\textit{Mean} & 0.697 & \textbf{0.716} & 0.711 & 0.678 & 0.674 & \textbf{0.026} \\
\bottomrule
\end{tabular}
\end{table}

% Table S3 — Routing grounding, full-montage
\begin{table}
\scriptsize
\centering
\setlength{\tabcolsep}{5pt}
\caption{%
Full-montage routing grounding interventions (mean AUROC, LOSO, seed-averaged across five seeds). The ERP-XTTN column reports the full two-factor model AUROC (from Table~\ref{sup:tab-fullmontage-auroc}) as a reference. All remaining columns report the routing-factor AUROC under the indicated intervention, applied to each fold's frozen trained model without retraining. Routing-only: unmodified prototypes (intact reference). Time-rev: prototype waveforms time-reversed. Noise null: prototypes replaced by variance-matched Gaussian noise, establishing the baseline discriminability recoverable from routing-pattern geometry alone. Phase-rand: prototype waveforms phase-randomized (amplitude spectrum preserved, phase shuffled). Polarity-inv: prototypes negated. Cross-comp: match-pathway states are cyclically shifted across prototype slots while routing keys remain fixed. Proto-perm: prototype columns of the match matrix are permuted independently within each trial. Trial-perm: which trial supplies each match value is permuted. Datasets ordered by mean AUROC across all five methods, descending. Standard deviations are omitted for compactness; fold-level and seed-level variability are available in the released code \cite{wyman2026erpxttn_code}. Abbreviations as in Table~\ref{sup:tab-fullmontage-auroc}.%
}
\label{sup:tab-grounding-full-routing}
\begin{tabular}{lccccccccc}
\toprule
Dataset & ERP-XTTN & Routing-only & Time-rev. & Noise null & Phase-rand. & Polarity-inv. & Cross-comp. & Proto-perm & Trial-perm \\
\midrule
HRI & 0.85 & 0.78 & 0.75 & 0.51 & 0.53 & 0.22 & 0.24 & 0.50 & 0.52 \\
ERN & 0.93 & 0.90 & 0.88 & 0.49 & 0.50 & 0.10 & 0.14 & 0.49 & 0.55 \\
LRP & 0.85 & 0.80 & 0.75 & 0.49 & 0.31 & 0.20 & 0.20 & 0.51 & 0.51 \\
BNCI & 0.78 & 0.74 & 0.72 & 0.48 & 0.55 & 0.26 & 0.30 & 0.49 & 0.50 \\
N170 & 0.77 & 0.66 & 0.59 & 0.50 & 0.49 & 0.34 & 0.38 & 0.49 & 0.50 \\
P300 & 0.70 & 0.66 & 0.64 & 0.49 & 0.50 & 0.34 & 0.37 & 0.49 & 0.50 \\
N2pc & 0.68 & 0.65 & 0.63 & 0.49 & 0.46 & 0.35 & 0.43 & 0.50 & 0.50 \\
MMN & 0.61 & 0.57 & 0.56 & 0.50 & 0.50 & 0.43 & 0.45 & 0.50 & 0.50 \\
N400 & 0.72 & 0.70 & 0.66 & 0.51 & 0.59 & 0.30 & 0.35 & 0.48 & 0.51 \\
\midrule
\textit{Mean} & 0.76 & 0.72 & 0.68 & 0.50 & 0.49 & 0.28 & 0.32 & 0.49 & 0.51 \\
\bottomrule
\end{tabular}
\end{table}

% Table S4 — Amplitude grounding, full-montage
\begin{sidewaystable}[p]
\centering
\setlength{\tabcolsep}{5pt}
\caption{%
Full-montage amplitude grounding controls (mean AUROC, LOSO, seed-averaged across five seeds). The ERP-XTTN column reports the full two-factor model AUROC (from Table~\ref{sup:tab-fullmontage-auroc}) as a reference. Amplitude features are evaluated separately for the monopolar channel matched filter (Channel MF; Equation~\ref{eq:mfchan}) and the bipolar contrast matched filter (Contrast MF; Equation~\ref{eq:mfctr}). For each derivation: On-target = matched filter computed over the prototype's detected component window; Off-target (early) = window displaced $-300$\,ms from the detected latency (clipped to epoch start); Off-target (late) = window displaced $+150$\,ms; Permutation = matched-filter features shuffled across trials. For each fold and feature family, AUROC is the maximum orientation-invariant AUROC over individual features. These values constitute a descriptive best-feature diagnostic rather than the performance of a trained classifier or the logistic combiner. Datasets ordered by mean AUROC across all five methods, descending. Standard deviations are omitted and are available in the released code \cite{wyman2026erpxttn_code}. Abbreviations as in Table~\ref{sup:tab-fullmontage-auroc}.%
}
\label{sup:tab-grounding-full-amp}
\begin{tabular}{l c cccc cccc}
\toprule
 & & \multicolumn{4}{c}{Channel MF} & \multicolumn{4}{c}{Contrast MF} \\
\cmidrule(lr){3-6} \cmidrule(lr){7-10}
Dataset & ERP-XTTN & On-target & Off-target (early) & Off-target (late) & Permutation & On-target & Off-target (early) & Off-target (late) & Permutation \\
\midrule
HRI & 0.85 & 0.77 & 0.59 & 0.76 & 0.56 & 0.83 & 0.65 & 0.81 & 0.59 \\
ERN & 0.93 & 0.85 & 0.82 & 0.83 & 0.63 & 0.90 & 0.88 & 0.88 & 0.67 \\
LRP & 0.85 & 0.70 & 0.67 & 0.65 & 0.57 & 0.78 & 0.75 & 0.68 & 0.59 \\
BNCI & 0.78 & 0.74 & 0.60 & 0.72 & 0.56 & 0.78 & 0.65 & 0.76 & 0.58 \\
N170 & 0.77 & 0.78 & 0.61 & 0.72 & 0.62 & 0.82 & 0.65 & 0.77 & 0.64 \\
P300 & 0.70 & 0.69 & 0.62 & 0.72 & 0.64 & 0.72 & 0.66 & 0.75 & 0.67 \\
N2pc & 0.68 & 0.66 & 0.60 & 0.63 & 0.59 & 0.70 & 0.64 & 0.68 & 0.61 \\
MMN & 0.61 & 0.60 & 0.56 & 0.58 & 0.56 & 0.62 & 0.58 & 0.60 & 0.57 \\
N400 & 0.72 & 0.73 & 0.67 & 0.70 & 0.64 & 0.75 & 0.70 & 0.74 & 0.67 \\
\midrule
\textit{Mean} & 0.76 & 0.72 & 0.64 & 0.70 & 0.60 & 0.77 & 0.68 & 0.74 & 0.62 \\
\bottomrule
\end{tabular}
\end{sidewaystable}

% Table S5 — EPMN native vs shared protocol
\begin{table}
\centering
\setlength{\tabcolsep}{6pt}
\caption{%
EPMN native-recipe versus shared-protocol comparison (mean $\pm$ standard deviation across subjects, LOSO) at the three-channel montage. The native recipe follows the training procedure of Wei et al.\ \cite{wei2022}: input resampled to 64 time points; SGD optimizer (learning rate $10^{-4}$, momentum 0.9) with learning-rate halving every 5 epochs; episodic training with $N_q = 12$ query trials and $N_s = 18$ support trials per class, re-sampled each episode; majority-class down-sampling with unweighted cross-entropy; no augmentation. The shared protocol uses the AdamW optimizer, cosine-annealed learning rate, temporal jitter, and additive noise described in Section~\ref{sec:training} of the main text. Both protocols use the same two-phase subject-level validation scaffolding and LOSO folds. \textbf{Bold} marks the higher mean per dataset. $\Delta$ = shared-protocol minus native AUROC (positive indicates the shared protocol is higher). The comparison was run on the three datasets for which the native EPMN episodic recipe could be implemented from the original description. Abbreviations as in Table~\ref{sup:tab-fullmontage-auroc}.%
}
\label{sup:tab-epmn-native}
\begin{tabular}{lccc}
\toprule
Dataset & Native AUROC & Shared-protocol AUROC & $\Delta$ \\
\midrule
ERN & 0.776 $\pm$ 0.121 & \textbf{0.795 $\pm$ 0.096} & \textit{+0.019} \\
P300 & 0.629 $\pm$ 0.078 & \textbf{0.674 $\pm$ 0.077} & \textit{+0.045} \\
N400 & 0.565 $\pm$ 0.076 & \textbf{0.573 $\pm$ 0.077} & \textit{+0.008} \\
\midrule
\textit{Mean} & 0.656 & \textbf{0.681} & \textbf{+0.024} \\
\bottomrule
\end{tabular}
\end{table}

% Table S6 — 3-channel balanced accuracy
\begin{table}
\centering
\setlength{\tabcolsep}{5pt}
\caption{%
Balanced accuracy (mean $\pm$ standard deviation across subjects, LOSO) at threshold 0.5, three-channel montage. \textbf{Bold} marks the highest mean per dataset. $\Delta$ = best baseline minus ERP-XTTN. Full-montage balanced accuracy is in Table~\ref{sup:tab-fullmontage-ba}; AUROC values at the three-channel montage are in main text Table~\ref{tab:auroc}. Abbreviations as in Table~\ref{sup:tab-fullmontage-auroc}.%
}
\label{sup:tab-balacc}
\begin{tabular}{lcccccc}
\toprule
Dataset & ERP-XTTN & EEGNet & EEG-Deformer & EPMN & xDAWN+RG & $\Delta$ \\
\midrule
HRI & 0.766 $\pm$ 0.096 & \textbf{0.775 $\pm$ 0.097} & 0.773 $\pm$ 0.090 & 0.738 $\pm$ 0.099 & 0.749 $\pm$ 0.106 & \textit{0.009} \\
ERN & 0.745 $\pm$ 0.097 & \textbf{0.773 $\pm$ 0.098} & 0.740 $\pm$ 0.103 & 0.697 $\pm$ 0.093 & 0.568 $\pm$ 0.074 & \textit{0.028} \\
LRP & 0.689 $\pm$ 0.065 & 0.705 $\pm$ 0.069 & 0.691 $\pm$ 0.070 & 0.694 $\pm$ 0.063 & \textbf{0.716 $\pm$ 0.064} & \textit{0.027} \\
BNCI & 0.708 $\pm$ 0.074 & \textbf{0.711 $\pm$ 0.078} & 0.707 $\pm$ 0.074 & 0.668 $\pm$ 0.075 & 0.637 $\pm$ 0.059 & \textit{0.003} \\
N170 & 0.661 $\pm$ 0.077 & 0.646 $\pm$ 0.090 & 0.644 $\pm$ 0.081 & 0.636 $\pm$ 0.077 & \textbf{0.667 $\pm$ 0.087} & \textit{0.006} \\
P300 & 0.618 $\pm$ 0.053 & \textbf{0.649 $\pm$ 0.070} & 0.643 $\pm$ 0.061 & 0.596 $\pm$ 0.054 & 0.512 $\pm$ 0.019 & \textit{0.031} \\
N2pc & 0.595 $\pm$ 0.061 & 0.599 $\pm$ 0.071 & 0.594 $\pm$ 0.066 & 0.588 $\pm$ 0.061 & \textbf{0.601 $\pm$ 0.070} & \textit{0.006} \\
MMN & 0.561 $\pm$ 0.036 & \textbf{0.564 $\pm$ 0.036} & 0.554 $\pm$ 0.031 & 0.534 $\pm$ 0.022 & 0.500 $\pm$ 0.000 & \textit{0.003} \\
N400 & 0.544 $\pm$ 0.058 & \textbf{0.563 $\pm$ 0.065} & 0.548 $\pm$ 0.052 & 0.550 $\pm$ 0.055 & 0.554 $\pm$ 0.059 & \textit{0.019} \\
\midrule
\textit{Mean} & 0.654 & \textbf{0.665} & 0.655 & 0.633 & 0.612 & \textbf{0.015} \\
\bottomrule
\end{tabular}
\end{table}

% Per-subject AUROC tables, one per dataset (S7--S15).

% Table S7 — Per-subject AUROC, HRI
\begin{table}
\centering
\caption{%
Per-subject AUROC (mean $\pm$ standard deviation across five training seeds, LOSO) at the three-channel montage for HRI ErrP ($n = 11$ subjects). Subject numbering follows the original dataset release, which does not include sub-01 or sub-12; all eleven subjects distributed in the public release were analyzed. Each row is one LOSO fold with the indicated subject held out; the reported value is that subject's test-set AUROC averaged across seeds (xDAWN+RG: single deterministic run, standard deviation omitted). \textbf{Bold} marks the highest mean per subject. The mean row matches the HRI entry of main text Table~\ref{tab:auroc}. $\Delta$ is computed per subject as that subject's best-baseline AUROC minus ERP-XTTN AUROC, then averaged; because the identity of the best baseline varies across subjects, the mean $\Delta$ here is generally larger than the fixed-best-baseline $\Delta$ reported in Table~\ref{tab:auroc}. Abbreviations as in Table~\ref{sup:tab-fullmontage-auroc}.%
}
\label{sup:tab-persubj-hri}
\begin{tabular}{lcccccc}
\toprule
Subject & ERP-XTTN & EEGNet & EEG-Deformer & EPMN & xDAWN+RG & $\Delta$ \\
\midrule
sub-02 & 0.753 $\pm$ 0.008 & 0.813 $\pm$ 0.004 & \textbf{0.814 $\pm$ 0.018} & 0.759 $\pm$ 0.041 & 0.736 & \textit{0.061} \\
sub-03 & 0.858 $\pm$ 0.004 & 0.863 $\pm$ 0.004 & \textbf{0.880 $\pm$ 0.008} & 0.872 $\pm$ 0.014 & 0.875 & \textit{0.023} \\
sub-04 & 0.976 $\pm$ 0.001 & \textbf{0.984 $\pm$ 0.002} & 0.984 $\pm$ 0.002 & 0.971 $\pm$ 0.006 & 0.976 & \textit{0.008} \\
sub-05 & 0.735 $\pm$ 0.007 & 0.776 $\pm$ 0.004 & \textbf{0.780 $\pm$ 0.012} & 0.717 $\pm$ 0.015 & 0.742 & \textit{0.045} \\
sub-06 & 0.850 $\pm$ 0.008 & \textbf{0.909 $\pm$ 0.006} & 0.889 $\pm$ 0.014 & 0.825 $\pm$ 0.044 & 0.877 & \textit{0.059} \\
sub-07 & 0.864 $\pm$ 0.009 & \textbf{0.884 $\pm$ 0.004} & 0.853 $\pm$ 0.012 & 0.852 $\pm$ 0.032 & 0.860 & \textit{0.020} \\
sub-08 & 0.877 $\pm$ 0.006 & 0.876 $\pm$ 0.008 & \textbf{0.884 $\pm$ 0.010} & 0.864 $\pm$ 0.025 & 0.854 & \textit{0.007} \\
sub-09 & 0.911 $\pm$ 0.004 & 0.912 $\pm$ 0.006 & 0.900 $\pm$ 0.012 & 0.900 $\pm$ 0.024 & \textbf{0.935} & \textit{0.025} \\
sub-10 & \textbf{0.612 $\pm$ 0.009} & 0.607 $\pm$ 0.010 & 0.606 $\pm$ 0.028 & 0.611 $\pm$ 0.021 & 0.597 & \textit{-0.001} \\
sub-11 & \textbf{0.874 $\pm$ 0.009} & 0.871 $\pm$ 0.017 & 0.858 $\pm$ 0.015 & 0.872 $\pm$ 0.016 & 0.846 & \textit{-0.002} \\
sub-13 & 0.900 $\pm$ 0.006 & \textbf{0.945 $\pm$ 0.002} & 0.918 $\pm$ 0.021 & 0.901 $\pm$ 0.009 & 0.905 & \textit{0.044} \\
\midrule
\textit{Mean} & 0.837 $\pm$ 0.101 & \textbf{0.858 $\pm$ 0.101} & 0.852 $\pm$ 0.097 & 0.831 $\pm$ 0.100 & 0.837 $\pm$ 0.107 & \textbf{0.026} \\
\bottomrule
\end{tabular}
\end{table}

% Table S8 — Per-subject AUROC, ERN
\begin{table}
\centering
\caption{%
Per-subject AUROC (mean $\pm$ standard deviation across five training seeds, LOSO) at the three-channel montage for ERP CORE ERN ($n = 40$ subjects). Conventions as in Table~\ref{sup:tab-persubj-hri}. The mean row matches the ERN entry of main text Table~\ref{tab:auroc}.%
}
\label{sup:tab-persubj-ern}
\begin{tabular}{lcccccc}
\toprule
Subject & ERP-XTTN & EEGNet & EEG-Deformer & EPMN & xDAWN+RG & $\Delta$ \\
\midrule
sub-01 & 0.759 $\pm$ 0.018 & \textbf{0.819 $\pm$ 0.011} & 0.792 $\pm$ 0.032 & 0.700 $\pm$ 0.027 & 0.689 & \textit{0.060} \\
sub-02 & 0.709 $\pm$ 0.023 & \textbf{0.892 $\pm$ 0.023} & 0.871 $\pm$ 0.069 & 0.748 $\pm$ 0.096 & 0.630 & \textit{0.182} \\
sub-03 & 0.920 $\pm$ 0.006 & \textbf{0.958 $\pm$ 0.006} & 0.937 $\pm$ 0.018 & 0.851 $\pm$ 0.086 & 0.856 & \textit{0.038} \\
sub-04 & \textbf{0.934 $\pm$ 0.008} & 0.931 $\pm$ 0.012 & 0.868 $\pm$ 0.063 & 0.911 $\pm$ 0.027 & 0.929 & \textit{-0.003} \\
sub-05 & \textbf{0.995 $\pm$ 0.002} & 0.948 $\pm$ 0.009 & 0.979 $\pm$ 0.017 & 0.880 $\pm$ 0.186 & 0.891 & \textit{-0.016} \\
sub-06 & 0.724 $\pm$ 0.013 & 0.777 $\pm$ 0.014 & 0.734 $\pm$ 0.037 & 0.683 $\pm$ 0.023 & \textbf{0.786} & \textit{0.062} \\
sub-07 & 0.555 $\pm$ 0.028 & 0.579 $\pm$ 0.043 & 0.581 $\pm$ 0.074 & \textbf{0.675 $\pm$ 0.067} & 0.402 & \textit{0.120} \\
sub-08 & 0.908 $\pm$ 0.011 & \textbf{0.931 $\pm$ 0.006} & 0.922 $\pm$ 0.007 & 0.871 $\pm$ 0.018 & 0.920 & \textit{0.024} \\
sub-09 & \textbf{0.769 $\pm$ 0.034} & 0.761 $\pm$ 0.014 & 0.754 $\pm$ 0.047 & 0.655 $\pm$ 0.122 & 0.694 & \textit{-0.008} \\
sub-10 & 0.663 $\pm$ 0.008 & \textbf{0.784 $\pm$ 0.009} & 0.710 $\pm$ 0.059 & 0.671 $\pm$ 0.059 & 0.722 & \textit{0.121} \\
sub-11 & 0.635 $\pm$ 0.018 & \textbf{0.686 $\pm$ 0.014} & 0.666 $\pm$ 0.009 & 0.617 $\pm$ 0.053 & 0.617 & \textit{0.051} \\
sub-12 & 0.858 $\pm$ 0.012 & \textbf{0.880 $\pm$ 0.007} & 0.835 $\pm$ 0.019 & 0.863 $\pm$ 0.032 & 0.804 & \textit{0.023} \\
sub-13 & 0.778 $\pm$ 0.018 & \textbf{0.852 $\pm$ 0.026} & 0.827 $\pm$ 0.034 & 0.728 $\pm$ 0.074 & 0.842 & \textit{0.074} \\
sub-14 & 0.886 $\pm$ 0.014 & \textbf{0.935 $\pm$ 0.009} & 0.870 $\pm$ 0.043 & 0.876 $\pm$ 0.054 & 0.846 & \textit{0.050} \\
sub-15 & 0.942 $\pm$ 0.010 & 0.946 $\pm$ 0.007 & \textbf{0.952 $\pm$ 0.015} & 0.915 $\pm$ 0.020 & 0.950 & \textit{0.010} \\
sub-16 & 0.927 $\pm$ 0.011 & \textbf{0.963 $\pm$ 0.004} & 0.951 $\pm$ 0.004 & 0.927 $\pm$ 0.013 & 0.918 & \textit{0.035} \\
sub-17 & 0.867 $\pm$ 0.013 & \textbf{0.936 $\pm$ 0.005} & 0.856 $\pm$ 0.026 & 0.837 $\pm$ 0.062 & 0.780 & \textit{0.069} \\
sub-18 & 0.844 $\pm$ 0.015 & \textbf{0.884 $\pm$ 0.009} & 0.823 $\pm$ 0.028 & 0.789 $\pm$ 0.060 & 0.829 & \textit{0.040} \\
sub-19 & 0.921 $\pm$ 0.015 & \textbf{0.969 $\pm$ 0.004} & 0.946 $\pm$ 0.016 & 0.930 $\pm$ 0.049 & 0.911 & \textit{0.048} \\
sub-20 & 0.768 $\pm$ 0.026 & \textbf{0.855 $\pm$ 0.029} & 0.797 $\pm$ 0.019 & 0.760 $\pm$ 0.076 & 0.565 & \textit{0.087} \\
sub-21 & 0.891 $\pm$ 0.009 & \textbf{0.954 $\pm$ 0.005} & 0.911 $\pm$ 0.030 & 0.843 $\pm$ 0.148 & 0.932 & \textit{0.063} \\
sub-22 & 0.855 $\pm$ 0.010 & \textbf{0.890 $\pm$ 0.003} & 0.873 $\pm$ 0.020 & 0.847 $\pm$ 0.024 & 0.853 & \textit{0.034} \\
sub-23 & 0.917 $\pm$ 0.011 & \textbf{0.964 $\pm$ 0.006} & 0.952 $\pm$ 0.015 & 0.860 $\pm$ 0.095 & 0.858 & \textit{0.047} \\
sub-24 & 0.941 $\pm$ 0.010 & \textbf{0.956 $\pm$ 0.008} & 0.940 $\pm$ 0.022 & 0.814 $\pm$ 0.226 & 0.945 & \textit{0.015} \\
sub-25 & 0.607 $\pm$ 0.020 & \textbf{0.796 $\pm$ 0.031} & 0.794 $\pm$ 0.040 & 0.655 $\pm$ 0.043 & 0.512 & \textit{0.189} \\
sub-26 & 0.990 $\pm$ 0.003 & \textbf{0.996 $\pm$ 0.000} & 0.994 $\pm$ 0.002 & 0.987 $\pm$ 0.002 & 0.992 & \textit{0.006} \\
sub-27 & 0.813 $\pm$ 0.016 & \textbf{0.882 $\pm$ 0.004} & 0.779 $\pm$ 0.043 & 0.788 $\pm$ 0.038 & 0.843 & \textit{0.068} \\
sub-28 & 0.715 $\pm$ 0.034 & \textbf{0.760 $\pm$ 0.009} & 0.720 $\pm$ 0.032 & 0.721 $\pm$ 0.050 & 0.687 & \textit{0.045} \\
sub-29 & 0.794 $\pm$ 0.011 & \textbf{0.798 $\pm$ 0.023} & 0.762 $\pm$ 0.054 & 0.722 $\pm$ 0.106 & 0.725 & \textit{0.004} \\
sub-30 & 0.932 $\pm$ 0.006 & \textbf{0.940 $\pm$ 0.009} & 0.903 $\pm$ 0.029 & 0.878 $\pm$ 0.054 & 0.897 & \textit{0.008} \\
sub-31 & 0.902 $\pm$ 0.009 & \textbf{0.950 $\pm$ 0.004} & 0.935 $\pm$ 0.013 & 0.928 $\pm$ 0.029 & 0.873 & \textit{0.048} \\
sub-32 & 0.938 $\pm$ 0.012 & \textbf{0.945 $\pm$ 0.004} & 0.917 $\pm$ 0.032 & 0.862 $\pm$ 0.018 & 0.899 & \textit{0.007} \\
sub-33 & 0.851 $\pm$ 0.011 & \textbf{0.886 $\pm$ 0.011} & 0.822 $\pm$ 0.023 & 0.801 $\pm$ 0.078 & 0.726 & \textit{0.035} \\
sub-34 & 0.726 $\pm$ 0.020 & 0.800 $\pm$ 0.007 & \textbf{0.804 $\pm$ 0.025} & 0.718 $\pm$ 0.017 & 0.709 & \textit{0.077} \\
sub-35 & 0.898 $\pm$ 0.008 & \textbf{0.949 $\pm$ 0.003} & 0.906 $\pm$ 0.009 & 0.830 $\pm$ 0.101 & 0.880 & \textit{0.051} \\
sub-36 & 0.776 $\pm$ 0.016 & 0.755 $\pm$ 0.012 & 0.744 $\pm$ 0.026 & 0.773 $\pm$ 0.029 & \textbf{0.806} & \textit{0.030} \\
sub-37 & 0.827 $\pm$ 0.020 & 0.861 $\pm$ 0.016 & \textbf{0.873 $\pm$ 0.021} & 0.789 $\pm$ 0.064 & 0.816 & \textit{0.046} \\
sub-38 & 0.771 $\pm$ 0.006 & \textbf{0.812 $\pm$ 0.011} & 0.788 $\pm$ 0.010 & 0.725 $\pm$ 0.116 & 0.740 & \textit{0.042} \\
sub-39 & 0.832 $\pm$ 0.009 & \textbf{0.882 $\pm$ 0.006} & 0.854 $\pm$ 0.049 & 0.775 $\pm$ 0.072 & 0.807 & \textit{0.050} \\
sub-40 & 0.639 $\pm$ 0.014 & \textbf{0.704 $\pm$ 0.026} & 0.651 $\pm$ 0.079 & 0.599 $\pm$ 0.115 & 0.651 & \textit{0.065} \\
\midrule
\textit{Mean} & 0.824 $\pm$ 0.110 & \textbf{0.869 $\pm$ 0.093} & 0.840 $\pm$ 0.097 & 0.795 $\pm$ 0.096 & 0.793 $\pm$ 0.130 & \textbf{0.050} \\
\bottomrule
\end{tabular}
\end{table}

% Table S9 — Per-subject AUROC, LRP
\begin{table}
\centering
\caption{%
Per-subject AUROC (mean $\pm$ standard deviation across five training seeds, LOSO) at the three-channel montage for ERP CORE LRP ($n = 40$ subjects). Conventions as in Table~\ref{sup:tab-persubj-hri}. The mean row matches the LRP entry of main text Table~\ref{tab:auroc}.%
}
\label{sup:tab-persubj-lrp}
\begin{tabular}{lcccccc}
\toprule
Subject & ERP-XTTN & EEGNet & EEG-Deformer & EPMN & xDAWN+RG & $\Delta$ \\
\midrule
sub-01 & 0.763 $\pm$ 0.005 & 0.786 $\pm$ 0.014 & 0.760 $\pm$ 0.043 & 0.786 $\pm$ 0.017 & \textbf{0.807} & \textit{0.043} \\
sub-02 & 0.831 $\pm$ 0.004 & 0.867 $\pm$ 0.008 & \textbf{0.875 $\pm$ 0.005} & 0.844 $\pm$ 0.009 & 0.860 & \textit{0.044} \\
sub-03 & 0.715 $\pm$ 0.016 & 0.688 $\pm$ 0.008 & 0.666 $\pm$ 0.046 & 0.639 $\pm$ 0.110 & \textbf{0.718} & \textit{0.004} \\
sub-04 & 0.774 $\pm$ 0.004 & 0.843 $\pm$ 0.012 & 0.851 $\pm$ 0.009 & 0.806 $\pm$ 0.030 & \textbf{0.871} & \textit{0.097} \\
sub-05 & 0.763 $\pm$ 0.007 & \textbf{0.795 $\pm$ 0.011} & 0.746 $\pm$ 0.036 & 0.772 $\pm$ 0.010 & 0.781 & \textit{0.033} \\
sub-06 & 0.747 $\pm$ 0.004 & 0.745 $\pm$ 0.008 & 0.750 $\pm$ 0.008 & 0.737 $\pm$ 0.008 & \textbf{0.758} & \textit{0.010} \\
sub-07 & 0.677 $\pm$ 0.002 & 0.701 $\pm$ 0.002 & 0.719 $\pm$ 0.015 & 0.702 $\pm$ 0.025 & \textbf{0.762} & \textit{0.085} \\
sub-08 & 0.833 $\pm$ 0.004 & 0.849 $\pm$ 0.005 & 0.831 $\pm$ 0.017 & 0.849 $\pm$ 0.020 & \textbf{0.856} & \textit{0.023} \\
sub-09 & 0.574 $\pm$ 0.007 & 0.589 $\pm$ 0.007 & \textbf{0.624 $\pm$ 0.016} & 0.586 $\pm$ 0.030 & 0.613 & \textit{0.050} \\
sub-10 & 0.584 $\pm$ 0.006 & 0.592 $\pm$ 0.011 & 0.582 $\pm$ 0.015 & 0.593 $\pm$ 0.015 & \textbf{0.635} & \textit{0.051} \\
sub-11 & 0.765 $\pm$ 0.005 & 0.814 $\pm$ 0.006 & 0.818 $\pm$ 0.010 & 0.800 $\pm$ 0.025 & \textbf{0.835} & \textit{0.070} \\
sub-12 & 0.849 $\pm$ 0.002 & 0.870 $\pm$ 0.004 & 0.872 $\pm$ 0.007 & 0.873 $\pm$ 0.013 & \textbf{0.876} & \textit{0.027} \\
sub-13 & 0.775 $\pm$ 0.012 & \textbf{0.792 $\pm$ 0.010} & 0.790 $\pm$ 0.012 & 0.762 $\pm$ 0.023 & 0.760 & \textit{0.017} \\
sub-14 & 0.809 $\pm$ 0.006 & 0.824 $\pm$ 0.008 & 0.848 $\pm$ 0.018 & 0.826 $\pm$ 0.024 & \textbf{0.880} & \textit{0.071} \\
sub-15 & 0.910 $\pm$ 0.003 & \textbf{0.939 $\pm$ 0.002} & 0.930 $\pm$ 0.030 & 0.899 $\pm$ 0.020 & 0.908 & \textit{0.030} \\
sub-16 & 0.697 $\pm$ 0.007 & \textbf{0.775 $\pm$ 0.006} & 0.734 $\pm$ 0.028 & 0.719 $\pm$ 0.063 & 0.759 & \textit{0.078} \\
sub-17 & 0.762 $\pm$ 0.006 & 0.796 $\pm$ 0.006 & \textbf{0.811 $\pm$ 0.013} & 0.760 $\pm$ 0.062 & 0.796 & \textit{0.049} \\
sub-18 & 0.640 $\pm$ 0.017 & \textbf{0.731 $\pm$ 0.004} & 0.730 $\pm$ 0.020 & 0.725 $\pm$ 0.019 & 0.706 & \textit{0.090} \\
sub-19 & 0.712 $\pm$ 0.004 & \textbf{0.734 $\pm$ 0.004} & 0.726 $\pm$ 0.024 & 0.715 $\pm$ 0.030 & 0.730 & \textit{0.022} \\
sub-20 & 0.790 $\pm$ 0.003 & 0.824 $\pm$ 0.005 & 0.819 $\pm$ 0.006 & 0.817 $\pm$ 0.016 & \textbf{0.830} & \textit{0.040} \\
sub-21 & 0.781 $\pm$ 0.008 & \textbf{0.848 $\pm$ 0.006} & 0.825 $\pm$ 0.013 & 0.797 $\pm$ 0.089 & 0.829 & \textit{0.067} \\
sub-22 & 0.891 $\pm$ 0.006 & 0.912 $\pm$ 0.003 & 0.905 $\pm$ 0.009 & 0.889 $\pm$ 0.018 & \textbf{0.915} & \textit{0.024} \\
sub-23 & 0.655 $\pm$ 0.011 & 0.722 $\pm$ 0.013 & 0.716 $\pm$ 0.007 & 0.690 $\pm$ 0.016 & \textbf{0.734} & \textit{0.079} \\
sub-24 & 0.722 $\pm$ 0.006 & 0.730 $\pm$ 0.006 & 0.746 $\pm$ 0.011 & 0.700 $\pm$ 0.015 & \textbf{0.756} & \textit{0.034} \\
sub-25 & 0.812 $\pm$ 0.011 & \textbf{0.874 $\pm$ 0.017} & 0.871 $\pm$ 0.017 & 0.856 $\pm$ 0.017 & 0.839 & \textit{0.062} \\
sub-26 & 0.831 $\pm$ 0.005 & 0.847 $\pm$ 0.003 & \textbf{0.852 $\pm$ 0.009} & 0.820 $\pm$ 0.027 & 0.827 & \textit{0.021} \\
sub-27 & 0.796 $\pm$ 0.005 & \textbf{0.808 $\pm$ 0.008} & 0.803 $\pm$ 0.010 & 0.793 $\pm$ 0.006 & 0.805 & \textit{0.012} \\
sub-28 & 0.734 $\pm$ 0.003 & 0.737 $\pm$ 0.005 & 0.718 $\pm$ 0.015 & 0.711 $\pm$ 0.044 & \textbf{0.764} & \textit{0.030} \\
sub-29 & 0.641 $\pm$ 0.005 & 0.665 $\pm$ 0.008 & 0.666 $\pm$ 0.009 & 0.628 $\pm$ 0.065 & \textbf{0.668} & \textit{0.027} \\
sub-30 & 0.712 $\pm$ 0.007 & 0.743 $\pm$ 0.016 & 0.755 $\pm$ 0.011 & 0.704 $\pm$ 0.027 & \textbf{0.759} & \textit{0.047} \\
sub-31 & 0.751 $\pm$ 0.005 & \textbf{0.777 $\pm$ 0.006} & 0.762 $\pm$ 0.023 & 0.761 $\pm$ 0.023 & 0.768 & \textit{0.026} \\
sub-32 & 0.891 $\pm$ 0.015 & 0.935 $\pm$ 0.011 & \textbf{0.939 $\pm$ 0.020} & 0.878 $\pm$ 0.055 & 0.894 & \textit{0.047} \\
sub-33 & 0.661 $\pm$ 0.006 & 0.703 $\pm$ 0.011 & 0.706 $\pm$ 0.015 & 0.687 $\pm$ 0.015 & \textbf{0.709} & \textit{0.048} \\
sub-34 & 0.780 $\pm$ 0.003 & 0.792 $\pm$ 0.005 & 0.770 $\pm$ 0.020 & 0.769 $\pm$ 0.028 & \textbf{0.794} & \textit{0.014} \\
sub-35 & 0.874 $\pm$ 0.005 & \textbf{0.910 $\pm$ 0.005} & 0.893 $\pm$ 0.019 & 0.879 $\pm$ 0.037 & 0.899 & \textit{0.036} \\
sub-36 & 0.853 $\pm$ 0.006 & \textbf{0.904 $\pm$ 0.004} & 0.844 $\pm$ 0.059 & 0.832 $\pm$ 0.101 & 0.887 & \textit{0.050} \\
sub-37 & 0.808 $\pm$ 0.004 & 0.852 $\pm$ 0.006 & 0.837 $\pm$ 0.007 & 0.836 $\pm$ 0.020 & \textbf{0.860} & \textit{0.052} \\
sub-38 & 0.721 $\pm$ 0.004 & 0.741 $\pm$ 0.007 & 0.742 $\pm$ 0.013 & 0.734 $\pm$ 0.017 & \textbf{0.769} & \textit{0.047} \\
sub-39 & 0.800 $\pm$ 0.004 & 0.830 $\pm$ 0.004 & 0.833 $\pm$ 0.012 & 0.812 $\pm$ 0.012 & \textbf{0.839} & \textit{0.039} \\
sub-40 & 0.771 $\pm$ 0.005 & \textbf{0.801 $\pm$ 0.007} & 0.788 $\pm$ 0.022 & 0.795 $\pm$ 0.012 & 0.800 & \textit{0.030} \\
\midrule
\textit{Mean} & 0.761 $\pm$ 0.080 & 0.792 $\pm$ 0.084 & 0.786 $\pm$ 0.081 & 0.769 $\pm$ 0.080 & \textbf{0.796 $\pm$ 0.074} & \textbf{0.043} \\
\bottomrule
\end{tabular}
\end{table}

% Table S10 — Per-subject AUROC, BNCI
\begin{table}
\centering
\caption{%
Per-subject AUROC (mean $\pm$ standard deviation across five training seeds, LOSO) at the three-channel montage for BNCI ErrP ($n = 6$ subjects). Conventions as in Table~\ref{sup:tab-persubj-hri}. The mean row matches the BNCI entry of main text Table~\ref{tab:auroc}.%
}
\label{sup:tab-persubj-bnci}
\begin{tabular}{lcccccc}
\toprule
Subject & ERP-XTTN & EEGNet & EEG-Deformer & EPMN & xDAWN+RG & $\Delta$ \\
\midrule
sub-01 & 0.881 $\pm$ 0.010 & \textbf{0.915 $\pm$ 0.008} & 0.882 $\pm$ 0.004 & 0.823 $\pm$ 0.028 & 0.897 & \textit{0.034} \\
sub-02 & 0.745 $\pm$ 0.006 & \textbf{0.795 $\pm$ 0.004} & 0.750 $\pm$ 0.010 & 0.741 $\pm$ 0.061 & 0.778 & \textit{0.050} \\
sub-03 & 0.825 $\pm$ 0.015 & 0.817 $\pm$ 0.013 & \textbf{0.852 $\pm$ 0.020} & 0.779 $\pm$ 0.016 & 0.823 & \textit{0.027} \\
sub-04 & 0.697 $\pm$ 0.010 & \textbf{0.754 $\pm$ 0.006} & 0.736 $\pm$ 0.018 & 0.693 $\pm$ 0.026 & 0.753 & \textit{0.057} \\
sub-05 & 0.816 $\pm$ 0.012 & 0.825 $\pm$ 0.005 & \textbf{0.840 $\pm$ 0.010} & 0.789 $\pm$ 0.021 & 0.811 & \textit{0.024} \\
sub-06 & 0.660 $\pm$ 0.007 & 0.648 $\pm$ 0.011 & \textbf{0.664 $\pm$ 0.019} & 0.539 $\pm$ 0.076 & 0.650 & \textit{0.004} \\
\midrule
\textit{Mean} & 0.771 $\pm$ 0.084 & \textbf{0.792 $\pm$ 0.088} & 0.787 $\pm$ 0.084 & 0.727 $\pm$ 0.102 & 0.785 $\pm$ 0.082 & \textbf{0.033} \\
\bottomrule
\end{tabular}
\end{table}

% Table S11 — Per-subject AUROC, N170
\begin{table}
\centering
\caption{%
Per-subject AUROC (mean $\pm$ standard deviation across five training seeds, LOSO) at the three-channel montage for ERP CORE N170 ($n = 40$ subjects). Conventions as in Table~\ref{sup:tab-persubj-hri}. The mean row matches the N170 entry of main text Table~\ref{tab:auroc}.%
}
\label{sup:tab-persubj-n170}
\begin{tabular}{lcccccc}
\toprule
Subject & ERP-XTTN & EEGNet & EEG-Deformer & EPMN & xDAWN+RG & $\Delta$ \\
\midrule
sub-01 & 0.577 $\pm$ 0.006 & 0.594 $\pm$ 0.019 & 0.583 $\pm$ 0.016 & 0.585 $\pm$ 0.014 & \textbf{0.648} & \textit{0.071} \\
sub-02 & 0.663 $\pm$ 0.007 & 0.663 $\pm$ 0.011 & 0.649 $\pm$ 0.023 & 0.669 $\pm$ 0.013 & \textbf{0.716} & \textit{0.053} \\
sub-03 & 0.680 $\pm$ 0.010 & 0.659 $\pm$ 0.025 & \textbf{0.681 $\pm$ 0.037} & 0.650 $\pm$ 0.008 & 0.665 & \textit{0.001} \\
sub-04 & 0.758 $\pm$ 0.032 & 0.720 $\pm$ 0.027 & \textbf{0.776 $\pm$ 0.022} & 0.711 $\pm$ 0.042 & 0.686 & \textit{0.018} \\
sub-05 & 0.703 $\pm$ 0.018 & 0.658 $\pm$ 0.019 & 0.707 $\pm$ 0.012 & 0.667 $\pm$ 0.017 & \textbf{0.740} & \textit{0.037} \\
sub-06 & 0.832 $\pm$ 0.018 & 0.842 $\pm$ 0.024 & 0.843 $\pm$ 0.019 & 0.843 $\pm$ 0.038 & \textbf{0.907} & \textit{0.075} \\
sub-07 & \textbf{0.844 $\pm$ 0.013} & 0.806 $\pm$ 0.020 & 0.794 $\pm$ 0.029 & 0.801 $\pm$ 0.031 & 0.765 & \textit{-0.039} \\
sub-08 & 0.744 $\pm$ 0.024 & 0.698 $\pm$ 0.013 & 0.621 $\pm$ 0.013 & 0.683 $\pm$ 0.083 & \textbf{0.828} & \textit{0.084} \\
sub-09 & 0.835 $\pm$ 0.010 & 0.850 $\pm$ 0.015 & 0.853 $\pm$ 0.015 & 0.808 $\pm$ 0.047 & \textbf{0.862} & \textit{0.028} \\
sub-10 & \textbf{0.741 $\pm$ 0.016} & 0.733 $\pm$ 0.018 & 0.725 $\pm$ 0.016 & 0.703 $\pm$ 0.029 & 0.721 & \textit{-0.008} \\
sub-11 & 0.650 $\pm$ 0.019 & 0.618 $\pm$ 0.026 & 0.570 $\pm$ 0.031 & 0.528 $\pm$ 0.074 & \textbf{0.657} & \textit{0.007} \\
sub-12 & 0.636 $\pm$ 0.016 & 0.612 $\pm$ 0.022 & 0.631 $\pm$ 0.010 & 0.655 $\pm$ 0.057 & \textbf{0.720} & \textit{0.084} \\
sub-13 & 0.718 $\pm$ 0.035 & 0.726 $\pm$ 0.032 & \textbf{0.732 $\pm$ 0.027} & 0.718 $\pm$ 0.009 & 0.696 & \textit{0.014} \\
sub-14 & \textbf{0.855 $\pm$ 0.019} & 0.777 $\pm$ 0.015 & 0.762 $\pm$ 0.034 & 0.771 $\pm$ 0.045 & 0.805 & \textit{-0.050} \\
sub-15 & 0.720 $\pm$ 0.016 & 0.703 $\pm$ 0.042 & \textbf{0.735 $\pm$ 0.012} & 0.691 $\pm$ 0.019 & 0.709 & \textit{0.016} \\
sub-16 & 0.879 $\pm$ 0.009 & 0.918 $\pm$ 0.014 & \textbf{0.937 $\pm$ 0.015} & 0.914 $\pm$ 0.005 & 0.869 & \textit{0.058} \\
sub-17 & 0.685 $\pm$ 0.008 & 0.640 $\pm$ 0.025 & 0.623 $\pm$ 0.045 & 0.606 $\pm$ 0.105 & \textbf{0.694} & \textit{0.009} \\
sub-18 & 0.634 $\pm$ 0.013 & 0.676 $\pm$ 0.032 & \textbf{0.688 $\pm$ 0.013} & 0.615 $\pm$ 0.024 & 0.617 & \textit{0.053} \\
sub-19 & 0.614 $\pm$ 0.014 & 0.654 $\pm$ 0.023 & 0.613 $\pm$ 0.012 & 0.664 $\pm$ 0.015 & \textbf{0.679} & \textit{0.066} \\
sub-20 & 0.752 $\pm$ 0.027 & 0.757 $\pm$ 0.046 & 0.789 $\pm$ 0.037 & 0.765 $\pm$ 0.038 & \textbf{0.849} & \textit{0.097} \\
sub-21 & 0.695 $\pm$ 0.015 & 0.637 $\pm$ 0.017 & 0.618 $\pm$ 0.031 & 0.639 $\pm$ 0.038 & \textbf{0.724} & \textit{0.029} \\
sub-22 & 0.594 $\pm$ 0.022 & 0.760 $\pm$ 0.020 & \textbf{0.848 $\pm$ 0.013} & 0.683 $\pm$ 0.119 & 0.573 & \textit{0.254} \\
sub-23 & 0.602 $\pm$ 0.008 & 0.615 $\pm$ 0.010 & 0.592 $\pm$ 0.031 & 0.599 $\pm$ 0.025 & \textbf{0.638} & \textit{0.035} \\
sub-24 & \textbf{0.919 $\pm$ 0.007} & 0.824 $\pm$ 0.014 & 0.823 $\pm$ 0.026 & 0.859 $\pm$ 0.027 & 0.909 & \textit{-0.010} \\
sub-25 & 0.614 $\pm$ 0.011 & 0.555 $\pm$ 0.025 & 0.531 $\pm$ 0.027 & 0.565 $\pm$ 0.031 & \textbf{0.654} & \textit{0.040} \\
sub-26 & \textbf{0.821 $\pm$ 0.030} & 0.752 $\pm$ 0.026 & 0.673 $\pm$ 0.044 & 0.782 $\pm$ 0.043 & 0.799 & \textit{-0.021} \\
sub-27 & 0.735 $\pm$ 0.022 & 0.759 $\pm$ 0.012 & \textbf{0.769 $\pm$ 0.009} & \textbf{0.770 $\pm$ 0.010} & 0.732 & \textit{0.035} \\
sub-28 & 0.846 $\pm$ 0.016 & \textbf{0.878 $\pm$ 0.018} & 0.830 $\pm$ 0.027 & 0.866 $\pm$ 0.014 & 0.878 & \textit{0.032} \\
sub-29 & 0.589 $\pm$ 0.016 & 0.588 $\pm$ 0.017 & \textbf{0.659 $\pm$ 0.012} & 0.532 $\pm$ 0.049 & 0.510 & \textit{0.070} \\
sub-30 & 0.850 $\pm$ 0.010 & 0.866 $\pm$ 0.009 & 0.837 $\pm$ 0.008 & 0.844 $\pm$ 0.041 & \textbf{0.896} & \textit{0.046} \\
sub-31 & 0.774 $\pm$ 0.010 & 0.694 $\pm$ 0.017 & 0.712 $\pm$ 0.025 & 0.647 $\pm$ 0.066 & \textbf{0.792} & \textit{0.018} \\
sub-32 & 0.740 $\pm$ 0.033 & 0.808 $\pm$ 0.022 & \textbf{0.836 $\pm$ 0.013} & 0.695 $\pm$ 0.045 & 0.754 & \textit{0.097} \\
sub-33 & 0.589 $\pm$ 0.021 & 0.562 $\pm$ 0.014 & 0.596 $\pm$ 0.015 & 0.580 $\pm$ 0.015 & \textbf{0.608} & \textit{0.019} \\
sub-34 & 0.718 $\pm$ 0.018 & \textbf{0.743 $\pm$ 0.019} & 0.739 $\pm$ 0.013 & 0.712 $\pm$ 0.042 & 0.733 & \textit{0.025} \\
sub-35 & 0.894 $\pm$ 0.015 & 0.872 $\pm$ 0.018 & 0.890 $\pm$ 0.017 & 0.817 $\pm$ 0.087 & \textbf{0.903} & \textit{0.009} \\
sub-36 & 0.897 $\pm$ 0.020 & \textbf{0.907 $\pm$ 0.008} & 0.873 $\pm$ 0.011 & 0.818 $\pm$ 0.040 & 0.906 & \textit{0.010} \\
sub-37 & 0.704 $\pm$ 0.024 & \textbf{0.759 $\pm$ 0.041} & 0.756 $\pm$ 0.007 & 0.703 $\pm$ 0.033 & 0.758 & \textit{0.055} \\
sub-38 & 0.941 $\pm$ 0.007 & \textbf{0.958 $\pm$ 0.008} & 0.902 $\pm$ 0.030 & 0.941 $\pm$ 0.010 & 0.927 & \textit{0.016} \\
sub-39 & 0.619 $\pm$ 0.027 & 0.617 $\pm$ 0.017 & 0.596 $\pm$ 0.011 & 0.612 $\pm$ 0.031 & \textbf{0.671} & \textit{0.052} \\
sub-40 & \textbf{0.817 $\pm$ 0.008} & 0.804 $\pm$ 0.016 & 0.790 $\pm$ 0.033 & 0.752 $\pm$ 0.075 & 0.723 & \textit{-0.013} \\
\midrule
\textit{Mean} & 0.737 $\pm$ 0.105 & 0.732 $\pm$ 0.105 & 0.730 $\pm$ 0.106 & 0.712 $\pm$ 0.104 & \textbf{0.748 $\pm$ 0.104} & \textbf{0.037} \\
\bottomrule
\end{tabular}
\end{table}

% Table S12 — Per-subject AUROC, P300
\begin{table}
\centering
\caption{%
Per-subject AUROC (mean $\pm$ standard deviation across five training seeds, LOSO) at the three-channel montage for ERP CORE P300 ($n = 40$ subjects). Conventions as in Table~\ref{sup:tab-persubj-hri}. The mean row matches the P300 entry of main text Table~\ref{tab:auroc}.%
}
\label{sup:tab-persubj-p300}
\begin{tabular}{lcccccc}
\toprule
Subject & ERP-XTTN & EEGNet & EEG-Deformer & EPMN & xDAWN+RG & $\Delta$ \\
\midrule
sub-01 & 0.611 $\pm$ 0.020 & \textbf{0.691 $\pm$ 0.018} & 0.689 $\pm$ 0.034 & 0.632 $\pm$ 0.031 & 0.570 & \textit{0.080} \\
sub-02 & 0.719 $\pm$ 0.023 & \textbf{0.805 $\pm$ 0.015} & 0.800 $\pm$ 0.017 & 0.761 $\pm$ 0.032 & 0.716 & \textit{0.085} \\
sub-03 & 0.583 $\pm$ 0.037 & \textbf{0.737 $\pm$ 0.012} & 0.684 $\pm$ 0.023 & 0.648 $\pm$ 0.037 & 0.578 & \textit{0.154} \\
sub-04 & 0.750 $\pm$ 0.016 & \textbf{0.905 $\pm$ 0.009} & 0.826 $\pm$ 0.052 & 0.741 $\pm$ 0.054 & 0.746 & \textit{0.155} \\
sub-05 & \textbf{0.620 $\pm$ 0.016} & 0.607 $\pm$ 0.013 & 0.579 $\pm$ 0.015 & 0.552 $\pm$ 0.017 & 0.540 & \textit{-0.013} \\
sub-06 & 0.481 $\pm$ 0.039 & 0.463 $\pm$ 0.017 & \textbf{0.500 $\pm$ 0.028} & 0.454 $\pm$ 0.029 & 0.445 & \textit{0.020} \\
sub-07 & \textbf{0.730 $\pm$ 0.022} & 0.701 $\pm$ 0.018 & 0.722 $\pm$ 0.027 & 0.710 $\pm$ 0.025 & 0.608 & \textit{-0.007} \\
sub-08 & 0.635 $\pm$ 0.009 & \textbf{0.645 $\pm$ 0.014} & 0.629 $\pm$ 0.023 & 0.620 $\pm$ 0.017 & 0.583 & \textit{0.010} \\
sub-09 & 0.635 $\pm$ 0.015 & 0.676 $\pm$ 0.008 & 0.641 $\pm$ 0.021 & 0.655 $\pm$ 0.027 & \textbf{0.708} & \textit{0.073} \\
sub-10 & 0.623 $\pm$ 0.021 & 0.704 $\pm$ 0.017 & \textbf{0.706 $\pm$ 0.027} & 0.672 $\pm$ 0.047 & 0.628 & \textit{0.083} \\
sub-11 & 0.670 $\pm$ 0.012 & 0.736 $\pm$ 0.004 & \textbf{0.743 $\pm$ 0.005} & 0.685 $\pm$ 0.011 & 0.688 & \textit{0.073} \\
sub-12 & 0.682 $\pm$ 0.027 & 0.738 $\pm$ 0.007 & \textbf{0.765 $\pm$ 0.017} & 0.705 $\pm$ 0.038 & 0.611 & \textit{0.083} \\
sub-13 & 0.608 $\pm$ 0.035 & 0.651 $\pm$ 0.011 & \textbf{0.690 $\pm$ 0.031} & 0.672 $\pm$ 0.020 & 0.584 & \textit{0.082} \\
sub-14 & 0.545 $\pm$ 0.025 & 0.606 $\pm$ 0.008 & \textbf{0.608 $\pm$ 0.027} & 0.601 $\pm$ 0.039 & 0.538 & \textit{0.064} \\
sub-15 & 0.666 $\pm$ 0.018 & 0.739 $\pm$ 0.012 & 0.739 $\pm$ 0.017 & 0.692 $\pm$ 0.029 & \textbf{0.748} & \textit{0.082} \\
sub-16 & 0.769 $\pm$ 0.021 & \textbf{0.860 $\pm$ 0.013} & 0.846 $\pm$ 0.021 & 0.817 $\pm$ 0.021 & 0.788 & \textit{0.091} \\
sub-17 & 0.654 $\pm$ 0.013 & 0.720 $\pm$ 0.007 & 0.715 $\pm$ 0.032 & 0.657 $\pm$ 0.039 & \textbf{0.727} & \textit{0.073} \\
sub-18 & 0.636 $\pm$ 0.016 & \textbf{0.758 $\pm$ 0.021} & 0.716 $\pm$ 0.029 & 0.658 $\pm$ 0.022 & 0.712 & \textit{0.122} \\
sub-19 & 0.584 $\pm$ 0.021 & \textbf{0.655 $\pm$ 0.008} & 0.654 $\pm$ 0.016 & 0.642 $\pm$ 0.028 & 0.620 & \textit{0.071} \\
sub-20 & 0.716 $\pm$ 0.021 & \textbf{0.756 $\pm$ 0.009} & 0.731 $\pm$ 0.024 & 0.681 $\pm$ 0.028 & 0.686 & \textit{0.040} \\
sub-21 & 0.691 $\pm$ 0.017 & 0.723 $\pm$ 0.010 & \textbf{0.743 $\pm$ 0.032} & 0.669 $\pm$ 0.010 & 0.644 & \textit{0.052} \\
sub-22 & 0.642 $\pm$ 0.008 & \textbf{0.674 $\pm$ 0.013} & 0.638 $\pm$ 0.027 & 0.629 $\pm$ 0.020 & 0.557 & \textit{0.032} \\
sub-23 & 0.527 $\pm$ 0.012 & \textbf{0.661 $\pm$ 0.014} & 0.622 $\pm$ 0.021 & 0.602 $\pm$ 0.070 & 0.547 & \textit{0.135} \\
sub-24 & 0.657 $\pm$ 0.018 & 0.687 $\pm$ 0.010 & \textbf{0.730 $\pm$ 0.022} & 0.698 $\pm$ 0.039 & 0.699 & \textit{0.073} \\
sub-25 & 0.779 $\pm$ 0.030 & 0.796 $\pm$ 0.011 & 0.797 $\pm$ 0.021 & \textbf{0.817 $\pm$ 0.012} & 0.709 & \textit{0.038} \\
sub-26 & 0.764 $\pm$ 0.031 & 0.798 $\pm$ 0.012 & \textbf{0.818 $\pm$ 0.019} & 0.748 $\pm$ 0.040 & 0.719 & \textit{0.053} \\
sub-27 & 0.702 $\pm$ 0.028 & 0.740 $\pm$ 0.009 & 0.725 $\pm$ 0.015 & 0.667 $\pm$ 0.045 & \textbf{0.787} & \textit{0.086} \\
sub-28 & 0.734 $\pm$ 0.013 & 0.731 $\pm$ 0.013 & \textbf{0.764 $\pm$ 0.016} & 0.727 $\pm$ 0.022 & 0.680 & \textit{0.030} \\
sub-29 & 0.678 $\pm$ 0.030 & \textbf{0.715 $\pm$ 0.013} & 0.685 $\pm$ 0.023 & 0.636 $\pm$ 0.020 & 0.647 & \textit{0.037} \\
sub-30 & 0.610 $\pm$ 0.022 & 0.779 $\pm$ 0.017 & \textbf{0.783 $\pm$ 0.031} & 0.660 $\pm$ 0.036 & 0.700 & \textit{0.173} \\
sub-31 & 0.775 $\pm$ 0.024 & \textbf{0.866 $\pm$ 0.012} & 0.851 $\pm$ 0.012 & 0.805 $\pm$ 0.034 & 0.720 & \textit{0.092} \\
sub-32 & 0.707 $\pm$ 0.021 & 0.799 $\pm$ 0.013 & \textbf{0.802 $\pm$ 0.011} & 0.745 $\pm$ 0.054 & 0.735 & \textit{0.095} \\
sub-33 & 0.607 $\pm$ 0.033 & \textbf{0.804 $\pm$ 0.016} & 0.758 $\pm$ 0.025 & 0.627 $\pm$ 0.046 & 0.785 & \textit{0.196} \\
sub-34 & 0.768 $\pm$ 0.016 & \textbf{0.846 $\pm$ 0.005} & 0.815 $\pm$ 0.025 & 0.827 $\pm$ 0.030 & 0.809 & \textit{0.078} \\
sub-35 & 0.681 $\pm$ 0.013 & 0.654 $\pm$ 0.021 & 0.633 $\pm$ 0.024 & 0.629 $\pm$ 0.060 & \textbf{0.701} & \textit{0.019} \\
sub-36 & \textbf{0.613 $\pm$ 0.022} & 0.588 $\pm$ 0.013 & 0.595 $\pm$ 0.038 & 0.564 $\pm$ 0.054 & 0.502 & \textit{-0.018} \\
sub-37 & 0.730 $\pm$ 0.018 & 0.810 $\pm$ 0.023 & \textbf{0.832 $\pm$ 0.014} & 0.775 $\pm$ 0.035 & 0.764 & \textit{0.102} \\
sub-38 & 0.636 $\pm$ 0.018 & \textbf{0.793 $\pm$ 0.010} & 0.776 $\pm$ 0.027 & 0.674 $\pm$ 0.035 & 0.673 & \textit{0.157} \\
sub-39 & \textbf{0.719 $\pm$ 0.006} & 0.672 $\pm$ 0.013 & 0.688 $\pm$ 0.023 & 0.643 $\pm$ 0.053 & 0.715 & \textit{-0.003} \\
sub-40 & \textbf{0.637 $\pm$ 0.020} & 0.625 $\pm$ 0.018 & 0.624 $\pm$ 0.025 & 0.566 $\pm$ 0.024 & 0.485 & \textit{-0.012} \\
\midrule
\textit{Mean} & 0.664 $\pm$ 0.071 & \textbf{0.723 $\pm$ 0.087} & 0.717 $\pm$ 0.082 & 0.674 $\pm$ 0.077 & 0.660 $\pm$ 0.091 & \textbf{0.071} \\
\bottomrule
\end{tabular}
\end{table}

% Table S13 — Per-subject AUROC, N2pc
\begin{table}
\centering
\caption{%
Per-subject AUROC (mean $\pm$ standard deviation across five training seeds, LOSO) at the three-channel montage for ERP CORE N2pc ($n = 40$ subjects). Conventions as in Table~\ref{sup:tab-persubj-hri}. The mean row matches the N2pc entry of main text Table~\ref{tab:auroc}.%
}
\label{sup:tab-persubj-n2pc}
\begin{tabular}{lcccccc}
\toprule
Subject & ERP-XTTN & EEGNet & EEG-Deformer & EPMN & xDAWN+RG & $\Delta$ \\
\midrule
sub-01 & 0.640 $\pm$ 0.007 & \textbf{0.645 $\pm$ 0.014} & 0.645 $\pm$ 0.009 & 0.615 $\pm$ 0.012 & 0.605 & \textit{0.005} \\
sub-02 & 0.617 $\pm$ 0.018 & \textbf{0.645 $\pm$ 0.013} & 0.610 $\pm$ 0.009 & 0.611 $\pm$ 0.018 & 0.626 & \textit{0.028} \\
sub-03 & \textbf{0.568 $\pm$ 0.003} & 0.555 $\pm$ 0.010 & 0.541 $\pm$ 0.007 & 0.554 $\pm$ 0.018 & 0.549 & \textit{-0.013} \\
sub-04 & 0.783 $\pm$ 0.021 & 0.818 $\pm$ 0.012 & \textbf{0.834 $\pm$ 0.020} & 0.773 $\pm$ 0.028 & 0.817 & \textit{0.051} \\
sub-05 & 0.756 $\pm$ 0.008 & 0.743 $\pm$ 0.011 & \textbf{0.778 $\pm$ 0.013} & 0.724 $\pm$ 0.004 & 0.766 & \textit{0.022} \\
sub-06 & 0.845 $\pm$ 0.009 & \textbf{0.897 $\pm$ 0.003} & 0.871 $\pm$ 0.010 & 0.885 $\pm$ 0.011 & 0.886 & \textit{0.053} \\
sub-07 & 0.681 $\pm$ 0.012 & \textbf{0.713 $\pm$ 0.014} & 0.701 $\pm$ 0.013 & 0.677 $\pm$ 0.013 & 0.690 & \textit{0.031} \\
sub-08 & 0.698 $\pm$ 0.014 & \textbf{0.702 $\pm$ 0.013} & 0.698 $\pm$ 0.032 & 0.695 $\pm$ 0.012 & 0.688 & \textit{0.004} \\
sub-09 & 0.682 $\pm$ 0.011 & \textbf{0.724 $\pm$ 0.011} & 0.688 $\pm$ 0.021 & 0.682 $\pm$ 0.026 & 0.665 & \textit{0.042} \\
sub-10 & \textbf{0.540 $\pm$ 0.006} & 0.516 $\pm$ 0.005 & 0.507 $\pm$ 0.010 & 0.530 $\pm$ 0.016 & 0.540 & \textit{-0.001} \\
sub-11 & 0.639 $\pm$ 0.012 & 0.664 $\pm$ 0.010 & 0.629 $\pm$ 0.029 & 0.651 $\pm$ 0.011 & \textbf{0.684} & \textit{0.045} \\
sub-12 & 0.482 $\pm$ 0.010 & 0.484 $\pm$ 0.008 & \textbf{0.508 $\pm$ 0.008} & 0.482 $\pm$ 0.029 & 0.449 & \textit{0.026} \\
sub-13 & 0.674 $\pm$ 0.021 & 0.700 $\pm$ 0.006 & \textbf{0.703 $\pm$ 0.017} & 0.640 $\pm$ 0.011 & 0.671 & \textit{0.029} \\
sub-14 & 0.631 $\pm$ 0.016 & 0.654 $\pm$ 0.012 & 0.596 $\pm$ 0.032 & 0.657 $\pm$ 0.019 & \textbf{0.670} & \textit{0.039} \\
sub-15 & 0.671 $\pm$ 0.011 & \textbf{0.721 $\pm$ 0.008} & 0.707 $\pm$ 0.015 & 0.715 $\pm$ 0.022 & 0.717 & \textit{0.050} \\
sub-16 & 0.840 $\pm$ 0.009 & \textbf{0.844 $\pm$ 0.017} & 0.835 $\pm$ 0.018 & 0.794 $\pm$ 0.022 & 0.832 & \textit{0.004} \\
sub-17 & 0.636 $\pm$ 0.010 & 0.601 $\pm$ 0.008 & 0.635 $\pm$ 0.013 & 0.589 $\pm$ 0.012 & \textbf{0.652} & \textit{0.016} \\
sub-18 & 0.712 $\pm$ 0.011 & 0.732 $\pm$ 0.019 & 0.735 $\pm$ 0.009 & 0.716 $\pm$ 0.020 & \textbf{0.737} & \textit{0.025} \\
sub-19 & 0.671 $\pm$ 0.012 & 0.681 $\pm$ 0.016 & 0.689 $\pm$ 0.013 & 0.686 $\pm$ 0.013 & \textbf{0.717} & \textit{0.047} \\
sub-20 & 0.517 $\pm$ 0.009 & 0.525 $\pm$ 0.002 & 0.536 $\pm$ 0.021 & \textbf{0.539 $\pm$ 0.017} & 0.501 & \textit{0.022} \\
sub-21 & 0.588 $\pm$ 0.008 & 0.587 $\pm$ 0.009 & 0.591 $\pm$ 0.011 & 0.608 $\pm$ 0.010 & \textbf{0.627} & \textit{0.039} \\
sub-22 & 0.669 $\pm$ 0.009 & 0.716 $\pm$ 0.005 & \textbf{0.725 $\pm$ 0.009} & 0.697 $\pm$ 0.011 & 0.650 & \textit{0.057} \\
sub-23 & \textbf{0.618 $\pm$ 0.007} & 0.567 $\pm$ 0.005 & 0.558 $\pm$ 0.007 & 0.567 $\pm$ 0.018 & 0.585 & \textit{-0.034} \\
sub-24 & 0.694 $\pm$ 0.015 & 0.706 $\pm$ 0.005 & 0.708 $\pm$ 0.024 & 0.690 $\pm$ 0.008 & \textbf{0.750} & \textit{0.056} \\
sub-25 & \textbf{0.635 $\pm$ 0.006} & 0.631 $\pm$ 0.016 & 0.601 $\pm$ 0.010 & 0.613 $\pm$ 0.014 & 0.619 & \textit{-0.004} \\
sub-26 & 0.521 $\pm$ 0.023 & 0.516 $\pm$ 0.005 & 0.510 $\pm$ 0.011 & \textbf{0.532 $\pm$ 0.013} & 0.516 & \textit{0.010} \\
sub-27 & 0.589 $\pm$ 0.007 & 0.587 $\pm$ 0.013 & 0.580 $\pm$ 0.019 & 0.564 $\pm$ 0.022 & \textbf{0.614} & \textit{0.025} \\
sub-28 & 0.613 $\pm$ 0.008 & 0.610 $\pm$ 0.012 & \textbf{0.624 $\pm$ 0.023} & 0.594 $\pm$ 0.022 & 0.608 & \textit{0.012} \\
sub-29 & 0.515 $\pm$ 0.012 & 0.510 $\pm$ 0.010 & 0.513 $\pm$ 0.005 & \textbf{0.528 $\pm$ 0.013} & 0.517 & \textit{0.014} \\
sub-30 & 0.547 $\pm$ 0.008 & \textbf{0.566 $\pm$ 0.011} & 0.534 $\pm$ 0.018 & 0.549 $\pm$ 0.026 & 0.539 & \textit{0.019} \\
sub-31 & \textbf{0.592 $\pm$ 0.012} & 0.591 $\pm$ 0.007 & 0.588 $\pm$ 0.013 & 0.559 $\pm$ 0.011 & 0.557 & \textit{-0.001} \\
sub-32 & 0.589 $\pm$ 0.009 & 0.641 $\pm$ 0.004 & 0.619 $\pm$ 0.018 & 0.596 $\pm$ 0.010 & \textbf{0.655} & \textit{0.066} \\
sub-33 & 0.672 $\pm$ 0.006 & 0.706 $\pm$ 0.009 & 0.679 $\pm$ 0.013 & 0.653 $\pm$ 0.016 & \textbf{0.740} & \textit{0.068} \\
sub-34 & 0.624 $\pm$ 0.010 & \textbf{0.662 $\pm$ 0.013} & 0.662 $\pm$ 0.017 & 0.635 $\pm$ 0.030 & 0.608 & \textit{0.038} \\
sub-35 & 0.616 $\pm$ 0.010 & \textbf{0.661 $\pm$ 0.008} & 0.645 $\pm$ 0.018 & 0.613 $\pm$ 0.030 & 0.655 & \textit{0.045} \\
sub-36 & \textbf{0.636 $\pm$ 0.008} & 0.608 $\pm$ 0.009 & 0.595 $\pm$ 0.019 & 0.610 $\pm$ 0.010 & 0.625 & \textit{-0.011} \\
sub-37 & \textbf{0.701 $\pm$ 0.006} & 0.684 $\pm$ 0.005 & 0.652 $\pm$ 0.030 & 0.694 $\pm$ 0.008 & 0.688 & \textit{-0.006} \\
sub-38 & 0.681 $\pm$ 0.008 & 0.675 $\pm$ 0.006 & 0.680 $\pm$ 0.022 & 0.664 $\pm$ 0.023 & \textbf{0.700} & \textit{0.020} \\
sub-39 & 0.621 $\pm$ 0.005 & 0.600 $\pm$ 0.027 & 0.589 $\pm$ 0.010 & 0.638 $\pm$ 0.017 & \textbf{0.679} & \textit{0.058} \\
sub-40 & \textbf{0.563 $\pm$ 0.014} & 0.546 $\pm$ 0.016 & 0.522 $\pm$ 0.014 & 0.547 $\pm$ 0.011 & 0.555 & \textit{-0.008} \\
\midrule
\textit{Mean} & 0.639 $\pm$ 0.081 & 0.648 $\pm$ 0.092 & 0.641 $\pm$ 0.092 & 0.634 $\pm$ 0.082 & \textbf{0.649 $\pm$ 0.093} & \textbf{0.025} \\
\bottomrule
\end{tabular}
\end{table}

% Table S14 — Per-subject AUROC, MMN
\begin{table}
\centering
\caption{%
Per-subject AUROC (mean $\pm$ standard deviation across five training seeds, LOSO) at the three-channel montage for ERP CORE MMN ($n = 40$ subjects). Conventions as in Table~\ref{sup:tab-persubj-hri}. The mean row matches the MMN entry of main text Table~\ref{tab:auroc}.%
}
\label{sup:tab-persubj-mmn}
\begin{tabular}{lcccccc}
\toprule
Subject & ERP-XTTN & EEGNet & EEG-Deformer & EPMN & xDAWN+RG & $\Delta$ \\
\midrule
sub-01 & 0.676 $\pm$ 0.006 & 0.687 $\pm$ 0.007 & \textbf{0.689 $\pm$ 0.005} & 0.646 $\pm$ 0.022 & 0.665 & \textit{0.013} \\
sub-02 & \textbf{0.573 $\pm$ 0.003} & 0.566 $\pm$ 0.004 & 0.557 $\pm$ 0.010 & 0.570 $\pm$ 0.018 & 0.568 & \textit{-0.003} \\
sub-03 & 0.560 $\pm$ 0.003 & \textbf{0.588 $\pm$ 0.005} & 0.576 $\pm$ 0.009 & 0.557 $\pm$ 0.030 & 0.583 & \textit{0.028} \\
sub-04 & 0.499 $\pm$ 0.004 & \textbf{0.520 $\pm$ 0.007} & 0.519 $\pm$ 0.005 & 0.503 $\pm$ 0.008 & 0.504 & \textit{0.021} \\
sub-05 & \textbf{0.624 $\pm$ 0.006} & 0.616 $\pm$ 0.012 & 0.611 $\pm$ 0.021 & 0.570 $\pm$ 0.020 & 0.609 & \textit{-0.008} \\
sub-06 & 0.555 $\pm$ 0.005 & 0.556 $\pm$ 0.008 & 0.555 $\pm$ 0.010 & 0.544 $\pm$ 0.021 & \textbf{0.557} & \textit{0.002} \\
sub-07 & 0.620 $\pm$ 0.004 & 0.626 $\pm$ 0.005 & 0.626 $\pm$ 0.009 & 0.598 $\pm$ 0.024 & \textbf{0.642} & \textit{0.022} \\
sub-08 & 0.618 $\pm$ 0.002 & 0.619 $\pm$ 0.006 & \textbf{0.633 $\pm$ 0.014} & 0.588 $\pm$ 0.014 & 0.623 & \textit{0.016} \\
sub-09 & 0.615 $\pm$ 0.006 & 0.624 $\pm$ 0.008 & \textbf{0.631 $\pm$ 0.014} & 0.603 $\pm$ 0.019 & 0.625 & \textit{0.016} \\
sub-10 & 0.564 $\pm$ 0.004 & 0.568 $\pm$ 0.005 & \textbf{0.570 $\pm$ 0.006} & 0.543 $\pm$ 0.025 & 0.562 & \textit{0.006} \\
sub-11 & 0.583 $\pm$ 0.007 & 0.585 $\pm$ 0.009 & 0.583 $\pm$ 0.014 & 0.551 $\pm$ 0.023 & \textbf{0.586} & \textit{0.003} \\
sub-12 & 0.569 $\pm$ 0.007 & 0.569 $\pm$ 0.004 & 0.569 $\pm$ 0.014 & 0.540 $\pm$ 0.048 & \textbf{0.570} & \textit{0.002} \\
sub-13 & 0.522 $\pm$ 0.006 & 0.503 $\pm$ 0.005 & 0.511 $\pm$ 0.009 & 0.507 $\pm$ 0.019 & \textbf{0.530} & \textit{0.007} \\
sub-14 & 0.563 $\pm$ 0.001 & 0.561 $\pm$ 0.006 & 0.568 $\pm$ 0.008 & \textbf{0.568 $\pm$ 0.026} & 0.567 & \textit{0.005} \\
sub-15 & 0.579 $\pm$ 0.005 & 0.598 $\pm$ 0.006 & 0.594 $\pm$ 0.006 & 0.552 $\pm$ 0.059 & \textbf{0.608} & \textit{0.030} \\
sub-16 & 0.625 $\pm$ 0.007 & 0.616 $\pm$ 0.004 & 0.624 $\pm$ 0.008 & 0.575 $\pm$ 0.040 & \textbf{0.627} & \textit{0.002} \\
sub-17 & 0.507 $\pm$ 0.004 & 0.511 $\pm$ 0.009 & 0.496 $\pm$ 0.026 & 0.509 $\pm$ 0.021 & \textbf{0.514} & \textit{0.007} \\
sub-18 & 0.544 $\pm$ 0.005 & 0.531 $\pm$ 0.004 & 0.538 $\pm$ 0.005 & 0.513 $\pm$ 0.006 & \textbf{0.551} & \textit{0.007} \\
sub-19 & 0.600 $\pm$ 0.008 & 0.591 $\pm$ 0.008 & 0.594 $\pm$ 0.007 & 0.548 $\pm$ 0.016 & \textbf{0.616} & \textit{0.016} \\
sub-20 & 0.648 $\pm$ 0.004 & 0.648 $\pm$ 0.011 & 0.654 $\pm$ 0.007 & 0.609 $\pm$ 0.037 & \textbf{0.655} & \textit{0.008} \\
sub-21 & 0.528 $\pm$ 0.006 & 0.538 $\pm$ 0.007 & 0.529 $\pm$ 0.011 & 0.524 $\pm$ 0.024 & \textbf{0.547} & \textit{0.019} \\
sub-22 & 0.659 $\pm$ 0.003 & 0.665 $\pm$ 0.008 & \textbf{0.668 $\pm$ 0.006} & 0.612 $\pm$ 0.026 & 0.653 & \textit{0.009} \\
sub-23 & 0.544 $\pm$ 0.005 & \textbf{0.555 $\pm$ 0.010} & 0.550 $\pm$ 0.005 & 0.549 $\pm$ 0.009 & 0.551 & \textit{0.011} \\
sub-24 & 0.600 $\pm$ 0.004 & 0.601 $\pm$ 0.004 & 0.602 $\pm$ 0.010 & 0.586 $\pm$ 0.019 & \textbf{0.604} & \textit{0.004} \\
sub-25 & 0.549 $\pm$ 0.004 & 0.554 $\pm$ 0.005 & 0.563 $\pm$ 0.002 & 0.526 $\pm$ 0.017 & \textbf{0.577} & \textit{0.027} \\
sub-26 & \textbf{0.620 $\pm$ 0.005} & 0.604 $\pm$ 0.004 & 0.601 $\pm$ 0.005 & 0.567 $\pm$ 0.022 & 0.616 & \textit{-0.004} \\
sub-27 & 0.621 $\pm$ 0.005 & \textbf{0.646 $\pm$ 0.006} & 0.642 $\pm$ 0.007 & 0.609 $\pm$ 0.019 & 0.635 & \textit{0.025} \\
sub-28 & 0.555 $\pm$ 0.006 & 0.577 $\pm$ 0.003 & 0.562 $\pm$ 0.006 & 0.541 $\pm$ 0.019 & \textbf{0.582} & \textit{0.027} \\
sub-29 & 0.557 $\pm$ 0.002 & \textbf{0.563 $\pm$ 0.005} & 0.553 $\pm$ 0.007 & 0.527 $\pm$ 0.035 & 0.558 & \textit{0.006} \\
sub-30 & 0.534 $\pm$ 0.005 & \textbf{0.555 $\pm$ 0.007} & 0.546 $\pm$ 0.018 & 0.533 $\pm$ 0.017 & 0.545 & \textit{0.021} \\
sub-31 & 0.612 $\pm$ 0.004 & 0.621 $\pm$ 0.008 & \textbf{0.624 $\pm$ 0.015} & 0.571 $\pm$ 0.023 & 0.621 & \textit{0.012} \\
sub-32 & 0.559 $\pm$ 0.003 & 0.558 $\pm$ 0.009 & 0.556 $\pm$ 0.013 & 0.555 $\pm$ 0.018 & \textbf{0.570} & \textit{0.011} \\
sub-33 & 0.649 $\pm$ 0.004 & 0.634 $\pm$ 0.012 & 0.645 $\pm$ 0.011 & 0.619 $\pm$ 0.018 & \textbf{0.652} & \textit{0.002} \\
sub-34 & \textbf{0.573 $\pm$ 0.004} & 0.547 $\pm$ 0.008 & 0.537 $\pm$ 0.013 & 0.526 $\pm$ 0.011 & 0.545 & \textit{-0.026} \\
sub-35 & 0.594 $\pm$ 0.006 & 0.593 $\pm$ 0.007 & 0.592 $\pm$ 0.024 & 0.571 $\pm$ 0.028 & \textbf{0.599} & \textit{0.005} \\
sub-36 & 0.507 $\pm$ 0.002 & 0.514 $\pm$ 0.005 & 0.501 $\pm$ 0.008 & 0.506 $\pm$ 0.015 & \textbf{0.516} & \textit{0.010} \\
sub-37 & \textbf{0.642 $\pm$ 0.006} & 0.633 $\pm$ 0.007 & 0.631 $\pm$ 0.015 & 0.583 $\pm$ 0.027 & 0.637 & \textit{-0.005} \\
sub-38 & 0.664 $\pm$ 0.003 & \textbf{0.666 $\pm$ 0.014} & 0.646 $\pm$ 0.015 & 0.598 $\pm$ 0.019 & 0.663 & \textit{0.003} \\
sub-39 & 0.702 $\pm$ 0.002 & 0.707 $\pm$ 0.006 & 0.695 $\pm$ 0.016 & 0.651 $\pm$ 0.020 & \textbf{0.718} & \textit{0.017} \\
sub-40 & 0.553 $\pm$ 0.003 & 0.560 $\pm$ 0.006 & 0.555 $\pm$ 0.008 & 0.547 $\pm$ 0.011 & \textbf{0.565} & \textit{0.012} \\
\midrule
\textit{Mean} & 0.587 $\pm$ 0.049 & 0.589 $\pm$ 0.049 & 0.587 $\pm$ 0.050 & 0.562 $\pm$ 0.038 & \textbf{0.593 $\pm$ 0.048} & \textbf{0.010} \\
\bottomrule
\end{tabular}
\end{table}

% Table S15 — Per-subject AUROC, N400
\begin{table}
\centering
\caption{%
Per-subject AUROC (mean $\pm$ standard deviation across five training seeds, LOSO) at the three-channel montage for ERP CORE N400 ($n = 40$ subjects). Conventions as in Table~\ref{sup:tab-persubj-hri}. The mean row matches the N400 entry of main text Table~\ref{tab:auroc}.%
}
\label{sup:tab-persubj-n400}
\begin{tabular}{lcccccc}
\toprule
Subject & ERP-XTTN & EEGNet & EEG-Deformer & EPMN & xDAWN+RG & $\Delta$ \\
\midrule
sub-01 & 0.531 $\pm$ 0.024 & 0.531 $\pm$ 0.020 & 0.541 $\pm$ 0.015 & 0.569 $\pm$ 0.023 & \textbf{0.597} & \textit{0.066} \\
sub-02 & 0.601 $\pm$ 0.023 & \textbf{0.669 $\pm$ 0.013} & 0.658 $\pm$ 0.013 & 0.620 $\pm$ 0.019 & 0.611 & \textit{0.068} \\
sub-03 & 0.564 $\pm$ 0.021 & \textbf{0.623 $\pm$ 0.015} & 0.603 $\pm$ 0.037 & 0.603 $\pm$ 0.059 & 0.608 & \textit{0.059} \\
sub-04 & 0.668 $\pm$ 0.032 & \textbf{0.719 $\pm$ 0.008} & 0.650 $\pm$ 0.105 & 0.602 $\pm$ 0.046 & 0.650 & \textit{0.051} \\
sub-05 & 0.468 $\pm$ 0.025 & 0.493 $\pm$ 0.013 & 0.494 $\pm$ 0.036 & 0.495 $\pm$ 0.043 & \textbf{0.549} & \textit{0.081} \\
sub-06 & 0.572 $\pm$ 0.022 & 0.578 $\pm$ 0.029 & \textbf{0.623 $\pm$ 0.025} & 0.615 $\pm$ 0.040 & 0.599 & \textit{0.051} \\
sub-07 & 0.626 $\pm$ 0.014 & \textbf{0.643 $\pm$ 0.013} & 0.623 $\pm$ 0.037 & 0.566 $\pm$ 0.065 & 0.620 & \textit{0.017} \\
sub-08 & 0.512 $\pm$ 0.024 & 0.598 $\pm$ 0.021 & 0.596 $\pm$ 0.027 & \textbf{0.606 $\pm$ 0.021} & 0.519 & \textit{0.094} \\
sub-09 & 0.503 $\pm$ 0.007 & \textbf{0.560 $\pm$ 0.012} & 0.528 $\pm$ 0.040 & 0.514 $\pm$ 0.046 & 0.549 & \textit{0.057} \\
sub-10 & 0.568 $\pm$ 0.016 & \textbf{0.570 $\pm$ 0.020} & 0.566 $\pm$ 0.025 & 0.534 $\pm$ 0.037 & 0.547 & \textit{0.002} \\
sub-11 & 0.522 $\pm$ 0.025 & 0.475 $\pm$ 0.016 & 0.489 $\pm$ 0.007 & 0.526 $\pm$ 0.023 & \textbf{0.544} & \textit{0.021} \\
sub-12 & 0.646 $\pm$ 0.030 & \textbf{0.677 $\pm$ 0.020} & 0.657 $\pm$ 0.027 & 0.630 $\pm$ 0.067 & 0.666 & \textit{0.031} \\
sub-13 & 0.445 $\pm$ 0.021 & 0.513 $\pm$ 0.005 & 0.513 $\pm$ 0.020 & \textbf{0.514 $\pm$ 0.039} & 0.514 & \textit{0.070} \\
sub-14 & \textbf{0.499 $\pm$ 0.025} & 0.469 $\pm$ 0.011 & 0.479 $\pm$ 0.020 & 0.478 $\pm$ 0.033 & 0.494 & \textit{-0.004} \\
sub-15 & 0.616 $\pm$ 0.016 & 0.629 $\pm$ 0.017 & 0.656 $\pm$ 0.054 & 0.610 $\pm$ 0.077 & \textbf{0.656} & \textit{0.040} \\
sub-16 & 0.713 $\pm$ 0.013 & 0.704 $\pm$ 0.012 & 0.641 $\pm$ 0.032 & 0.674 $\pm$ 0.031 & \textbf{0.742} & \textit{0.029} \\
sub-17 & 0.480 $\pm$ 0.028 & 0.515 $\pm$ 0.018 & 0.480 $\pm$ 0.034 & 0.487 $\pm$ 0.012 & \textbf{0.560} & \textit{0.080} \\
sub-18 & \textbf{0.459 $\pm$ 0.046} & 0.387 $\pm$ 0.024 & 0.426 $\pm$ 0.050 & 0.431 $\pm$ 0.071 & 0.416 & \textit{-0.028} \\
sub-19 & 0.754 $\pm$ 0.028 & \textbf{0.798 $\pm$ 0.017} & 0.795 $\pm$ 0.025 & 0.748 $\pm$ 0.060 & 0.774 & \textit{0.045} \\
sub-20 & 0.623 $\pm$ 0.031 & \textbf{0.699 $\pm$ 0.014} & 0.667 $\pm$ 0.034 & 0.684 $\pm$ 0.050 & 0.693 & \textit{0.077} \\
sub-21 & 0.479 $\pm$ 0.021 & 0.434 $\pm$ 0.014 & 0.465 $\pm$ 0.030 & \textbf{0.487 $\pm$ 0.032} & 0.436 & \textit{0.008} \\
sub-22 & 0.567 $\pm$ 0.024 & 0.567 $\pm$ 0.037 & 0.587 $\pm$ 0.051 & 0.588 $\pm$ 0.021 & \textbf{0.610} & \textit{0.043} \\
sub-23 & 0.560 $\pm$ 0.013 & 0.548 $\pm$ 0.018 & 0.555 $\pm$ 0.041 & 0.521 $\pm$ 0.024 & \textbf{0.562} & \textit{0.002} \\
sub-24 & 0.579 $\pm$ 0.021 & 0.587 $\pm$ 0.015 & 0.557 $\pm$ 0.033 & \textbf{0.604 $\pm$ 0.055} & 0.603 & \textit{0.025} \\
sub-25 & 0.496 $\pm$ 0.011 & 0.511 $\pm$ 0.012 & 0.507 $\pm$ 0.008 & \textbf{0.545 $\pm$ 0.046} & 0.527 & \textit{0.049} \\
sub-26 & 0.683 $\pm$ 0.022 & 0.686 $\pm$ 0.016 & 0.676 $\pm$ 0.037 & \textbf{0.732 $\pm$ 0.023} & 0.687 & \textit{0.049} \\
sub-27 & 0.551 $\pm$ 0.024 & \textbf{0.560 $\pm$ 0.015} & 0.545 $\pm$ 0.027 & 0.546 $\pm$ 0.023 & 0.519 & \textit{0.010} \\
sub-28 & 0.465 $\pm$ 0.021 & 0.491 $\pm$ 0.013 & 0.490 $\pm$ 0.032 & 0.471 $\pm$ 0.044 & \textbf{0.491} & \textit{0.027} \\
sub-29 & 0.567 $\pm$ 0.027 & 0.598 $\pm$ 0.008 & 0.585 $\pm$ 0.037 & \textbf{0.603 $\pm$ 0.030} & 0.589 & \textit{0.037} \\
sub-30 & 0.516 $\pm$ 0.021 & 0.550 $\pm$ 0.012 & 0.551 $\pm$ 0.033 & \textbf{0.574 $\pm$ 0.039} & 0.567 & \textit{0.059} \\
sub-31 & 0.706 $\pm$ 0.021 & 0.744 $\pm$ 0.016 & \textbf{0.752 $\pm$ 0.019} & 0.716 $\pm$ 0.021 & 0.743 & \textit{0.046} \\
sub-32 & 0.452 $\pm$ 0.015 & 0.503 $\pm$ 0.005 & \textbf{0.505 $\pm$ 0.024} & 0.473 $\pm$ 0.023 & 0.432 & \textit{0.053} \\
sub-33 & 0.594 $\pm$ 0.031 & 0.636 $\pm$ 0.027 & 0.610 $\pm$ 0.011 & \textbf{0.639 $\pm$ 0.016} & 0.574 & \textit{0.046} \\
sub-34 & 0.567 $\pm$ 0.016 & 0.572 $\pm$ 0.020 & \textbf{0.597 $\pm$ 0.029} & 0.531 $\pm$ 0.049 & 0.520 & \textit{0.030} \\
sub-35 & 0.542 $\pm$ 0.030 & 0.550 $\pm$ 0.035 & 0.528 $\pm$ 0.071 & 0.504 $\pm$ 0.043 & \textbf{0.560} & \textit{0.017} \\
sub-36 & 0.563 $\pm$ 0.019 & 0.554 $\pm$ 0.015 & 0.511 $\pm$ 0.036 & 0.468 $\pm$ 0.019 & \textbf{0.575} & \textit{0.012} \\
sub-37 & 0.531 $\pm$ 0.021 & \textbf{0.585 $\pm$ 0.019} & 0.567 $\pm$ 0.020 & 0.552 $\pm$ 0.023 & 0.540 & \textit{0.053} \\
sub-38 & \textbf{0.676 $\pm$ 0.016} & 0.654 $\pm$ 0.033 & 0.667 $\pm$ 0.026 & 0.672 $\pm$ 0.029 & 0.642 & \textit{-0.003} \\
sub-39 & 0.597 $\pm$ 0.017 & \textbf{0.646 $\pm$ 0.022} & 0.596 $\pm$ 0.019 & 0.624 $\pm$ 0.036 & 0.576 & \textit{0.049} \\
sub-40 & 0.539 $\pm$ 0.013 & \textbf{0.598 $\pm$ 0.027} & 0.587 $\pm$ 0.022 & 0.574 $\pm$ 0.037 & 0.527 & \textit{0.058} \\
\midrule
\textit{Mean} & 0.565 $\pm$ 0.077 & \textbf{0.586 $\pm$ 0.087} & 0.578 $\pm$ 0.079 & 0.573 $\pm$ 0.077 & 0.580 $\pm$ 0.081 & \textbf{0.039} \\
\bottomrule
\end{tabular}
\end{table}

% Table S16 — Wilcoxon signed-rank test statistics
\begin{table}
\centering
\setlength{\tabcolsep}{4pt}
\caption{%
Paired Wilcoxon signed-rank tests for ERP-XTTN versus each baseline at the three-channel montage (companion to main text Figure~\ref{fig:fig_paired_heatmap}). Per-subject AUROCs are averaged over five training seeds before pairing (xDAWN+RG: single deterministic run). $\hat{\Delta}$: Hodges-Lehmann estimate of the paired difference (ERP-XTTN minus baseline; negative = ERP-XTTN lower). 95\% CI: exact distribution-free signed-rank confidence interval. $p$: two-sided Wilcoxon signed-rank raw $p$-value. $q$: Benjamini-Hochberg adjusted $p$-value controlling the false-discovery rate across all 36 comparisons (9 datasets $\times$ 4 baselines). $r_{rb}$: rank-biserial correlation (matched-pairs), computed as $r_{rb} = 2W^{+} / [n(n+1)/2] - 1$ where $W^{+}$ is the sum of positive-difference ranks; values near $+1$/$-1$ indicate ERP-XTTN consistently higher/lower. \textbf{Bold} $q$ marks comparisons significant at $q < 0.05$. $^\dagger$BNCI ($n=6$): the exact signed-rank test floors at $p = 0.031$ and cannot survive FDR correction regardless of effect size; results are reported descriptively (CI omitted). Datasets ordered by mean AUROC across all five methods, descending. Abbreviations as in Table~\ref{sup:tab-fullmontage-auroc}.%
}
\label{sup:tab-wilcoxon}
\begin{tabular}{llccccc}
\toprule
Dataset & Baseline & $\hat{\Delta}$ & 95\% CI & $p$ & $q$ & $r_{rb}$ \\
\midrule
HRI ($n{=}11$) & EEGNet & $-0.021$ & [$-0.040$, $-0.001$] & 0.024 & 0.055 & $-0.758$ \\
 & EEG-Deformer & $-0.014$ & [$-0.034$, $+0.004$] & 0.147 & 0.197 & $-0.515$ \\
 & EPMN & $+0.006$ & [$-0.002$, $+0.013$] & 0.123 & 0.180 & $+0.545$ \\
 & xDAWN+RG & $+0.000$ & [$-0.014$, $+0.015$] & 1.000 & 1.000 & $+0.000$ \\
\midrule
ERN ($n{=}40$) & EEGNet & $-0.041$ & [$-0.051$, $-0.029$] & $<$0.001 & \textbf{$<$0.001} & $-0.910$ \\
 & EEG-Deformer & $-0.010$ & [$-0.022$, $+0.001$] & 0.062 & 0.112 & $-0.339$ \\
 & EPMN & $+0.031$ & [$+0.017$, $+0.044$] & $<$0.001 & \textbf{$<$0.001} & $+0.698$ \\
 & xDAWN+RG & $+0.027$ & [$+0.011$, $+0.046$] & $<$0.001 & \textbf{0.003} & $+0.580$ \\
\midrule
LRP ($n{=}40$) & EEGNet & $-0.029$ & [$-0.037$, $-0.023$] & $<$0.001 & \textbf{$<$0.001} & $-0.949$ \\
 & EEG-Deformer & $-0.025$ & [$-0.034$, $-0.017$] & $<$0.001 & \textbf{$<$0.001} & $-0.807$ \\
 & EPMN & $-0.008$ & [$-0.016$, $-0.001$] & 0.015 & \textbf{0.039} & $-0.437$ \\
 & xDAWN+RG & $-0.034$ & [$-0.042$, $-0.026$] & $<$0.001 & \textbf{$<$0.001} & $-0.971$ \\
\midrule
BNCI ($n{=}6$)$^\dagger$ & EEGNet & $-0.022$ & -- & 0.219 & 0.272 & $-0.619$ \\
 & EEG-Deformer & $-0.015$ & -- & 0.031 & 0.062 & $-1.000$ \\
 & EPMN & $+0.037$ & -- & 0.031 & 0.062 & $+1.000$ \\
 & xDAWN+RG & $-0.014$ & -- & 0.438 & 0.492 & $-0.429$ \\
\midrule
N170 ($n{=}40$) & EEGNet & $+0.007$ & [$-0.008$, $+0.021$] & 0.301 & 0.361 & $+0.190$ \\
 & EEG-Deformer & $+0.008$ & [$-0.008$, $+0.028$] & 0.327 & 0.380 & $+0.180$ \\
 & EPMN & $+0.025$ & [$+0.011$, $+0.039$] & $<$0.001 & \textbf{0.003} & $+0.580$ \\
 & xDAWN+RG & $-0.012$ & [$-0.028$, $+0.003$] & 0.135 & 0.187 & $-0.273$ \\
\midrule
P300 ($n{=}40$) & EEGNet & $-0.057$ & [$-0.076$, $-0.036$] & $<$0.001 & \textbf{$<$0.001} & $-0.861$ \\
 & EEG-Deformer & $-0.051$ & [$-0.073$, $-0.035$] & $<$0.001 & \textbf{$<$0.001} & $-0.839$ \\
 & EPMN & $-0.011$ & [$-0.026$, $+0.003$] & 0.121 & 0.180 & $-0.283$ \\
 & xDAWN+RG & $+0.005$ & [$-0.017$, $+0.025$] & 0.646 & 0.684 & $+0.085$ \\
\midrule
N2pc ($n{=}40$) & EEGNet & $-0.009$ & [$-0.018$, $0.000$] & 0.050 & 0.094 & $-0.356$ \\
 & EEG-Deformer & $-0.002$ & [$-0.011$, $+0.007$] & 0.745 & 0.766 & $-0.061$ \\
 & EPMN & $+0.005$ & [$-0.003$, $+0.013$] & 0.179 & 0.230 & $+0.246$ \\
 & xDAWN+RG & $-0.010$ & [$-0.019$, $+0.003$] & 0.118 & 0.180 & $-0.285$ \\
\midrule
MMN ($n{=}40$) & EEGNet & $-0.003$ & [$-0.006$, $+0.001$] & 0.112 & 0.180 & $-0.290$ \\
 & EEG-Deformer & $-0.001$ & [$-0.005$, $+0.003$] & 0.476 & 0.519 & $-0.132$ \\
 & EPMN & $+0.024$ & [$+0.017$, $+0.031$] & $<$0.001 & \textbf{$<$0.001} & $+0.929$ \\
 & xDAWN+RG & $-0.006$ & [$-0.010$, $-0.003$] & $<$0.001 & \textbf{0.002} & $-0.620$ \\
\midrule
N400 ($n{=}40$) & EEGNet & $-0.022$ & [$-0.033$, $-0.010$] & $<$0.001 & \textbf{0.002} & $-0.612$ \\
 & EEG-Deformer & $-0.014$ & [$-0.025$, $-0.003$] & 0.018 & \textbf{0.043} & $-0.427$ \\
 & EPMN & $-0.010$ & [$-0.023$, $+0.004$] & 0.125 & 0.180 & $-0.280$ \\
 & xDAWN+RG & $-0.015$ & [$-0.026$, $-0.003$] & 0.014 & \textbf{0.038} & $-0.444$ \\
\bottomrule
\end{tabular}
\end{table}

% Table S17 — TP/FP and TP/TN waveform correlations
\begin{table}
\centering
\setlength{\tabcolsep}{6pt}
\caption{%
Outcome-conditioned waveform correlations at the detection channel (three-channel montage, fused decision). For each subject, single-trial waveforms at the detection channel (Table~\ref{tab:datasets}) were averaged separately for the four prediction$\times$label outcomes: true positive (TP), false positive (FP), true negative (TN), and false negative (FN). TP$\leftrightarrow$FP and TP$\leftrightarrow$TN report the per-subject Pearson correlation between the mean TP waveform and the mean FP (respectively TN) waveform, summarized as mean $\pm$ standard deviation across subjects. Values are computed from a single reference training seed (seed 1); the standard deviations therefore reflect cross-subject variability. Table~\ref{sup:tab-corr-perseed} reports the corresponding cross-subject means aggregated across all five seeds. TP$\leftrightarrow$FP $>$ TP$\leftrightarrow$TN indicates that false positives morphologically resemble true positives more than true negatives do, consistent with classification operating through waveform--prototype correspondence (Section~\ref{sec:physiological-structure} of the main text). Det.\ ch: detection channel on which prototypes are defined (Table~\ref{tab:datasets}).%
}
\label{sup:tab-corr}
\begin{tabular}{llcc}
\toprule
Dataset & Det.\ ch & TP$\leftrightarrow$FP & TP$\leftrightarrow$TN \\
\midrule
HRI & Cz & $+0.61 \pm 0.20$ & $+0.33 \pm 0.27$ \\
ERN & FCz & $+0.69 \pm 0.22$ & $+0.33 \pm 0.28$ \\
LRP & C3 & $+0.84 \pm 0.16$ & $+0.80 \pm 0.26$ \\
BNCI & Cz & $+0.76 \pm 0.21$ & $+0.37 \pm 0.43$ \\
N170 & PO8 & $+0.70 \pm 0.22$ & $+0.56 \pm 0.26$ \\
P300 & Pz & $+0.60 \pm 0.29$ & $+0.55 \pm 0.34$ \\
N2pc & PO7 & $+0.86 \pm 0.12$ & $+0.81 \pm 0.14$ \\
MMN & FCz & $+0.77 \pm 0.13$ & $-0.31 \pm 0.31$ \\
N400 & CPz & $+0.71 \pm 0.20$ & $+0.46 \pm 0.31$ \\
\bottomrule
\end{tabular}
\end{table}

% Table S18 — Per-seed waveform correlation stability
\begin{table}
\centering
\setlength{\tabcolsep}{6pt}
\caption{%
Per-seed stability of the outcome-conditioned waveform ordering (three-channel montage, fused decision), companion to Table~\ref{sup:tab-corr}. Because Table~\ref{sup:tab-corr} reports a single reference seed, its values differ slightly from the five-seed means below. For each dataset and training seed, TP$\leftrightarrow$FP and TP$\leftrightarrow$TN are the cross-subject mean of the per-subject Pearson correlations defined in Table~\ref{sup:tab-corr}; columns report the mean $\pm$ standard deviation of those cross-subject means across the five seeds. FP${>}$TN seeds: number of seeds (out of five) on which TP$\leftrightarrow$FP exceeded TP$\leftrightarrow$TN. Min.\ margin: smallest per-seed difference (TP$\leftrightarrow$FP $-$ TP$\leftrightarrow$TN). The ordering TP$\leftrightarrow$FP $>$ TP$\leftrightarrow$TN held on all five seeds for every dataset, indicating that the result is not an artifact of a particular random initialization.%
}
\label{sup:tab-corr-perseed}
\begin{tabular}{llcccc}
\toprule
Dataset & Det.\ ch & TP$\leftrightarrow$FP & TP$\leftrightarrow$TN & FP$>$TN seeds & Min.\ margin \\
\midrule
HRI & Cz & $+0.63 \pm 0.01$ & $+0.32 \pm 0.01$ & 5/5 & $+0.29$ \\
ERN & FCz & $+0.68 \pm 0.01$ & $+0.33 \pm 0.01$ & 5/5 & $+0.34$ \\
LRP & C3 & $+0.83 \pm 0.01$ & $+0.81 \pm 0.00$ & 5/5 & $+0.01$ \\
BNCI & Cz & $+0.75 \pm 0.01$ & $+0.37 \pm 0.00$ & 5/5 & $+0.37$ \\
N170 & PO8 & $+0.70 \pm 0.00$ & $+0.55 \pm 0.01$ & 5/5 & $+0.14$ \\
P300 & Pz & $+0.60 \pm 0.02$ & $+0.55 \pm 0.01$ & 5/5 & $+0.03$ \\
N2pc & PO7 & $+0.84 \pm 0.01$ & $+0.80 \pm 0.00$ & 5/5 & $+0.03$ \\
MMN & FCz & $+0.77 \pm 0.00$ & $-0.30 \pm 0.01$ & 5/5 & $+1.05$ \\
N400 & CPz & $+0.70 \pm 0.01$ & $+0.47 \pm 0.01$ & 5/5 & $+0.21$ \\
\bottomrule
\end{tabular}
\end{table}

% Table S19 — Routing grounding, 3-channel
\begin{table}
\scriptsize
\centering
\setlength{\tabcolsep}{5pt}
\caption{%
Routing grounding interventions at the three-channel montage (mean AUROC, LOSO, seed-averaged across five seeds). The ERP-XTTN column reports the full two-factor model AUROC (from main text Table~\ref{tab:auroc}) as a reference. Column definitions are identical to Table~\ref{sup:tab-grounding-full-routing}. Datasets ordered by mean AUROC across all five methods, descending. Standard deviations are omitted and are available in the released code \cite{wyman2026erpxttn_code}. Abbreviations as in Table~\ref{sup:tab-fullmontage-auroc}.%
}
\label{sup:tab-grounding-3-routing}
\begin{tabular}{lccccccccc}
\toprule
Dataset & ERP-XTTN & Routing-only & Time-rev. & Noise null & Phase-rand. & Polarity-inv. & Cross-comp. & Proto-perm & Trial-perm \\
\midrule
HRI & 0.84 & 0.81 & 0.80 & 0.50 & 0.45 & 0.19 & 0.19 & 0.48 & 0.51 \\
ERN & 0.82 & 0.81 & 0.81 & 0.46 & 0.45 & 0.19 & 0.24 & 0.49 & 0.51 \\
LRP & 0.76 & 0.75 & 0.67 & 0.48 & 0.34 & 0.25 & 0.25 & 0.49 & 0.51 \\
BNCI & 0.77 & 0.74 & 0.73 & 0.56 & 0.56 & 0.26 & 0.27 & 0.49 & 0.51 \\
N170 & 0.74 & 0.68 & 0.67 & 0.48 & 0.50 & 0.32 & 0.36 & 0.50 & 0.51 \\
P300 & 0.66 & 0.64 & 0.63 & 0.50 & 0.53 & 0.36 & 0.45 & 0.52 & 0.51 \\
N2pc & 0.64 & 0.61 & 0.60 & 0.51 & 0.48 & 0.39 & 0.48 & 0.51 & 0.50 \\
MMN & 0.59 & 0.58 & 0.58 & 0.50 & 0.50 & 0.42 & 0.44 & 0.50 & 0.50 \\
N400 & 0.56 & 0.54 & 0.54 & 0.50 & 0.52 & 0.46 & 0.50 & 0.50 & 0.51 \\
\midrule
\textit{Mean} & 0.71 & 0.69 & 0.67 & 0.50 & 0.48 & 0.31 & 0.35 & 0.50 & 0.51 \\
\bottomrule
\end{tabular}
\end{table}

% Table S20 — Amplitude grounding, 3-channel
\begin{sidewaystable}
\centering
\setlength{\tabcolsep}{5pt}
\caption{%
Amplitude grounding controls at the three-channel montage (mean AUROC, LOSO, seed-averaged across five seeds). The ERP-XTTN column reports the full two-factor model AUROC from main text Table~\ref{tab:auroc} as a reference. Column definitions are identical to Table~\ref{sup:tab-grounding-full-amp}. For each fold and feature family, AUROC is the maximum orientation-invariant AUROC over individual features. These values constitute a descriptive best-feature diagnostic rather than the performance of a trained classifier or the logistic combiner. At three channels the diagnostic searches across $KC$ monopolar features and $K\binom{C}{2}$ bipolar features, for up to 24 features in total at $K=4$ and $C=3$. Datasets ordered by mean AUROC across all five methods, descending. Standard deviations are omitted and are available in the released code \cite{wyman2026erpxttn_code}. Abbreviations as in Table~\ref{sup:tab-fullmontage-auroc}.%
}
\label{sup:tab-grounding-3-amp}
\begin{tabular}{l c cccc cccc}
\toprule
 & & \multicolumn{4}{c}{Channel MF} & \multicolumn{4}{c}{Contrast MF} \\
\cmidrule(lr){3-6} \cmidrule(lr){7-10}
Dataset & ERP-XTTN & On-target & Off-target (early) & Off-target (late) & Permutation & On-target & Off-target (early) & Off-target (late) & Permutation \\
\midrule
HRI & 0.84 & 0.74 & 0.58 & 0.74 & 0.55 & 0.75 & 0.59 & 0.72 & 0.55 \\
ERN & 0.82 & 0.79 & 0.77 & 0.78 & 0.59 & 0.78 & 0.77 & 0.78 & 0.59 \\
LRP & 0.76 & 0.66 & 0.63 & 0.59 & 0.55 & 0.74 & 0.72 & 0.61 & 0.55 \\
BNCI & 0.77 & 0.71 & 0.57 & 0.70 & 0.55 & 0.70 & 0.57 & 0.70 & 0.54 \\
N170 & 0.74 & 0.76 & 0.58 & 0.69 & 0.59 & 0.74 & 0.58 & 0.69 & 0.59 \\
P300 & 0.66 & 0.64 & 0.58 & 0.66 & 0.59 & 0.63 & 0.59 & 0.65 & 0.60 \\
N2pc & 0.64 & 0.64 & 0.58 & 0.60 & 0.56 & 0.64 & 0.59 & 0.62 & 0.56 \\
MMN & 0.59 & 0.58 & 0.53 & 0.56 & 0.54 & 0.56 & 0.55 & 0.55 & 0.54 \\
N400 & 0.56 & 0.63 & 0.60 & 0.63 & 0.59 & 0.63 & 0.62 & 0.63 & 0.60 \\
\midrule
\textit{Mean} & 0.71 & 0.68 & 0.60 & 0.66 & 0.57 & 0.69 & 0.62 & 0.66 & 0.57 \\
\bottomrule
\end{tabular}
\end{sidewaystable}

% Table S21 — Two-factor ablation grid
\begin{table}
\centering
\footnotesize
\setlength{\tabcolsep}{6pt}
\caption{%
Architecture ablation grid at the three-channel montage (seed-averaged across five seeds). Each ablation is a single-knob deviation from the default ERP-XTTN configuration of Section~\ref{sec:architecture}, evaluated on four datasets spanning the difficulty range (HRI, ERN, P300, N400) under the five-seed LOSO protocol. AUROC is reported as the mean $\pm$ standard deviation across subjects. The base configuration uses both routing and amplitude factors combined by $L_2$-penalized logistic regression on frozen per-trial outputs (Section~\ref{sec:amplitude}). Ablation arms: routing only and amplitude only remove one factor from the combiner; end-to-end trains the routing pathway and the combiner jointly by backpropagation rather than fitting the combiner on frozen outputs; no whitening replaces the whitened cosine (Equation~\ref{eq:match}) with a raw cosine; no self-attention removes the self-attention layer from the peak-embedding pathway; $K = 2$ and $K = 6$ vary the maximum prototype count against the default $K = 4$; prominence varies the peak-detection threshold (default $0.02$); 2 heads and 8 heads vary the attention-head count (default $H = 4$); learned free head replaces the grounded readout with the unconstrained learned classification head of the original ERP-XTTN \cite{wyman2026}. $\Delta$ = ablation arm AUROC minus base AUROC (negative indicates the ablation reduced performance). Abbreviations as in Table~\ref{sup:tab-fullmontage-auroc}.%
}
\label{sup:tab-ablation}
\begin{tabular}{llcc}
\toprule
Dataset & Ablation & AUROC (mean $\pm$ SD) & $\Delta$ vs base \\
\midrule
HRI & Base (routing+amplitude) & 0.837 $\pm$ 0.101 & -- \\
 & Routing only & 0.814 $\pm$ 0.098 & -0.024 \\
 & Amplitude only & 0.820 $\pm$ 0.100 & -0.017 \\
 & End-to-end head & 0.819 $\pm$ 0.103 & -0.018 \\
 & No whitening & 0.839 $\pm$ 0.103 & +0.002 \\
 & No self-attention & 0.828 $\pm$ 0.109 & -0.009 \\
 & $K=2$ & 0.824 $\pm$ 0.103 & -0.013 \\
 & $K=6$ & 0.836 $\pm$ 0.102 & -0.001 \\
 & Prominence 0.01 & 0.835 $\pm$ 0.102 & -0.002 \\
 & Prominence 0.05 & 0.834 $\pm$ 0.102 & -0.003 \\
 & 2 heads & 0.833 $\pm$ 0.103 & -0.004 \\
 & 8 heads & 0.838 $\pm$ 0.102 & +0.001 \\
 & Learned free head & 0.823 $\pm$ 0.106 & -0.014 \\
\midrule
ERN & Base (routing+amplitude) & 0.824 $\pm$ 0.110 & -- \\
 & Routing only & 0.813 $\pm$ 0.107 & -0.011 \\
 & Amplitude only & 0.797 $\pm$ 0.121 & -0.027 \\
 & End-to-end head & 0.810 $\pm$ 0.111 & -0.015 \\
 & No whitening & 0.829 $\pm$ 0.107 & +0.005 \\
 & No self-attention & 0.807 $\pm$ 0.115 & -0.017 \\
 & $K=2$ & 0.815 $\pm$ 0.114 & -0.009 \\
 & $K=6$ & 0.824 $\pm$ 0.110 & +0.000 \\
 & Prominence 0.01 & 0.824 $\pm$ 0.113 & 0.000 \\
 & Prominence 0.05 & 0.823 $\pm$ 0.115 & -0.002 \\
 & 2 heads & 0.819 $\pm$ 0.115 & -0.006 \\
 & 8 heads & 0.826 $\pm$ 0.109 & +0.002 \\
 & Learned free head & 0.823 $\pm$ 0.113 & -0.002 \\
\midrule
P300 & Base (routing+amplitude) & 0.664 $\pm$ 0.071 & -- \\
 & Routing only & 0.644 $\pm$ 0.068 & -0.020 \\
 & Amplitude only & 0.635 $\pm$ 0.071 & -0.030 \\
 & End-to-end head & 0.645 $\pm$ 0.073 & -0.020 \\
 & No whitening & 0.667 $\pm$ 0.076 & +0.003 \\
 & No self-attention & 0.661 $\pm$ 0.078 & -0.003 \\
 & $K=2$ & 0.662 $\pm$ 0.066 & -0.002 \\
 & $K=6$ & 0.669 $\pm$ 0.075 & +0.005 \\
 & Prominence 0.01 & 0.667 $\pm$ 0.072 & +0.002 \\
 & Prominence 0.05 & 0.664 $\pm$ 0.067 & 0.000 \\
 & 2 heads & 0.664 $\pm$ 0.072 & 0.000 \\
 & 8 heads & 0.662 $\pm$ 0.075 & -0.003 \\
 & Learned free head & 0.629 $\pm$ 0.071 & -0.035 \\
\midrule
N400 & Base (routing+amplitude) & 0.565 $\pm$ 0.077 & -- \\
 & Routing only & 0.544 $\pm$ 0.061 & -0.020 \\
 & Amplitude only & 0.578 $\pm$ 0.084 & +0.013 \\
 & End-to-end head & 0.571 $\pm$ 0.081 & +0.006 \\
 & No whitening & 0.567 $\pm$ 0.080 & +0.003 \\
 & No self-attention & 0.574 $\pm$ 0.083 & +0.009 \\
 & $K=2$ & 0.564 $\pm$ 0.074 & -0.001 \\
 & $K=6$ & 0.566 $\pm$ 0.076 & +0.001 \\
 & Prominence 0.01 & 0.565 $\pm$ 0.077 & 0.000 \\
 & Prominence 0.05 & 0.563 $\pm$ 0.071 & -0.001 \\
 & 2 heads & 0.566 $\pm$ 0.080 & +0.001 \\
 & 8 heads & 0.568 $\pm$ 0.080 & +0.003 \\
 & Learned free head & 0.513 $\pm$ 0.045 & -0.052 \\
\bottomrule
\end{tabular}
\end{table}

% Table S22 — Polarity-inversion contrast: grounded vs learned readout
\begin{table}[t]
\centering
\setlength{\tabcolsep}{6pt}
\caption{%
Effect of prototype polarity inversion on the grounded routing readout (routing factor only) versus the unconstrained learned classification head of the original ERP-XTTN \cite{wyman2026}, at the three-channel montage (mean AUROC, LOSO, seed-averaged across five seeds). Both readouts operate over the identical attention tensor and identical frozen prototypes; only the readout differs. Inversion is applied at inference on the frozen model. Noise-null values for the grounded arm are in Table~\ref{sup:tab-grounding-3-routing}. Datasets ordered by mean AUROC across all five methods, descending. Abbreviations as in Table~\ref{sup:tab-fullmontage-auroc}.%
}
\label{sup:tab-polarity-contrast}
\begin{tabular}{lcccc}
\toprule
Dataset & ERP-XTTN Routing-only & Polarity-inv. & Learned Head (Intact) & Learned Head (Polarity-inv.) \\
\midrule
HRI & 0.81 $\pm$ 0.10 & 0.19 $\pm$ 0.10 & 0.82 $\pm$ 0.11 & 0.82 $\pm$ 0.11 \\
ERN & 0.81 $\pm$ 0.11 & 0.19 $\pm$ 0.11 & 0.82 $\pm$ 0.12 & 0.82 $\pm$ 0.12 \\
P300 & 0.64 $\pm$ 0.07 & 0.36 $\pm$ 0.07 & 0.63 $\pm$ 0.08 & 0.63 $\pm$ 0.08 \\
N400 & 0.54 $\pm$ 0.07 & 0.46 $\pm$ 0.07 & 0.51 $\pm$ 0.06 & 0.51 $\pm$ 0.06 \\
\bottomrule
\end{tabular}
\end{table}

%% file: supplementary-figures.tex
\begin{figure}[p]
\centering
\includegraphics[width=0.8\textwidth]{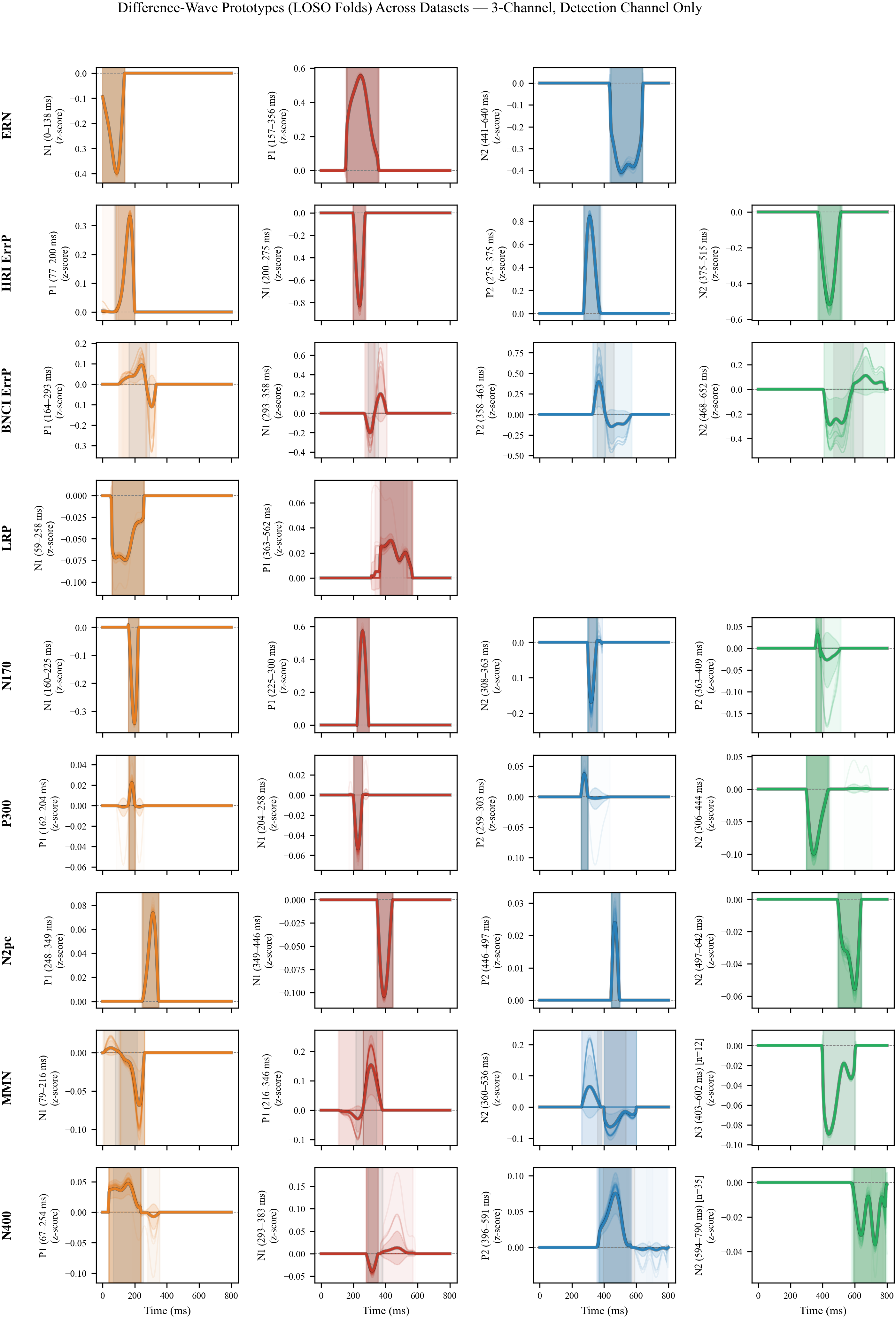}
\caption{%
Difference-wave prototypes (detection channel only, 3-channel condition) across all nine datasets. Each row corresponds to one dataset; each column to one prototype slot. The retained prototype count $K$ was determined independently within each training fold. Modal $K$ was 2 for LRP, 3 for ERN and MMN, and 4 for BNCI, HRI, N170, N2pc, N400, and P300. Only MMN and N400 varied across folds, both between $K = 3$ and $K = 4$. Thick lines show the mean across LOSO folds; thin lines show individual fold prototypes. Prototype windows (shaded) are derived automatically from prominent extrema in the training-fold grand-average difference wave (Section~\ref{sec:prototypes}). Prototypes are labeled by polarity and order of occurrence within each dataset. Prototype morphology was stable across folds for most datasets: mean pairwise correlations exceeded $r=0.88$ for seven of the nine datasets. BNCI showed the lowest stability ($r=0.798$), followed by MMN ($r=0.838$); per-fold values are available in the released code \cite{wyman2026erpxttn_code}. Abbreviations as in Table~\ref{sup:tab-fullmontage-auroc}.%
}
\label{sup:fig-prototypes-all}
\end{figure}

\begin{figure}[p]
\centering
\includegraphics[width=0.9\textwidth]{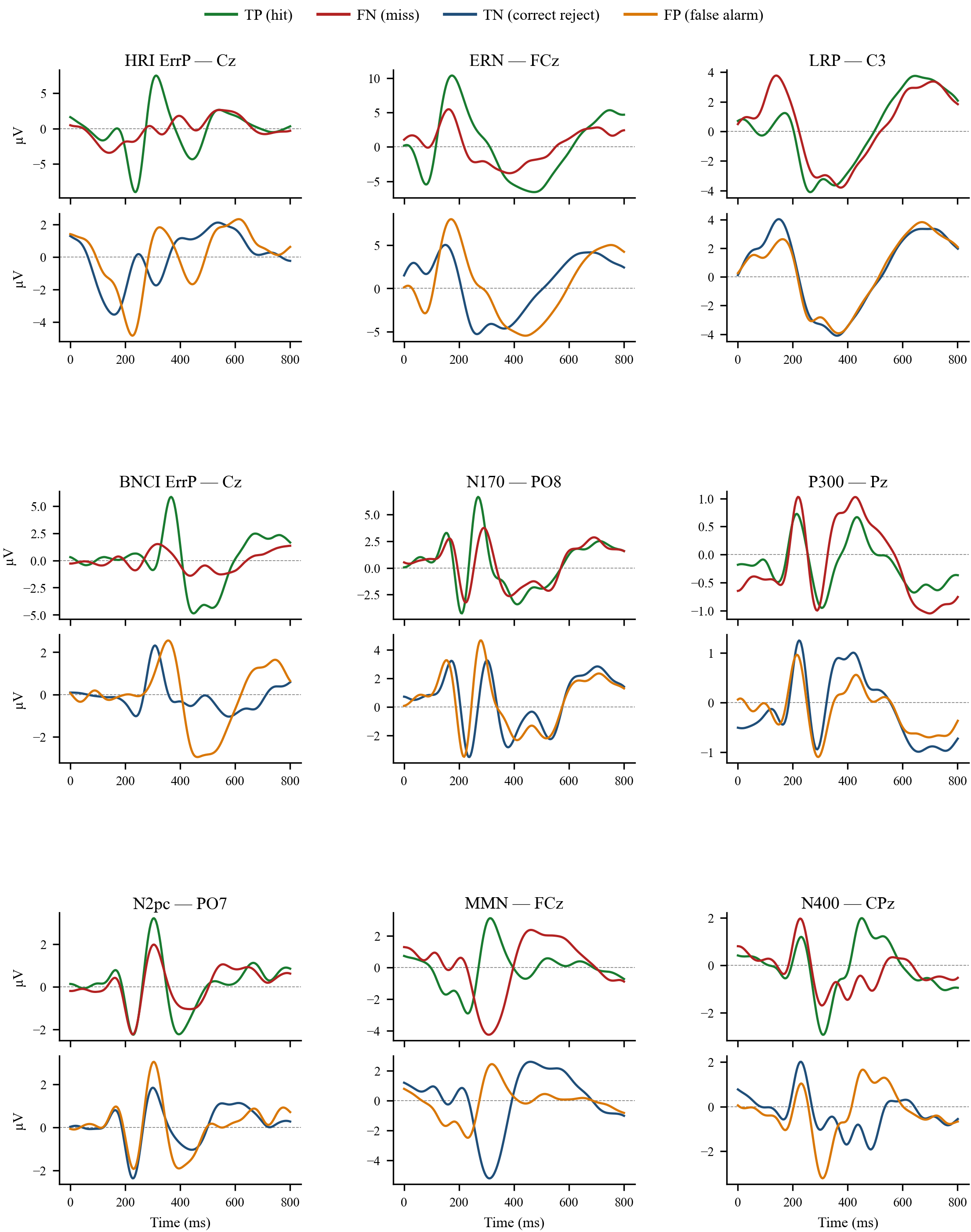}
\caption{%
Outcome-conditioned grand-average waveforms at the detection channel for all nine datasets (3-channel configuration), arranged in main text Table~\ref{tab:auroc} order. Each panel shows two subplots: top, true positive (TP, correctly classified positive-class trials) and false negative (FN, missed positive-class trials); bottom, true negative (TN, correctly classified negative-class trials) and false positive (FP, falsely flagged negative-class trials). Waveforms are grand-averaged across all subjects within each dataset. Averaging across subjects suppresses subject-idiosyncratic morphology, so the traces shown here are more similar across outcome conditions than the within-subject correlations of Table~\ref{sup:tab-corr}, which are computed per subject and are correspondingly lower. Dataset name, detection channel, and trial counts are shown per panel. The main-text analysis (Section~\ref{sec:physiological-structure}, Figure~\ref{fig:morphology}) is based on HRI ErrP; this figure extends the comparison to all components. In these grand averages, false positives visibly resemble true positives more than true negatives do, consistent with the TP$\leftrightarrow$FP $>$ TP$\leftrightarrow$TN pattern reported in Table~\ref{sup:tab-corr}. Notable exceptions include LRP, where both correlations are similar, and MMN, where true negatives show an inverted waveform relative to true positives. Abbreviations as in Table~\ref{sup:tab-fullmontage-auroc}.%
}
\label{sup:fig-morphology-all}
\end{figure}